\documentclass[runningheads]{llncs}

\usepackage{eccv}

\usepackage{eccvabbrv}

\usepackage{graphicx}
\usepackage{overpic}
\usepackage{booktabs}
\usepackage[utf8]{inputenc}
\usepackage{multirow}
\usepackage[table,xcdraw]{xcolor}
\usepackage{bm}
\usepackage{pifont}
\usepackage{array}
\usepackage{soul}
\usepackage{listings}
\usepackage{chngcntr}

\lstdefinestyle{prompt}{
  basicstyle=\ttfamily\footnotesize,
  breaklines=true,
  breakatwhitespace=false,
  columns=fullflexible,
  keepspaces=true,
  showstringspaces=false,
  frame=single,
  rulecolor=\color{black!30},
  backgroundcolor=\color{black!2},
  xleftmargin=0.6em,
  framexleftmargin=0.6em,
  framesep=0.4em
}

\AtBeginDocument{\AtBeginDocument{\counterwithout{lstlisting}{section}}}

\usepackage[accsupp]{axessibility}  

\definecolor{lightcyanbg}{RGB}{225, 240, 255} 
\definecolor{lightorangebg}{RGB}{255, 245, 235} 
\definecolor{rank1green}{RGB}{110, 215, 110} 
\definecolor{rank2green}{RGB}{180, 240, 180} 
\definecolor{rank3green}{RGB}{225, 250, 225} 
\definecolor{lightgreen}{RGB}{230, 245, 235} 
\definecolor{lightgraybg}{RGB}{245, 245, 245} 
\definecolor{highlightgray}{RGB}{160, 160, 160} 
\definecolor{highlightorange}{RGB}{250, 230, 215} 

\usepackage{hyperref}

\usepackage{orcidlink}

\begin{document}

\title{EXPLORE-Bench: Egocentric Scene Prediction with Long-Horizon Reasoning} 

\titlerunning{EXPLORE-Bench}

\author{Chengjun Yu\inst{1,*} \and
Xuhan Zhu\inst{2,3,*} \and
Chaoqun Du\inst{2} \and
Pengfei Yu\inst{2,\dagger} \and
Wei Zhai\inst{1,\dagger} \and
Yang Cao\inst{1} \and
Zheng-Jun Zha\inst{1}}

\authorrunning{Yu, C et al.}

\institute{University of Science and Technology of China\\
\and
Foundation Model Team, Li Auto Inc.\\
\and
Tsinghua University *Equal Contribution $\dagger$Corresponding Author\\
\email{\{yucj@mail., wzhai056@, forrest@, zhazj@\}ustc.edu.cn}\\
\email{\{zhuxuhan, duchaoqun, yupengfei1\}@lixiang.com}
\url{https://github.com/JackYu6/EXPLORE-Bench/}}
\maketitle

\begin{abstract}
Multimodal large language models (MLLMs) are increasingly considered as a foundation for embodied agents, yet it remains unclear whether they can reliably reason about the long-term physical consequences of actions from an egocentric viewpoint. We study this gap through a new task, \textbf{E}gocentric \textbf{S}cene \textbf{P}rediction with \textbf{LO}ng-horizon \textbf{RE}asoning: given an initial-scene image and a sequence of atomic action descriptions, a model is asked to predict the final scene after all actions are executed.
To enable systematic evaluation, we introduce \textbf{EXPLORE-Bench}, a benchmark curated from real first-person videos spanning diverse scenarios. Each instance pairs long action sequences with structured final-scene annotations, including object categories, visual attributes, and inter-object relations, which supports fine-grained, quantitative assessment.
Experiments on a range of proprietary and open-source MLLMs reveal a significant performance gap to humans, indicating that long-horizon egocentric reasoning remains a major challenge. We further analyze test-time scaling via stepwise reasoning and show that decomposing long action sequences can improve performance to some extent, while incurring non-trivial computational overhead. Overall, EXPLORE-Bench provides a principled testbed for measuring and advancing long-horizon reasoning for egocentric embodied perception.
\end{abstract}

\begin{figure}[t]
  \centering
  \includegraphics[width=\linewidth]{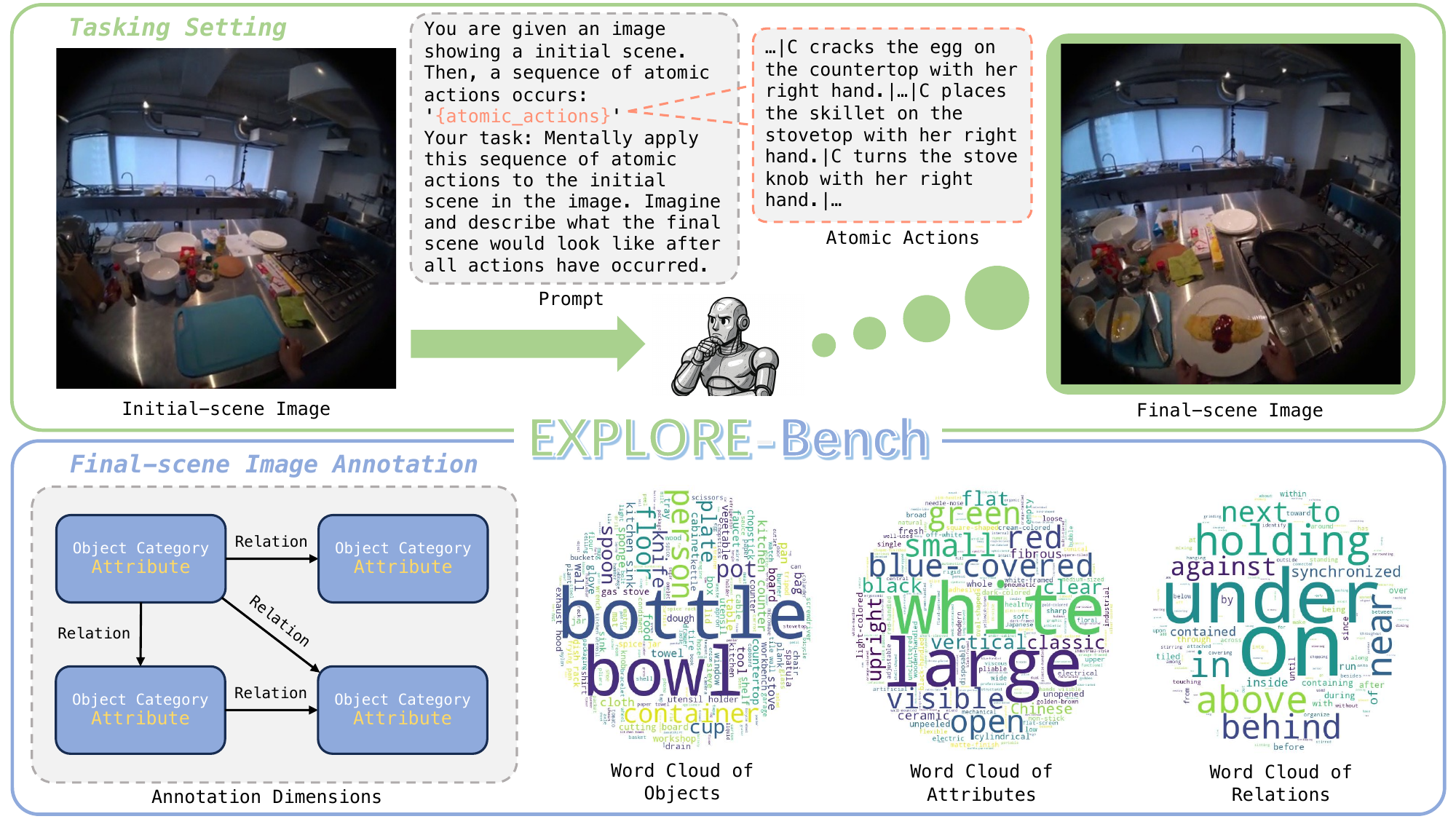}
  \caption{\textbf{Overview of EXPLORE-Bench.} EXPLORE-Bench evaluates MLLMs on a new task: egocentric scene prediction with long-horizon reasoning. We annotate the final scene at the object, attribute, and relation levels to enable fine-grained scene-level evaluation. Note that the prompt is abbreviated for brevity in this figure.}
  \label{fig:teaser}
\end{figure}

\section{Introduction}
\label{sec:intro} 
Humans perceive and act in the physical world primarily from an egocentric (first-person) view. This perspective naturally couples vision with action: what we see affects what we will do, and what we do in turn changes what we will see. Motivated by this tight perception--action loop, egocentric benchmarks \cite{jia2022egotaskqa,lin2022egocentric, mangalam2023egoschema,chen2023egoplan,team2025gemini} have recently drawn increasing attention as a promising means to evaluate the embodied potential of multimodal large language models (MLLMs) \cite {hou2026survey}.
Existing egocentric benchmarks cover a broad set of capabilities. However, a fundamental ability remains under-explored: \emph{egocentric scene prediction that requires long-horizon reasoning}---anticipating the final state of an entire scene after executing a long sequence of actions. 

In this work, we formalize a new task: given an initial-scene image and a long sequence of atomic action descriptions, an MLLM is asked to predict the final scene after all actions are executed, as illustrated in \cref{fig:teaser}. We refer to this as \emph{egocentric scene prediction with long-horizon reasoning}. This capability is essential for MLLM-driven embodied agents to understand how their actions causally affect the physical world, which may support better decision making and planning. Importantly, it also helps avoid unintended negative consequences when pursuing a goal---for instance, removing a book from the bottom of a stack may destabilize the pile and cause other objects to fall. Such outcomes often depend on reasoning over atomic-action steps and maintaining a coherent representation of what changes and what remains invariant throughout the interaction.

While several egocentric benchmarks \cite{yuaneoc, li2025egocross} include evaluations related to state prediction, they typically focus on instant (or short-term) and localized state changes (\eg, the state of a single object after a visual cue) and do not provide a systematic protocol to evaluate scene-level predictions under long action horizons (see \cref{tab:bench_comparison}). Moreover, without structured annotations of objects, attributes, and relations in the final scene, it is difficult to quantify what an MLLM gets right or wrong beyond coarse textual similarity. To bridge this gap, we design a scalable annotation pipeline and build a new benchmark that enables fine-grained and quantitative evaluation of long-horizon egocentric scene prediction.

Specifically, we introduce \textbf{EXPLORE-Bench} (\cref{sec:benchmark}), a well-curated benchmark for evaluating MLLMs on egocentric scene prediction with long-horizon reasoning. Our benchmark contains 1,157 instances. Each instance consists of (i) an initial-scene image, (ii) a sequence of atomic actions with an average length of 113, and (iii) structured annotations for the final-scene image, including object categories, visual attributes, and inter-object relations. The dataset is derived from real first-person videos spanning diverse scenarios, and its structured annotations support systematic evaluation of models' ability to track and reason about long-term consequences of actions.
We benchmark multiple proprietary and open-source MLLMs on EXPLORE-Bench and find a substantial gap to human performance, indicating that long-horizon egocentric scene prediction remains challenging for current models (\cref{sec:eval}). We further explore multi-step (decomposed) inference strategies and analyze when stepwise reasoning helps or hurts, providing insights for test-time scaling \cite{zhang2025survey} in this setting (\cref{sec:infer}). Moreover, we specifically examine the models’ performance in abnormal cases (\cref{sec:abnormal}). 
Our main contributions are summarized as follows:
\begin{itemize}
  \item We propose a new task, \emph{egocentric scene prediction with long-horizon reasoning}, to evaluate whether MLLM-driven agents can anticipate how their actions affect the physical world over long action horizons.
  \item We design a scalable data annotation pipeline and construct \textbf{EXPLORE-Bench} to support systematic, fine-grained evaluation of this task.
  \item We evaluate a range of proprietary and open-source MLLMs and show that, compared to humans, current models perform poorly on long-horizon egocentric scene prediction, especially in abnormal cases.
  \item We explore and analyze the effect of stepwise reasoning on model performance, offering practical insights for test-time scaling on egocentric scene prediction with long-horizon reasoning.
\end{itemize}

\begin{table*}[t]
\caption{\textbf{Comparison of widely adopted egocentric/embodied benchmarks with EXPLORE-Bench.} Here, \textbf{completeness} refers to whether the atomic actions can describe a complete long-horizon task (\eg, making an omelet).}
\label{tab:bench_comparison}
\centering
\resizebox{\textwidth}{!}{
\begin{tabular}{l|ccc|cc|cccc}
\toprule
\multirow{2}{*}{\textbf{Benchmarks}} & \multicolumn{3}{c|}{\textbf{Atomic Actions}} & \multicolumn{2}{c|}{\textbf{Scene Types}} & \multicolumn{4}{c}{\textbf{Evaluation Dimensions}} \\
& \textbf{length} & \textbf{influence} & \textbf{completeness} &  \textbf{normal} & \textbf{abnormal} & \textbf{object} & \textbf{attribute} & \textbf{relation} & \textbf{scene} \\
\midrule
EgoTaskQA \cite{jia2022egotaskqa} & short & regional & \ding{55} & \ding{51} & \ding{55} & \ding{51} & \ding{51} & \ding{51} & \ding{55} \\
EgoMCQ \cite{lin2022egocentric} & short  & regional & \ding{55} & \ding{51} & \ding{55} & \ding{51} & \ding{51} & \ding{51} & \ding{55} \\
EgoSchema \cite{mangalam2023egoschema} & short & regional & \ding{55} & \ding{51} & \ding{55} & \ding{51} & \ding{55} & \ding{51} & \ding{55} \\
EgoPlan-Bench \cite{chen2023egoplan} & short & regional & \ding{55} & \ding{51} & \ding{55} & \ding{51} & \ding{55} & \ding{51} & \ding{55} \\
ERQA \cite{team2025gemini} & short & regional & \ding{55} & \ding{51} & \ding{55} & \ding{51} & \ding{55} & \ding{51} & \ding{55} \\
\arrayrulecolor{black}\midrule
\rowcolor{gray!20} \textbf{EXPLORE-Bench } & long & global & \ding{51} & \ding{51}  & \ding{51} & \ding{51} & \ding{51} & \ding{51} & \ding{51} \\
\arrayrulecolor{black}\bottomrule
\end{tabular}
}
\end{table*}

\section{Related Work}
\noindent{\bf Long-horizon Video Benchmarks.}
Recently, a growing number of benchmarks \cite{wu2024longvideobench, zhou2025mlvu, fu2025video, huang2025online, song2024moviechat, yue2023movie101, mangalam2023egoschema} for long-horizon video understanding and reasoning have emerged, providing standardized platforms for evaluating models’ capabilities in temporal modeling, cross-frame relation extraction, and high-level semantic reasoning. These benchmarks emphasize sustained comprehension over long durations, requiring models to maintain long-term memory, track evolving events and object states, and integrate information across distant segments to answer queries about global narratives or long-range dependencies.
Nevertheless, current long-horizon video benchmarks primarily test coarse event-level understanding, leaving the evaluation of multi-step, fine-grained state changes induced by sequences of atomic actions largely under-explored. Our EXPLORE-Bench instead assesses long-horizon egocentric scene prediction by requiring models to infer the final scene state after executing a sequence of atomic actions, with an emphasis on global scene changes involving multiple objects.

\noindent{\bf Egocentric Benchmarks.}
Contrary to works that focus on the exocentric view \cite{luo2022learning, yang2023grounding, zhai2023exploring}, egocentric benchmarks probe models from the first-person perspective. Existing benchmarks evaluate a broad spectrum of abilities, including recognition \cite{yang2024egochoir,perrett2025hd}, memory \cite{zhou2025x,yang2025egolife}, understanding \cite{liu2026benchmarking,chen2025grounded,fan2019egovqa,jia2022egotaskqa}, planning \cite{yang2025embodiedbench,zhao2025urbanvideo}, generation \cite{yu2025hero,zhang2025egoreact,han2025touch}, and reasoning \cite{wu2025st,yan2025teleego}. Several benchmarks \cite{yuaneoc,dang2025ecbench,cheng2024egothink,zhou2025egotextvqa,li2025egocross} also emphasize broader embodied evaluation across tasks and settings. These efforts collectively demonstrate the promise of egocentric data for assessing embodied intelligence. 
Despite this progress, existing egocentric benchmarks do not include the task of egocentric scene prediction with long-horizon reasoning. Our task formulation differs from other egocentric reasoning benchmarks in that it requires reasoning over \emph{action sequences} rather than videos, which better simulates an agent's reasoning after planning actions but before execution. In terms of data, we provide structured scene annotations with objects, attributes, and relations, enabling systematic evaluation of scene-level predictions. 

\section{EXPLORE-Bench}
\label{sec:benchmark}

\subsection{Overview}
\label{sec:bench_overview}
\begin{figure}[t]
  \centering
  \includegraphics[width=\linewidth]{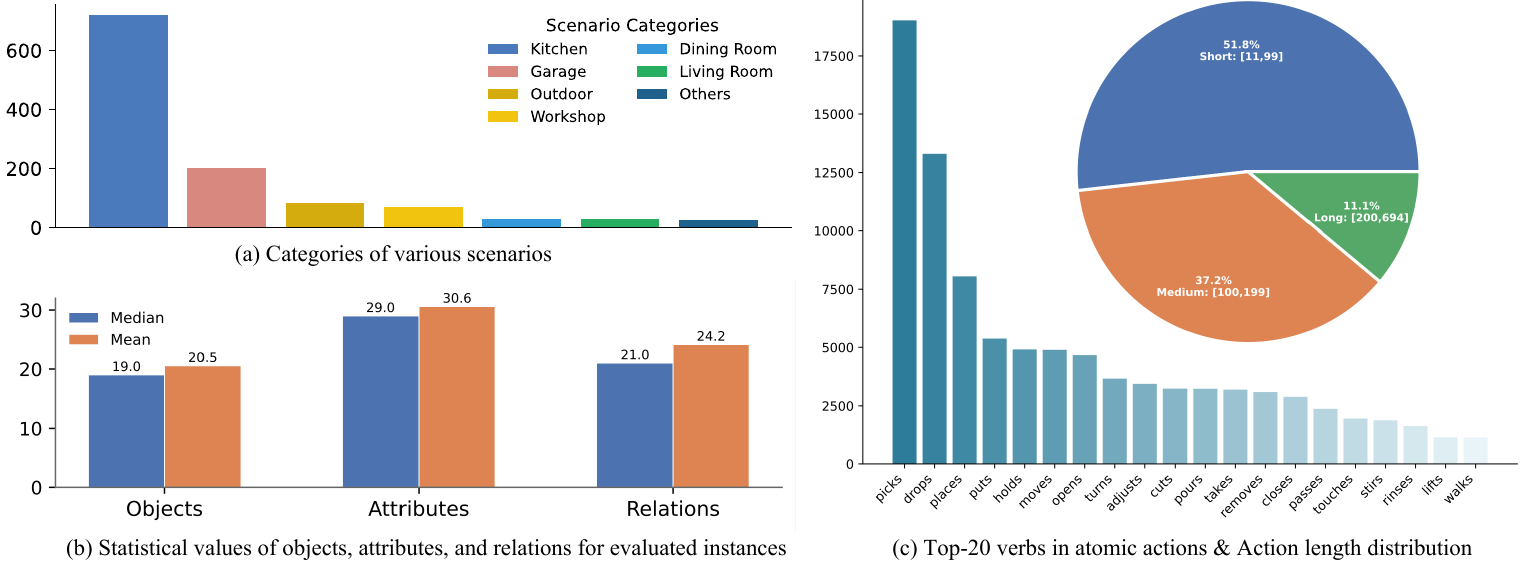}
  \caption{\textbf{Data analysis of EXPLORE-Bench} reflects the rich diversity of scenarios, objects, attributes, relations, and atomic actions.}
  \label{fig:data_stats}
\end{figure}

We introduce EXPLORE-Bench, a well-curated benchmark designed to quantitatively evaluate MLLMs' capability of egocentric scene prediction with long-horizon reasoning. EXPLORE-Bench comprises 1,157 instances derived from 1,157 real videos. These videos are sourced from two publicly available datasets: Ego4D \cite{grauman2022ego4d} and Ego-Exo4D \cite{grauman2024egoexo}, as well as our self-recorded videos captured in diverse scenarios. The videos have an average duration of 358.28 seconds, with the longest lasting 1,524.80 seconds.
Each instance in EXPLORE-Bench contains an image representing the initial scene, a sequence of atomic action descriptions, and annotations for the final-scene image. Across all instances, the number of atomic actions ranges from 11 to 694, with an average of 113 per instance. The final-scene annotations include the object categories present in the scene, the visual attributes of these objects, and the relations among them. These annotations provide a solid foundation for systematically evaluating MLLMs’ predictions of the final scene. We annotate 23,771 objects spanning 1,612 categories, with an average of more than 20 objects per instance. See \cref{fig:data_stats} for more data analysis.

\subsection{Dataset Construction}
\noindent{\bf Data Collection.}
Ego4D \cite{grauman2022ego4d} and Ego-Exo4D \cite{grauman2024egoexo} cover a wide range of human activities. We first filter videos from these two datasets based on the recorded activities to select those suitable for long-horizon reasoning—for example, cooking and bicycle repair are considered valid, whereas dancing or chatting with someone are treated as invalid. For each selected video, we then extract the start and end frames of the main activity as the initial- and final-scene images of a instance, respectively. The timestamped atomic-action annotations in these datasets facilitate this process: using the timestamps of the extracted start and end frames, we retrieve the atomic action descriptions occurring in between. For our self-recorded videos, we follow the same format and annotate atomic actions (as shown in \cref{fig:abnormal_case_1v2}) at a granularity comparable to that of these two datasets. 

\begin{figure}[t]
  \centering
  \includegraphics[width=\linewidth]{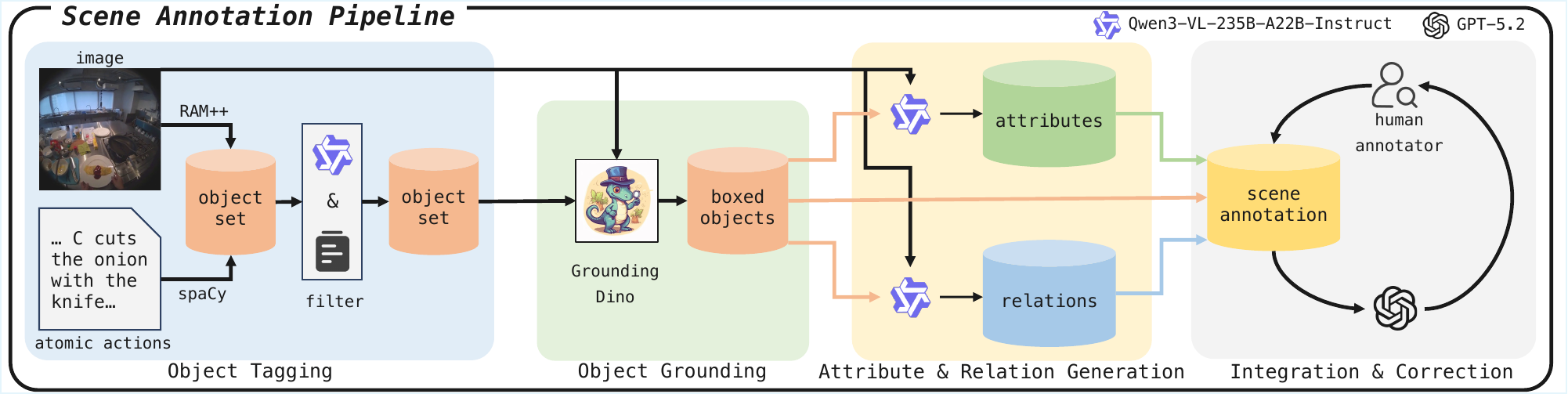}
  \caption{\textbf{Illustration of our scene annotation pipeline.} Rather than having the MLLM generate annotations directly from the images, we adopt a multi-step pipeline to ensure object coverage and the accuracy of attributes and relations, greatly reducing the load of manual annotation required.}
  \label{fig:anno_pipeline}
\end{figure}

\noindent{\bf Scene Annotation.} 
We develop a pipeline to produce high-quality scene annotations at scale, as illustrated in \cref{fig:anno_pipeline}. 

{\it Object Tagging.} 
We first extract object tags from the final-scene images and the atomic action descriptions using the Recognize Anything Plus Model \cite{huang2025open} and the spaCy \cite{spacy} library, respectively, and merge them to maximize object coverage. Next, we remove invalid tags (\eg, synonyms and human body parts) using a filtering scheme that combines an LLM-based filter with rule-based methods.

{\it Object Grounding.} 
The final-scene image and the filtered tags are then fed into Grounding DINO \cite{liu2024grounding} to obtain bounding boxes for the objects. Each detected object instance is represented by a category label with an index (\eg, \texttt{bowl.2}) to facilitate subsequent information matching and integration.

{\it Attribute Generation.} 
We then prompt Qwen3-VL-235B-A22B-Instruct \cite{qwen3vl} to generate descriptions of salient visual attributes for each object based on the final-scene image and the boxed objects, including shape, color, size, texture, and state, \etc. 

{\it Relation Generation.} 
Similarly, we prompt Qwen3-VL-235B-A22B-Instruct to produce inter-object relation triplets \texttt{\((\text{object.0}, \text{relation}, \text{object.1})\)}, covering both spatial (\eg, \texttt{under}) and interaction (\eg, \texttt{holding}) relations. 

{\it Information Integration.} 
We subsequently integrate objects in the final-scene image with their corresponding attributes and relations to form complete scene annotations. Within a single image annotation, different instances of the same category possess distinct attributes or relations.

{\it Annotation Correction.} 
Finally, we feed the image and the resulting annotations into GPT-5.2 \cite{gpt5_2} for annotation correction and augmentation. 


\noindent{\bf Human-in-the-loop Quality Control.}
Human quality control is integrated throughout our dataset construction pipeline. During data collection, videos are first pre-filtered based on the scenario labels provided by the source datasets and then further screened by human reviewers. When extracting initial- and final-scene images from each video, we ensure that the frames are free of motion blur, that only the main human activity occurs between them, and that there is no drastic change in camera viewpoint across the two scenes. We also filter out ambiguous instances in which the action descriptions are insufficient to infer the final scene (\eg, underspecified positions or severe occlusions).
During scene annotation, our prompts and filtering rules are refined through multiple iterations with human verification. More importantly, the final scene annotations are corrected by human annotators. The annotation guidelines are updated multiple times based on annotator feedback. 

\subsection{Evaluation Protocol}
\label{sec:eval_proc}
Our task formulation requires models to predict and describe the final scene, resulting in outputs in the same format as image captioning. Inspired by CompreCap \cite{lu2025benchmarking}, we comprehensively evaluate MLLMs’ predicted scene descriptions from three aspects: object-level coverage, the accuracy of attribute descriptions, and the score of relations. The procedure is as follows:

\noindent{\bf Description Decomposition.} 
The generated description is first split into sub-descriptions by sentence separators (\eg, periods). Nouns are then extracted and lemmatized using spaCy \cite{spacy} to form a set of candidate objects. This process establishes a standardized hierarchical structure, enabling accurate alignment with annotations and enhancing the stability of subsequent evaluations.

\noindent{\bf Object-level Evaluation.}
We employ Sentence-BERT \cite{reimers2019sentence} to extract word embeddings and compute a similarity matrix between the candidate objects and the annotated objects. Matched objects are identified by selecting the mutual maximum values in both rows and columns of the matrix. The object-level coverage $S_{obj}$ is calculated from the refined matrix, quantifying the soft coverage of the annotated objects in the generated description.

\noindent{\bf Attribute-level Evaluation.} 
In this step, we concatenate all sub-descriptions referring to the same annotated object. Then, a standardized prompt guides an LLM \cite{yang2025qwen3} to score attribute descriptions on a 0–5 scale based on the corresponding annotated phrases. The average score across all matched objects yields the attribute-level score $S_{att}$.

\noindent{\bf Relation-level Evaluation.}
The same scoring pipeline as in the attribute-level evaluation is applied to rate the alignment between generated and annotated relations. The resulting relation-level score $S_{rel}$ quantitatively measures the quality of relation descriptions.

\noindent{\bf Metric Unification.}
$S_{obj}$, $S_{att}$ and $S_{rel}$ are linearly scaled to a 0–100 range. Then, a weighted average of the scaled scores is computed to obtain the unified score $S_{uni}$, providing a comprehensive quantification of the scene description:
\begin{equation}
  S_{uni}=w_1 S_{obj}+w_2 (20 \cdot S_{att})+w_3(20 \cdot S_{rel}).
  \label{eq:s_uni}
\end{equation}
We assign weights following CompreCap \cite{lu2025benchmarking} ($w_1=0.25, w_2=0.35, w_3=0.40$).

\section{Evaluation on EXPLORE-Bench}
\label{sec:eval}

\subsection{Evaluation Setup}
\noindent{\bf Benchmark Models.} 
We comprehensively evaluate multimodal foundation models across diverse model families, including both proprietary MLLMs and open-source MLLMs. For proprietary MLLMs, we evaluate GPT-5.2 \cite{gpt5_2} and Gemini-3 \cite{gemini3}. For open-source MLLMs, we consider models from Qwen2-VL \cite{qwen2vl}, Qwen2.5-VL \cite{qwen2_5vl}, Qwen3-VL \cite{qwen3vl}, InternVL3.5 \cite{internvl3_5}, LLaVA-OneVision-1.5 \cite{llava_ov1_5}, Keye-VL-1.5 \cite{keyevl1_5}, Ovis2.5 \cite{ovis2_5}, MiMo-VL \cite{mimovl}, MiniCPM-V-4.5 \cite{minicpm_v4_5}, GLM-4.6V \cite{glm4_6v}, and Step3-VL \cite{step3vl}, encompassing both non-thinking and thinking models. 
In addition, we evulate Embodied-Reasoner \cite{zhang2025embodied} and EgoThinker \cite{pei2025egothinker}, which are tailored for embodied or egocentric reasoning.
For models that support non-thinking and thinking modes, we report their performance in both modes. All models use the same prompt for final-scene prediction. We use greedy decoding to ensure reproducibility. 

\noindent{\bf Dataset Partitioning.}
We partition all instances in EXPLORE-Bench into three subsets according to the length of the contained atomic-action sequence: \texttt{Short}, \texttt{Medium}, and \texttt{Long}. Specifically, the \texttt{Short} subset contains sequences of length 11–99, the \texttt{Medium} subset contains sequences of length 100–199, and the \texttt{Long} subset contains sequences ranging from 200 to 694. Note that the notion of ``short'' or ``long'' here is only relative. These three subsets contain 599, 430, and 128 instances, respectively (the distribution visualization is shown in \cref{fig:data_stats}). We report the performance of the models on each subset as well as on the full dataset (\texttt{Full}).

\noindent{\bf Human Performance Evaluation.} 
We sample a subset of 100 instances from the \texttt{Short}, \texttt{Medium}, and \texttt{Long} subsets in a 52:37:11 ratio, which we refer to as EXPLORE-Bench (tiny). Human participants independently describe the final scene of each instance and their performance is evaluated using the same metrics as those used for the models. For comparison, we also report the performance of several top-tier MLLMs on EXPLORE-Bench (tiny).

\noindent{\bf Evaluation Metrics.}  
We evaluate predicted scene descriptions using the protocol in \cref{sec:eval_proc} and report $S_{obj}$, $S_{att}$, $S_{rel}$, and $S_{uni}$ on the \texttt{Short}, \texttt{Medium}, \texttt{Long}, and \texttt{Full} sets, respectively. Using an LLM as an evaluator is increasingly common \cite{lu2025benchmarking, majumdar2024openeqa, zheng2023judging, wang2025vl4gaze}, and we further verify the consistency between our evaluation protocol and human judgments. We sample 100 instances from EXPLORE-Bench and collect final-scene descriptions generated by multiple models. Human evaluators then score each description in a source-blind manner. The Spearman correlation \(\rho\) between scores produced by Qwen3-8B \cite{yang2025qwen3} (used as the LLM-based scorer) and human scores is 0.919. In comparison, pairwise correlations among human evaluators range from 0.912 to 0.936, indicating that the LLM scorer is largely consistent with human judgments. 

\begin{table*}[!t]
\caption{\textbf{Evaluation results on EXPLORE-Bench.} \texttt{Short}, \texttt{Medium} and \texttt{Long} mean the subsets with short, medium and long atomic-action sequences, respectively. \texttt{Full} denotes the full dataset. \colorbox{rank1green}{Dark green} and \colorbox{rank2green}{light green} indicate the best and the second best result among all models. \colorbox{highlightorange}{Orange} denotes using Qwen2-VL-7B-Instruct as the base model. $^{\dagger}$ denotes results on EXPLORE-Bench (tiny) set. $^{\#}$ and $^*$ represent the non-thinking and thinking mode, respectively.}
\label{tab:eval_main}
\centering
\renewcommand{\arraystretch}{1.1} 
\setlength{\tabcolsep}{2.5pt} 

\resizebox{\textwidth}{!}{
\begin{tabular}{l|cccc|cccc|cccc|cccc}
\toprule
\multirow{2}{*}{\textbf{Methods}} & \multicolumn{4}{c|}{\texttt{Short}} & \multicolumn{4}{c|}{\texttt{Medium}} & \multicolumn{4}{c|}{\texttt{Long}} & \multicolumn{4}{c}{\texttt{Full}} \\
 & {$\bm{S_{obj}}$} & {$\bm{S_{att}}$} & {$\bm{S_{rel}}$} & {$\bm{S_{uni}}$} & {$\bm{S_{obj}}$} & {$\bm{S_{att}}$} & {$\bm{S_{rel}}$} & {$\bm{S_{uni}}$} & {$\bm{S_{obj}}$} & {$\bm{S_{att}}$} & {$\bm{S_{rel}}$} & {$\bm{S_{uni}}$} & {$\bm{S_{obj}}$} & {$\bm{S_{att}}$} & {$\bm{S_{rel}}$} & {$\bm{S_{uni}}$} \\ \midrule \multicolumn{17}{c}{\textit{Performance on EXPLORE-Bench (tiny)}} \\ \midrule
\rowcolor{lightcyanbg} Human$^{\dagger}$ & 72.81 & 2.64 & 3.11 & 61.56 & 70.43 & 2.46 & 2.88 & 57.87 & 67.86 & 2.11 & 2.47 & 51.50 & 71.38 & 2.51 & 2.95 & 59.08 \\
GPT-5.2-Chat$^{\dagger}$ & 59.93 & 1.73 & 2.79 & 49.42 & 60.39 & 1.72 & 2.66 & 48.40 & 58.25 & 1.65 & 2.61 & 46.97 & 59.92 & 1.72 & 2.72 & 48.77 \\
Gemini-3-Pro$^{\dagger}$ & 59.76 & 1.76 & 2.74 & 49.15 & 60.92 & 1.78 & 2.76 & 49.80 & 60.40 & 1.69 & 2.63 & 47.97 & 60.26 & 1.76 & 2.74 & 49.26 \\
Qwen3-VL-8B-Instruct$^{\dagger}$ & 61.71 & 1.91 & 2.89 & 51.95 & 62.03 & 1.90 & 2.84 & 51.55 & 55.54 & 1.69 & 2.56 & 46.17 & 61.15 & 1.88 & 2.84 & 51.16 \\ 
Qwen3-VL-8B-Thinking$^{\dagger}$ & 65.65 & 1.96 & 2.96 & 53.83 & 61.40 & 1.83 & 2.75 & 50.19 & 57.99 & 1.67 & 2.55 & 46.59 & 63.24 & 1.88 & 2.83 & 51.69 \\ \midrule
\multicolumn{17}{c}{\textit{Proprietary Multimodal Foundation Models (API)}} \\ \midrule
GPT-5.2-Chat & 59.91 & 1.74 & 2.70 & 48.71 & 59.88 & 1.67 & 2.65 & 47.85 & 58.06 & 1.65 & 2.61 & 46.91 & 59.69 & 1.70 & 2.67 & 48.19 \\
Gemini-3-Flash & 60.31 & 1.84 & 2.78 & 50.27 & 59.33 & 1.76 & 2.71 & 48.84 & 58.27 & \cellcolor{rank2green}1.72 & 2.69 & 48.18 & 59.72 & 1.80 & 2.75 & 49.51 \\
Gemini-3-Pro & 61.29 & 1.81 & 2.77 & 50.11 & \cellcolor{rank2green}60.99 & 1.75 & 2.74 & 49.44 & \cellcolor{rank1green}59.17 & 1.70 & \cellcolor{rank2green}2.70 & \cellcolor{rank1green}48.31 & \cellcolor{rank2green}60.94 & 1.77 & 2.75 & 49.66 \\ \midrule
\multicolumn{17}{c}{\textit{Open-source Non-thinking Multimodal Foundation Models}} \\ \midrule
Qwen2.5-VL-3B-Instruct & 48.17 & 1.36 & 2.27 & 39.70 & 44.19 & 1.22 & 2.17 & 36.92 & 39.92 & 1.09 & 2.08 & 34.28 & 45.78 & 1.28 & 2.21 & 38.07 \\
\rowcolor{highlightorange} Qwen2-VL-7B-Instruct & 47.64 & 1.34 & 2.23 & 39.14 & 46.04 & 1.27 & 2.16 & 37.73 & 43.79 & 1.22 & 2.15 & 36.73 & 46.62 & 1.30 & 2.20 & 38.35 \\
Keye-VL-1.5-8B$^{\#}$ & 49.70 & 1.34 & 2.24 & 39.78 & 49.49 & 1.34 & 2.24 & 39.69 & 46.15 & 1.25 & 2.17 & 37.68 & 49.23 & 1.33 & 2.23 & 39.51 \\
LLaVA-OneVision-1.5-4B & 51.15 & 1.42 & 2.33 & 41.37 & 49.17 & 1.33 & 2.28 & 39.85 & 46.35 & 1.28 & 2.26 & 38.64 & 49.88 & 1.37 & 2.31 & 40.50 \\
Qwen3-VL-2B-Instruct & 52.18 & 1.48 & 2.55 & 43.77 & 49.03 & 1.37 & 2.45 & 41.38 & 43.27 & 1.18 & 2.28 & 37.34 & 50.02 & 1.40 & 2.48 & 42.17 \\
Qwen2.5-VL-7B-Instruct & 52.80 & 1.52 & 2.46 & 43.58 & 51.25 & 1.46 & 2.41 & 42.32 & 47.78 & 1.34 & 2.36 & 40.22 & 51.67 & 1.48 & 2.43 & 42.74 \\
LLaVA-OneVision-1.5-8B & 53.25 & 1.54 & 2.51 & 44.13 & 51.21 & 1.47 & 2.44 & 42.63 & 47.62 & 1.38 & 2.41 & 40.83 & 51.87 & 1.49 & 2.47 & 43.21 \\
Ovis2.5-2B$^{\#}$ & 55.65 & 1.60 & 2.56 & 45.64 & 51.70 & 1.44 & 2.43 & 42.49 & 47.96 & 1.33 & 2.35 & 40.13 & 53.33 & 1.51 & 2.49 & 43.86 \\
InternVL3.5-2B & 57.51 & 1.64 & 2.62 & 46.80 & 55.35 & 1.53 & 2.50 & 44.60 & 52.40 & 1.42 & 2.41 & 42.32 & 56.14 & 1.58 & 2.55 & 45.48 \\
InternVL3.5-8B & 57.85 & 1.65 & 2.62 & 47.00 & 55.77 & 1.56 & 2.55 & 45.30 & 53.30 & 1.51 & 2.51 & 43.99 & 56.57 & 1.60 & 2.58 & 46.04 \\
MiMo-VL-7B-RL-2508$^{\#}$ & 56.13 & 1.67 & 2.60 & 46.53 & 55.82 & 1.64 & 2.59 & 46.14 & 53.42 & 1.58 & 2.54 & 44.68 & 55.71 & 1.65 & 2.59 & 46.18 \\
MiniCPM-V-4.5 (8B)$^{\#}$ & 58.86 & 1.70 & 2.63 & 47.67 & 57.45 & 1.64 & 2.58 & 46.49 & 54.69 & 1.56 & 2.55 & 44.95 & 57.87 & 1.66 & 2.60 & 46.93 \\
Ovis2.5-9B$^{\#}$ & 59.51 & 1.74 & 2.72 & 48.85 & 57.84 & 1.65 & 2.63 & 47.01 & 53.83 & 1.55 & 2.50 & 44.32 & 58.26 & 1.68 & 2.66 & 47.66 \\
Qwen3-VL-8B-Instruct & \cellcolor{rank2green}61.34 & \cellcolor{rank2green}1.88 & \cellcolor{rank2green}2.84 & \cellcolor{rank2green}51.23 & 60.78 & \cellcolor{rank1green}1.85 & \cellcolor{rank1green}2.81 & \cellcolor{rank1green}50.64 & 56.83 & \cellcolor{rank1green}1.73 & \cellcolor{rank1green}2.71 & \cellcolor{rank2green}48.00 & 60.63 & \cellcolor{rank1green}1.85 & \cellcolor{rank1green}2.82 & \cellcolor{rank2green}50.65 \\ \midrule
\multicolumn{17}{c}{\textit{Open-source Thinking Multimodal Foundation Models}} \\ \midrule
Ovis2.5-2B$^*$ & 51.10 & 1.40 & 2.43 & 42.02 & 44.52 & 1.17 & 2.20 & 36.89 & 40.70 & 1.04 & 2.10 & 34.25 & 47.50 & 1.28 & 2.31 & 39.25 \\
Qwen3-VL-2B-Thinking & 59.23 & 1.47 & 2.63 & 46.13 & 57.02 & 1.19 & 2.53 & 42.79 & 48.82 & 1.02 & 2.30 & 37.81 & 57.26 & 1.32 & 2.56 & 43.97 \\
Ovis2.5-9B$^*$ & 55.92 & 1.61 & 2.59 & 45.95 & 53.79 & 1.51 & 2.50 & 44.03 & 49.71 & 1.39 & 2.41 & 41.48 & 54.44 & 1.55 & 2.54 & 44.74 \\
Keye-VL-1.5-8B$^*$ & 56.58 & 1.58 & 2.55 & 45.64 & 56.54 & 1.54 & 2.53 & 45.14 & 53.13 & 1.62 & 2.43 & 44.09 & 56.18 & 1.57 & 2.53 & 45.28 \\
MiMo-VL-7B-RL-2508$^*$ & 57.32 & 1.64 & 2.56 & 46.26 & 56.85 & 1.59 & 2.53 & 45.63 & 56.58 & 1.57 & 2.50 & 45.17 & 57.06 & 1.61 & 2.54 & 45.90 \\
MiniCPM-V-4.5 (8B)$^*$ & 58.87 & 1.74 & 2.67 & 48.25 & 57.79 & 1.69 & 2.63 & 47.33 & 54.16 & 1.54 & 2.52 & 44.46 & 57.95 & 1.70 & 2.64 & 47.49 \\
GLM-4.6V-Flash (9B) & 60.70 & 1.78 & 2.71 & 49.32 & 58.44 & 1.70 & 2.65 & 47.67 & 54.95 & 1.60 & 2.60 & 45.75 & 59.22 & 1.73 & 2.67 & 48.31 \\
Step3-VL-10B & 61.32 & 1.76 & 2.78 & 49.87 & 60.06 & 1.62 & 2.71 & 48.01 & \cellcolor{rank2green}59.11 & 1.60 & 2.64 & 47.09 & 60.61 & 1.69 & 2.74 & 48.87 \\
Qwen3-VL-8B-Thinking & \cellcolor{rank1green}63.77 & \cellcolor{rank1green}1.91 & \cellcolor{rank1green}2.85 & \cellcolor{rank1green}52.08 & \cellcolor{rank1green}62.61 & \cellcolor{rank2green}1.81 & \cellcolor{rank2green}2.78 & \cellcolor{rank2green}50.56 & 58.02 & 1.64 & 2.63 & 47.07 & \cellcolor{rank1green}62.70 & \cellcolor{rank2green}1.84 & \cellcolor{rank2green}2.80 & \cellcolor{rank1green}50.96 \\ \midrule
\multicolumn{17}{c}{\textit{Embodied/Egocentric Multimodal Models}} \\ \midrule
\rowcolor{highlightorange} Embodied-Reasoner (7B) & 39.74 & 1.00 & 1.96 & 32.65 & 39.69 & 0.99 & 1.96 & 32.56 & 37.22 & 0.97 & 2.02 & 32.23 & 39.44 & 1.00 & 1.97 & 32.57 \\
\rowcolor{highlightorange} EgoThinker (7B) & 48.22 & 1.36 & 2.23 & 39.39 & 45.88 & 1.28 & 2.19 & 37.89 & 42.42 & 1.16 & 2.08 & 35.36 & 46.71 & 1.31 & 2.20 & 38.38 \\
\bottomrule
\end{tabular}%
}
\end{table*}

\subsection{Main Results}
Based on the results reported in \cref{tab:eval_main}, we make the following key observations:

\noindent{\bf Human Performance.} 
Human participants achieve an $S_{uni}$ score of 59.08 on the  EXPLORE-Bench (tiny) \texttt{Full} set, outperforming the best model by 7.39. Humans show a clear strength on the \texttt{Short} and \texttt{Medium} subsets. However, on the \texttt{Long} subset, the performance gap between humans and the best model narrows, suggesting that MLLMs may have a relative advantage on tasks requiring to process long action sequences, particularly in predicting inter-object relations. Although human participants outperform the top-tier MLLMs, the absolute score is not very high, highlighting the challenge posed by our benchmark.

\noindent{\bf Proprietary MLLMs.} 
Despite a substantial gap to human performance, leading proprietary MLLMs, such as Gemini-3-Pro, deliver competitive results. In particular, the $S_{uni}$ score of Gemini-3-Pro on the \texttt{Long} subset is close to that of humans. Gemini-3-Flash performs very similarly to Gemini-3-Pro overall, and slightly better than GPT-5.2-Chat.

\noindent{\bf Open-source MLLMs.} 
Notably, Qwen3-VL-8B achieves higher overall performance than Gemini-3-Pro on our benchmark, although the latter is stronger on the \texttt{Long} subset. Qwen3-VL-8B-Instruct and Qwen3-VL-8B-Thinking perform comparably overall: the former scores higher on the \texttt{Medium} and \texttt{Long} subsets, while the latter performs better on the \texttt{Short} subset. However, most open-source models still lag far behind human performance, indicating substantial limitations in their egocentric scene prediction with long-horizon reasoning. Observations on thinking models are provided in \cref{sec:thinking_models}.

\noindent{\bf Embodied \& Egocentric MLLMs.} 
Although Embodied-Reasoner and EgoThinker exhibit strong embodied or egocentric reasoning capabilities after training on specific datasets, they perform worse than most general-purpose MLLMs on our newly proposed task. Notably, compared to their base model (Qwen2-VL-7B-Instruct), they do not exhibit a clear performance improvement on our benchmark (Embodied-Reasoner even performs worse). This suggests that egocentric scene prediction with long-horizon reasoning has been largely overlooked by the community and remains in need of further exploration and development.

\noindent{\bf Description Length.}
We compute the average length (in words) of the scene descriptions generated by each model (excluding chain-of-thought for thinking models) and find no clear correlation between description length and performance. The top three models—Qwen3-VL-8B-Thinking, Qwen3-VL-8B-Instruct, and Gemini-3-Pro—produce descriptions with average lengths of 545.77, 312.46, and 186.60 words, respectively, indicating that Gemini-3-Pro is both accurate and concise. Qwen2.5-VL-3B-Instruct generates descriptions with an average length of 305.58 words, which is close to Qwen3-VL-8B-Instruct, yet their scores differ substantially. Embodied-Reasoner and EgoThinker produce long descriptions (812.00 and 808.92 words on average) but deliver unsatisfactory performance. In contrast, Step3-VL-10B also generates relatively long descriptions (747.27 words on average) while achieving highly competitive results. Additionally, Ovis2.5-9B in thinking mode produces the shortest descriptions, averaging 121.69 words.

\section{How Well Stepwise Reasoning Performs?}
\label{sec:infer}

\subsection{Thinking Model}
\label{sec:thinking_models}
As shown in \cref{tab:eval_main}, Keye-VL-1.5-8B \cite{keyevl1_5} in thinking mode improves $S_{uni}$ by 5.77 on the full EXPLORE-Bench compared to its non-thinking counterpart. In addition, native thinking models such as GLM-4.6V-Flash \cite{glm4_6v}, Step3-VL-10B \cite{step3vl}, and Qwen3-VL-8B-Thinking \cite{qwen3vl} deliver strong results, representing the leading performance among open-source models on our benchmark. However, for same-sized open-source MLLMs from the same model family, thinking-mode variants do not always outperform non-thinking ones, as observed for Ovis2.5-2B \cite{ovis2_5}, Ovis2.5-9B, and MiMo-VL-7B-RL-2508 \cite{mimovl}.

\subsection{Non-thinking Model}
\label{sec:non_thinking_model}
As an open-source family of foundation models with strong multimodal general-purpose capabilities, Qwen3-VL \cite{qwen3vl} has been widely adopted both in industry \cite{chen2025mindgpt, shen2026evolving, li2026qwen3vl_emb} and academia \cite{li2025mobileworldbench, wang2025n3d, wang2025vl4gaze}. Within this family, the \texttt{Instruct} model (non-thinking) is often considered to offer greater development potential than the \texttt{Thinking} variant. Since non-thinking models cannot spontaneously reason with chain-of-thought as thinking models do, we elicit stepwise reasoning in Qwen3-VL-8B-Instruct through prompt design and explicit inference strategies.

\begin{figure}[t]
  \centering
  \begin{subfigure}{0.49\linewidth}
    \centering
    \includegraphics[width=\linewidth]{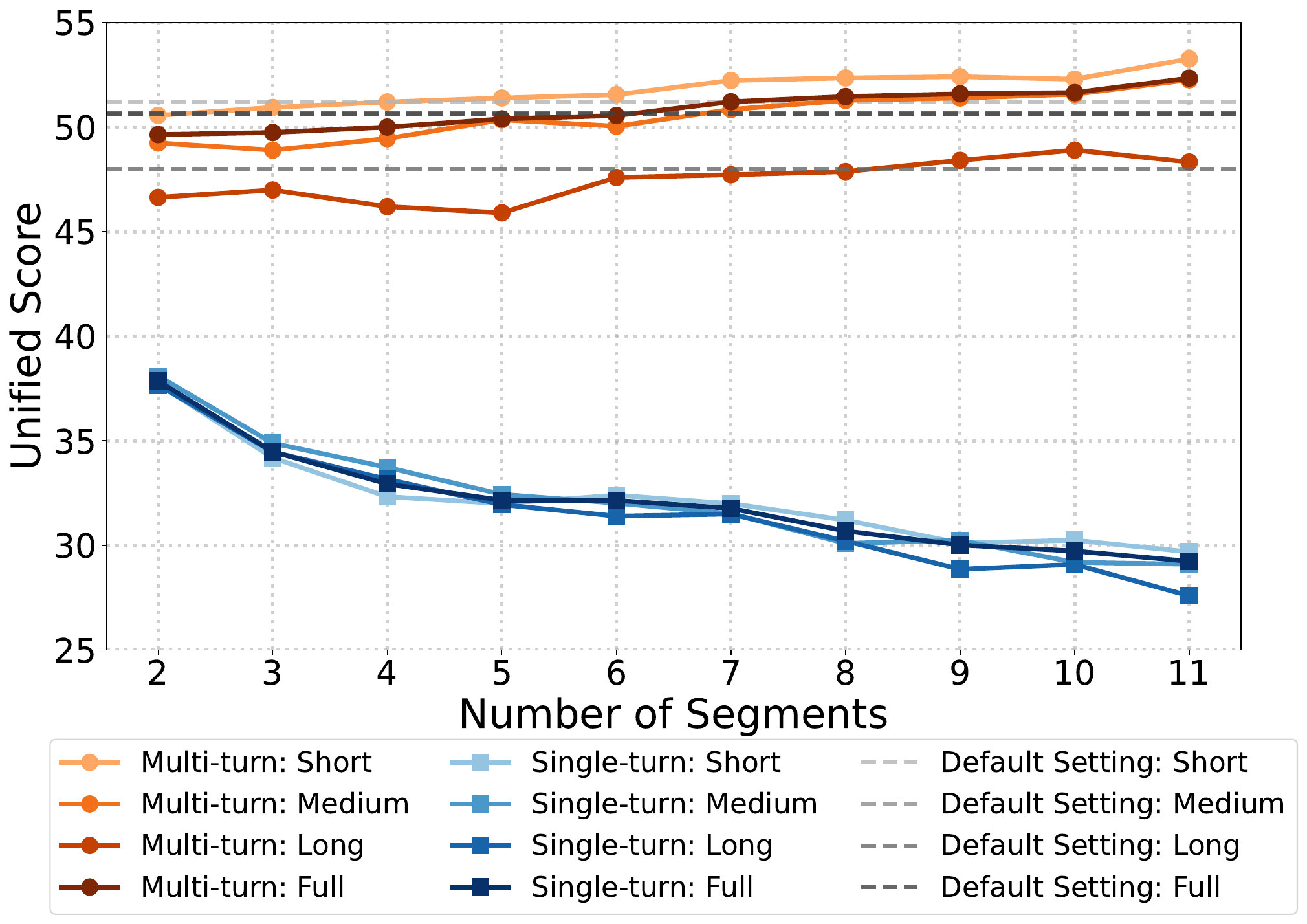} 
    \caption{\textbf{Unified score \vs number of segments.} Note that the two dashed lines corresponding to ``Default Setting: Medium'' and ``Default Setting: Full'' overlap.}
    \label{fig:score_segment}
  \end{subfigure}
  \hfill 
  \begin{subfigure}{0.49\linewidth}
    \centering
    \includegraphics[width=\linewidth]{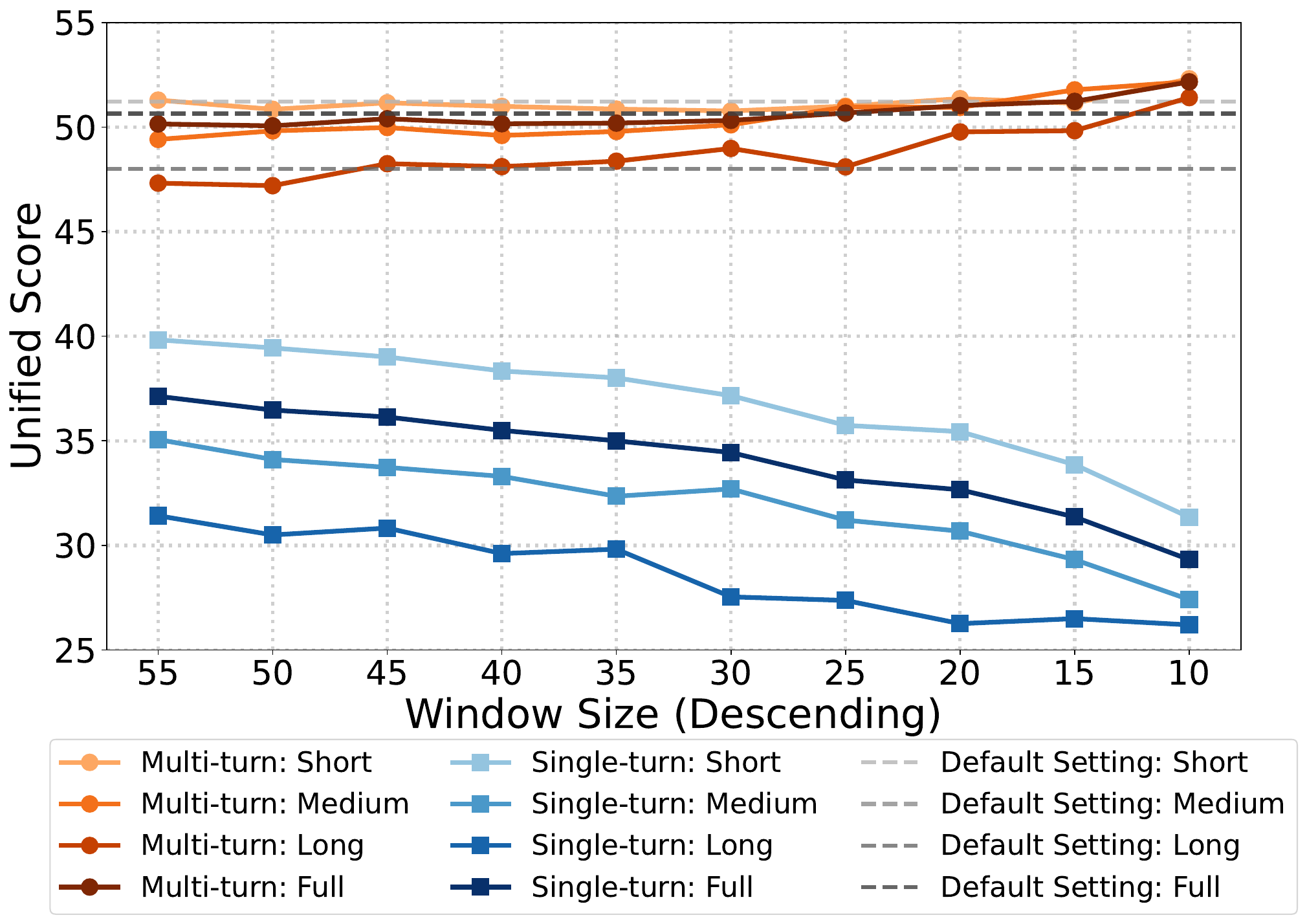}
    \caption{\textbf{Unified score \vs window size.} Note that the two dashed lines corresponding to ``Default Setting: Medium'' and ``Default Setting: Full'' overlap.}
    \label{fig:score_window}
  \end{subfigure}
  \caption{\textbf{Unified score $\bm{S_{uni}}$ of Qwen3-VL-8B-Instruct across subsets under different inference strategies.} \texttt{Short}, \texttt{Medium}, and \texttt{Long} denote the subsets with short, medium, and long atomic-action sequences. \texttt{Full} denotes the full dataset.}
  \label{fig:infer_score}
\end{figure}

\noindent{\bf Atomic Action Segments Partitioning.} 
To enable stepwise reasoning, we first segment the atomic action sequence in two ways. The first approach evenly partitions the action sequence of each instance into a predefined number of segments (\texttt{segment\_num}). This method is simple and straightforward, but does not account for variation in sequence lengths across instances. The second approach segments the sequence using a sliding window of size \texttt{window\_size}. Specifically, every \texttt{window\_size} atomic actions form one segment (the last segment may contain fewer actions), so longer sequences are divided into more segments. After segmenting the atomic action sequence, we apply two inference strategies to implement stepwise reasoning, which are introduced below. The experimental results of the above partitioning methods and inference strategies are provided in \cref{fig:infer_score}. 

\begin{figure}[t]
  \centering
  \begin{subfigure}{0.49\linewidth}
    \centering
    \includegraphics[width=\linewidth]{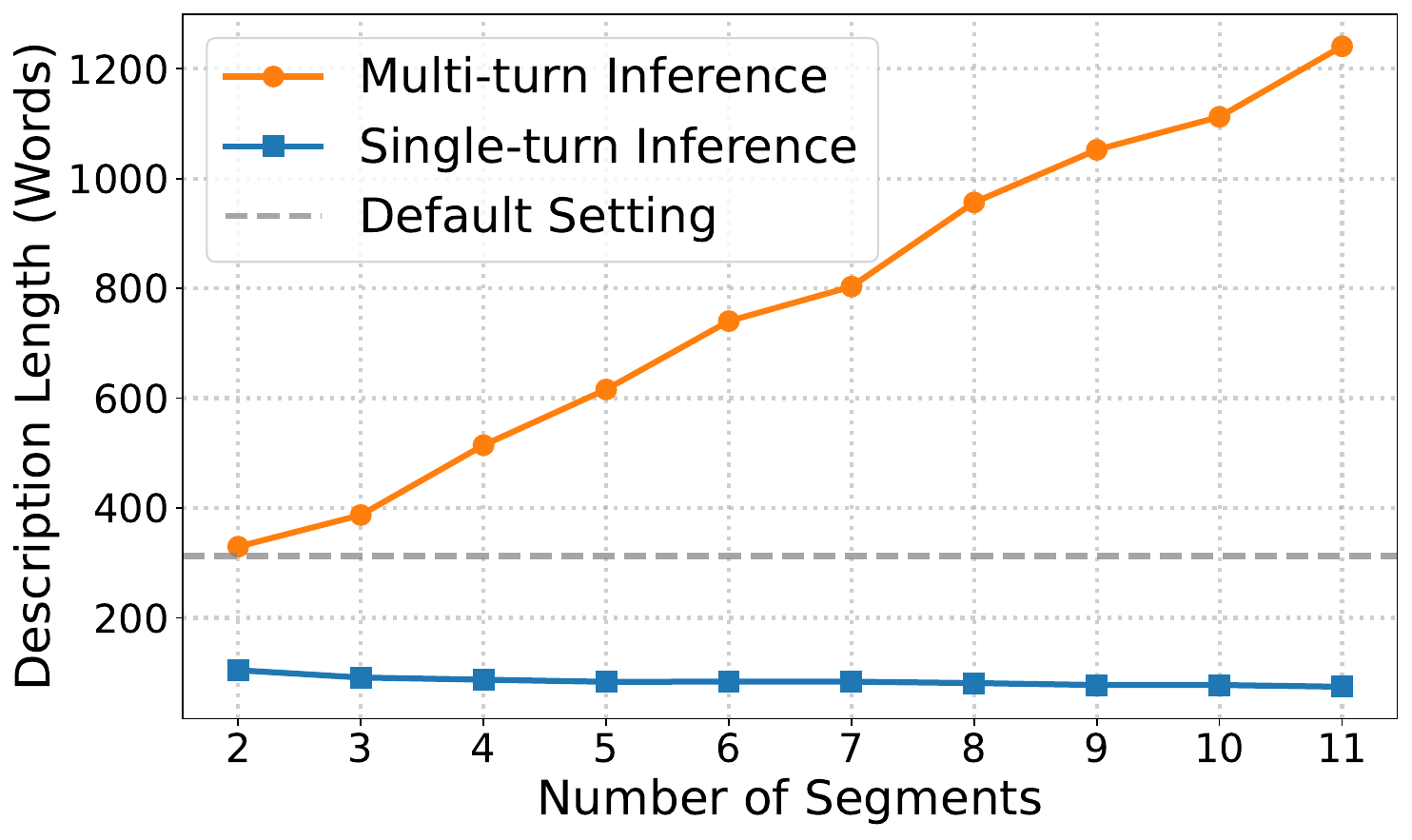} 
    \caption{Description length \vs number of segments}
    \label{fig:length_segment}
  \end{subfigure}
  \hfill 
  \begin{subfigure}{0.49\linewidth}
    \centering
    \includegraphics[width=\linewidth]{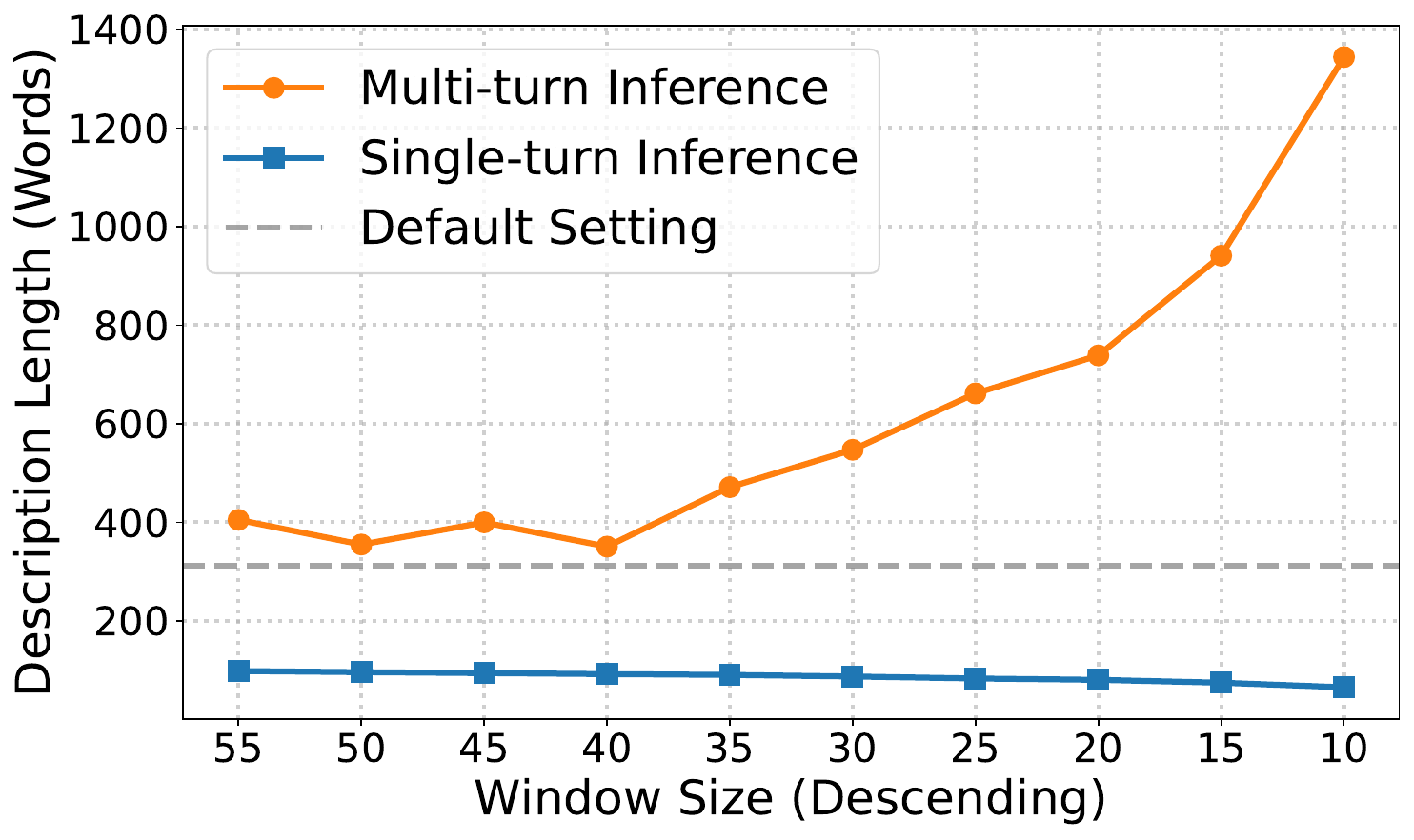}
    \caption{Description length \vs window size}
    \label{fig:length_window}
  \end{subfigure}
  \caption{\textbf{Average single-instance final-scene description length of Qwen3-VL-8B-Instruct under different inference strategies.}}
  \label{fig:infer_length}
\end{figure}

\begin{figure}[t]
  \centering
  \begin{subfigure}{0.49\linewidth}
    \centering
    \includegraphics[width=\linewidth]{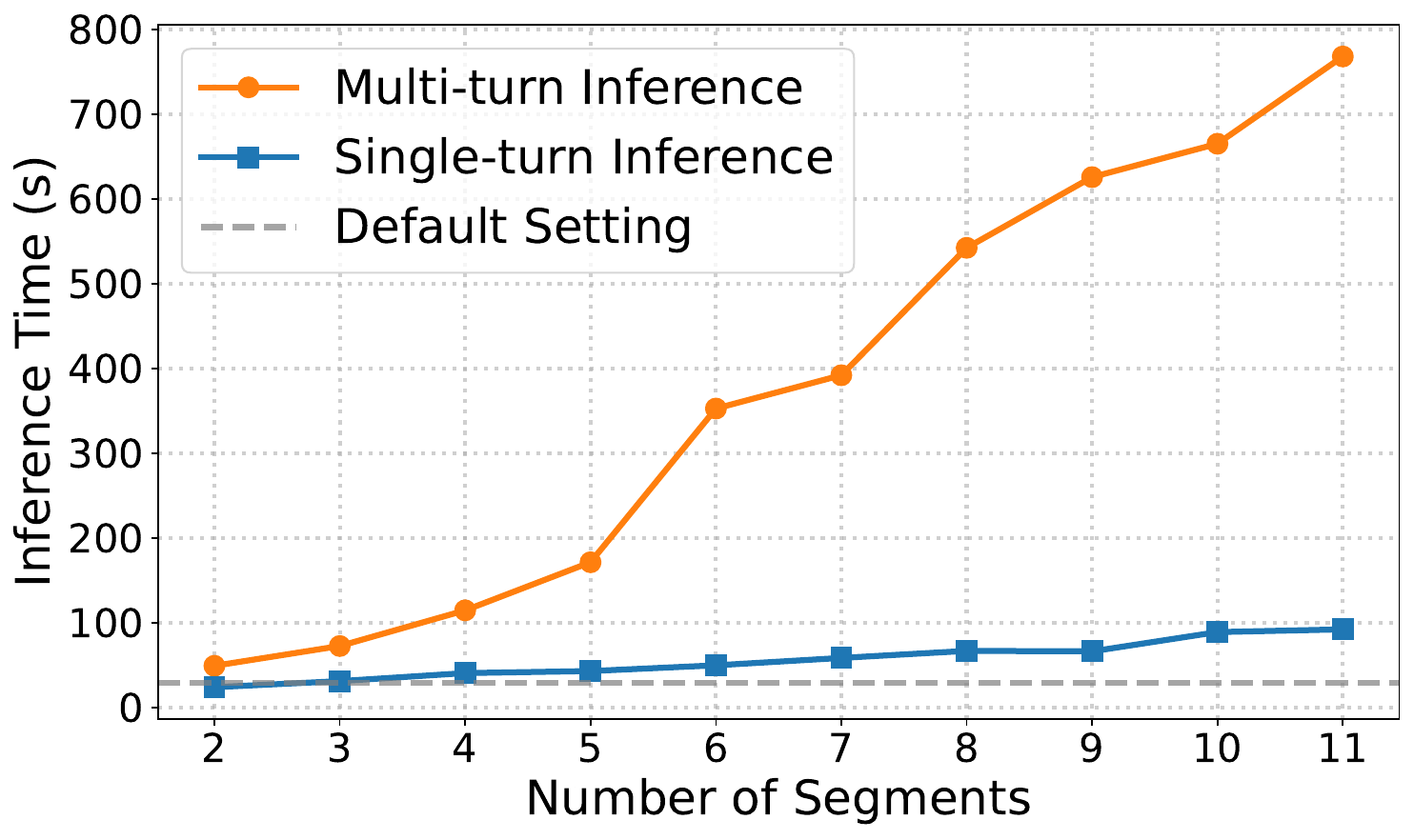} 
    \caption{Inference time \vs number of segments}
    \label{fig:time_segment}
  \end{subfigure}
  \hfill 
  \begin{subfigure}{0.49\linewidth}
    \centering
    \includegraphics[width=\linewidth]{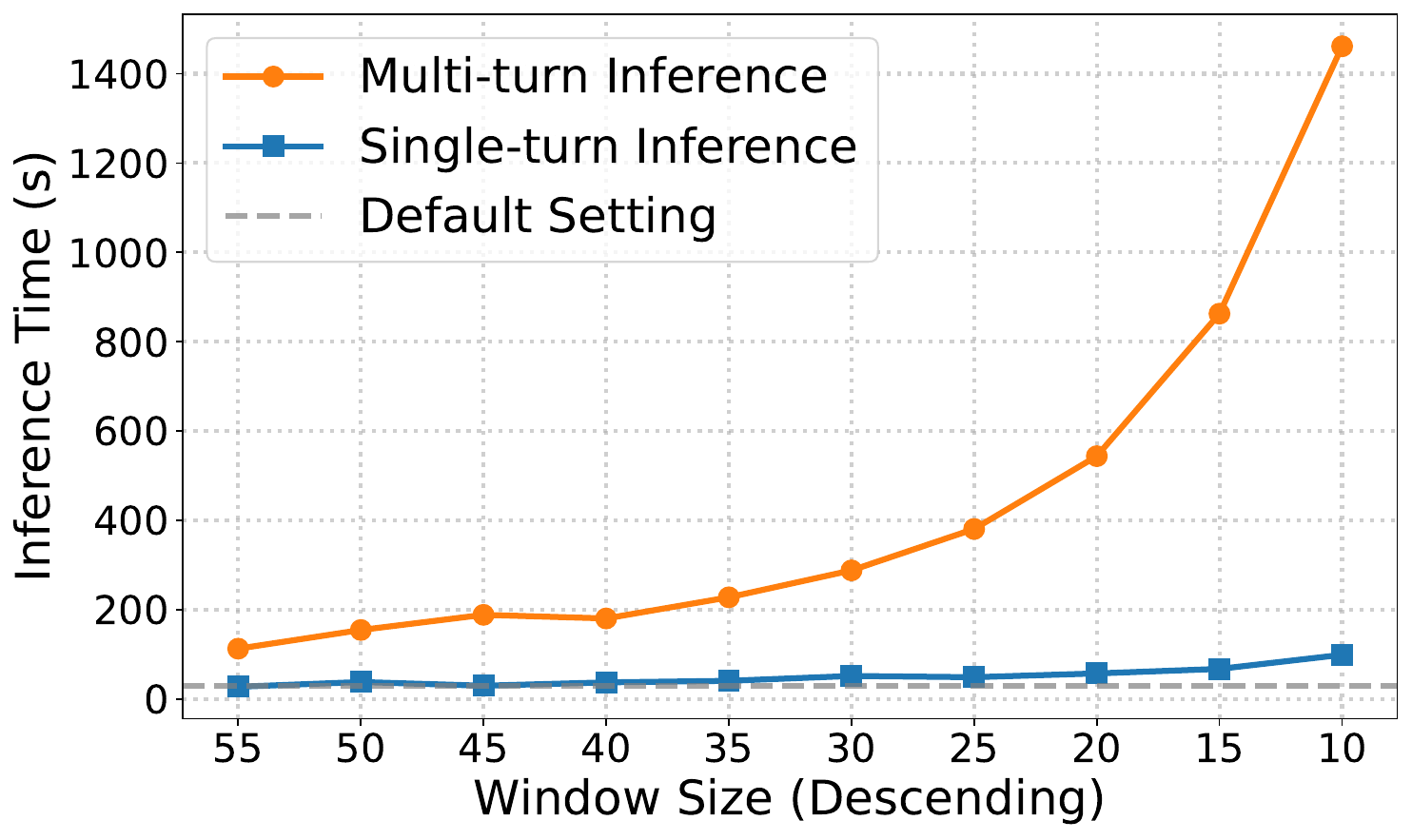}
    \caption{Inference time \vs window size}
    \label{fig:time_window}
  \end{subfigure}
  \caption{\textbf{Average single-instance inference time of Qwen3-VL-8B-Instruct on an NVIDIA H200 GPU under different inference strategies.}}
  \label{fig:infer_time}
\end{figure}

\noindent{\bf Single-turn Inference.} 
Inspired by the dynamic programming \cite{bellman1966dynamic} principle of decomposing the original problem into subproblems and performing state transitions, we prompt Qwen3-VL-8B-Instruct to predict the scene after executing each action segment in a single-turn inference, while conditioning each prediction on the scene description after the previous segment (if available) and the initial-scene image. 

We evaluate the resulting final-scene descriptions using the protocol in \cref{sec:eval_proc}. However, this strategy does not yield strong gains. Instead, compared to the default setting that directly prompts the model to predict the final scene, it leads to a substantial performance drop, which becomes more severe when \texttt{segment\_num} is larger or \texttt{window\_size} is smaller (see \cref{fig:infer_score}). 
We find through case study that under this inference strategy, the model’s generated scene descriptions mainly focus on the changes across scenes and tend to ignore the parts that remain unchanged. As a result, compared with the default setting, the final scene descriptions produced with this strategy are much shorter (see \cref{fig:infer_length}).

\noindent{\bf Multi-turn Inference.} 
To encourage the model to generate detailed and complete scene descriptions after executing each action segment, we propose a new inference strategy. This strategy follows the same underlying idea as the single-turn strategy above, but implements it via multi-turn inference. The key difference from the above strategy is that, in each inference round, the model is exposed to only one action segment and is asked to output the scene description after executing that segment, analogous to the final-scene prediction. As the number of rounds increases, the generated description becomes significantly longer (see \cref{fig:infer_length}).

As shown in \cref{fig:infer_score}, in contrast to the single-turn strategy, the multi-turn inference yields better performance when \texttt{segment\_num} is larger or \texttt{window\_size} is smaller. This suggests that a finer decomposition into subtasks benefits scene prediction with long-horizon reasoning, aligning with intuition. We also find that a large \texttt{segment\_num} (\eg, 11) and a small \texttt{window\_size} (\eg, 10) achieve similar $S_{uni}$ scores on the full dataset (52.34 \vs 52.16); however, the former performs better on the \texttt{Short} subset (53.26 \vs 52.31), whereas the latter is noticeably stronger on the \texttt{Long} subset (48.33 \vs 51.41). Additionally, multi-turn inference with \texttt{window\_size=10} improves $S_{uni}$ by 3.41 on the \texttt{Long} subset compared to the default setting. These observations suggest that adaptively segmenting action sequences based on their length may offer advantages for long-horizon reasoning tasks.
However, although larger \texttt{segment\_num} and smaller \texttt{window\_size} can improve performance over the default setting, model performance drops below the default when \texttt{segment\_num} is less than 7 or \texttt{window\_size} exceeds 25. Moreover, multi-turn inference incurs multiplicative computational overhead (see \cref{fig:infer_time}) while providing relatively limited gains.

\begin{table*}[t]
\caption{\textbf{Evaluation results of abnormal cases.} \colorbox{rank1green}{Dark green} and \colorbox{rank2green}{light green} indicate the best and the second best result among all models.}
\label{tab:eval_abnormal}
\centering
\setlength{\tabcolsep}{3pt} 

\begin{tabular}{l|ccc >{\columncolor{lightgraybg}}c c >{\columncolor{lightgraybg}}c}
\toprule
\textbf{Methods} & $\bm{S_{obj}}$ & $\bm{S_{att}}$ & $\bm{S_{rel}}$  & $\bm{S_{uni}}$ & $\bm{S_{key}}$  & $\bm{S_{abn}}$ \\
\midrule
\rowcolor{lightcyanbg} Human & 95.19 & 3.64 & 3.88 & 80.32 & 4.65 & 91.64 \\
Embodied-Reasoner & 63.75 & 1.62 & 2.75 & 49.28 & 1.45 & 30.95 \\
Qwen2-VL-7B-Instruct & 68.84 & 2.13 & 2.98 & 55.98 & 2.20 & 45.20 \\
EgoThinker & 72.33 & 2.21 & 2.92 & 56.92 & 2.38 & 48.44 \\
Qwen3-VL-8B-Instruct & 83.83 & 2.82 & 3.51 & 68.75 & 2.54 & 52.63 \\ 
Gemini-3-Pro & 80.98 & 2.70 & 3.41 & 66.47 & 2.74 & 56.00 \\
Qwen3-VL-8B-Thinking & \cellcolor{rank2green}86.03 & \cellcolor{rank1green}2.89 & \cellcolor{rank1green}3.62 & \cellcolor{rank1green}70.71 & \cellcolor{rank2green}2.84 & \cellcolor{rank2green}58.22 \\
GPT-5.2-Chat & \cellcolor{rank1green}87.32 & \cellcolor{rank2green}2.82 & \cellcolor{rank2green}3.55 & \cellcolor{rank2green}69.94 & \cellcolor{rank1green}3.10 & \cellcolor{rank1green}62.79 \\
\bottomrule
\end{tabular}
\end{table*}

\section{Can MLLMs Describe Abnormal States?}
\label{sec:abnormal}
\begin{figure}[t]
  \centering
  \includegraphics[width=\linewidth]{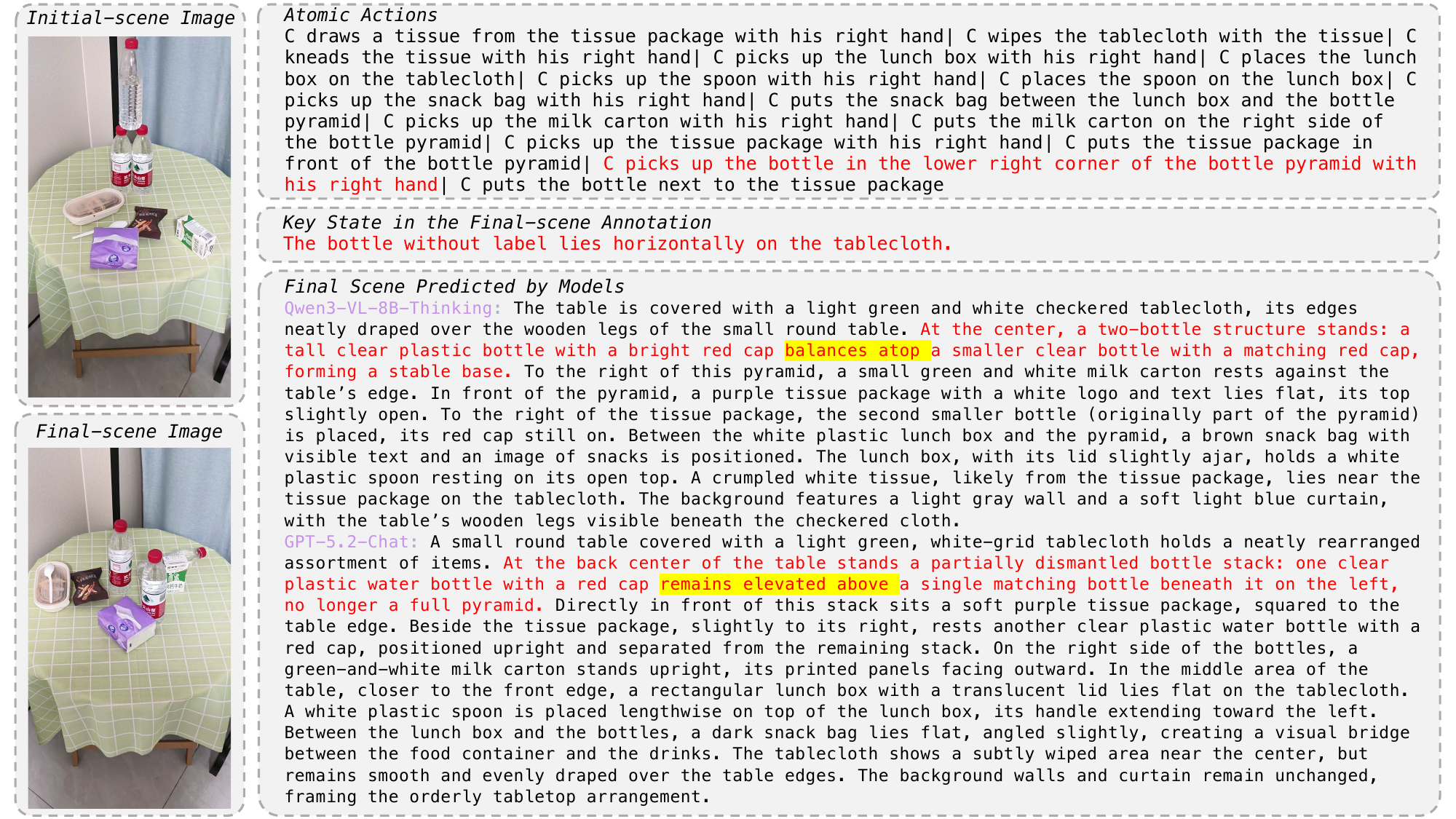}
  \caption{\textbf{Case study of an abnormal situation.} ``C'' refers to the camera wearer. Although both models recognize the changes, neither accurately describes the key state. Instead, they generate descriptions that \hl{violate physical commonsense}.}
  \label{fig:abnormal_case_1v2}
\end{figure}

We record and annotate 20 videos in which executing a sequence of actions leads to abnormal states in the final scene. The anomalies in these videos fall into two main categories: environmental damage (\eg, objects falling) and safety hazards (\eg, a faucet left running). In addition to annotating atomic actions and the objects, attributes, and relations in the final scene, we also annotate the key object states that reflect the scene anomalies (as shown in \cref{fig:abnormal_case_1v2}), enabling us to investigate whether MLLMs can infer the anomalies and describe these key states. We evaluate descriptions of key states in the same way as we evaluate object attributes, yielding a key-state score $S_{key}$ (0-5) . We then compute the score for abnormal-scene descriptions as
\begin{equation}
  S_{abn}=(1-\lambda)S_{uni}+\lambda(20 \cdot S_{key}).
  \label{eq:s_abn}
\end{equation}
We report results of these cases with $\lambda=0.9$ in \cref{tab:eval_abnormal}.

Compared with the full EXPLORE-Bench, these cases have much shorter action sequences (17 on average) and involve fewer objects (9 on average) per scene, leading to relatively higher $S_{uni}$ scores for the models. However, their $S_{abn}$ scores still lag far behind human performance. Benefiting from everyday experience, humans can easily detect abnormal states and achieve an $S_{key}$ score of 4.65. Relative to the base model (Qwen2-VL-7B-Instruct \cite{qwen2vl}), Embodied-Reasoner \cite{zhang2025embodied} performs worse overall, while EgoThinker \cite{pei2025egothinker} shows only marginal improvements. Qwen3-VL-8B-Thinking \cite{qwen3vl} and GPT-5.2-Chat \cite{gpt5_2} perform well among all models, yet our case study (shown in \cref{fig:abnormal_case_1v2}) demonstrates that they still fail to accurately describe abnormal states that humans can readily perceive.

\section{Conclusion}
\noindent In this paper, we introduce \textbf{EXPLORE-Bench}, a benchmark for \emph{egocentric scene prediction with long-horizon reasoning}, where models are asked to anticipate the final state of a scene given an initial image and a sequence of atomic actions. 
EXPLORE-Bench is built from real first-person videos across diverse scenarios and provides structured final-scene annotations---object categories, attributes, and relations---enabling fine-grained, quantitative evaluation of holistic scene-level prediction.
Through extensive evaluation of both proprietary and open-source MLLMs, we find that current models still struggle to reliably track long-term action consequences and remain substantially behind human performance, especially in abnormal cases. We further investigated stepwise reasoning at test time, showing that decomposing long action sequences can improve performance in challenging long-horizon settings, while also revealing non-trivial trade-offs between accuracy gains and computational overhead.

\noindent \textbf{Limitations and Future Work.} Currently, our exploration of test-time scaling is limited to a small set of inference strategies. Developing more effective \emph{and} more efficient test-time scaling methods for long-horizon reasoning remains open. 
While EXPLORE-Bench covers diverse everyday activities, rare or abnormal cases (\eg, unexpected interactions) are still under-represented. Collecting and annotating more such instances is an important next step. 
Moreover, as more data become available, it would be valuable to construct a dedicated training set to further advance models' long-horizon egocentric reasoning.

\section*{Acknowledgements}
This work is supported by the National Natural Science Foundation of China (NSFC) under Grants 62225207, 62436008, 62306295, 62576328 and 625B2175.
We sincerely thank the Foundation Model Team at Li Auto Inc. for providing computational resources and data support.

%
%
\bibliographystyle{splncs04}
\bibliography{main}

\clearpage

\appendix
\renewcommand{\theHsection}{appendix.\thesection}
\renewcommand{\theHsubsection}{appendix.\thesubsection}

\section{Appendix Outline}
\label{sec:outline} 

In this appendix, we provide:

\begin{itemize}
\item The prompts used in our work (\cref{sec:prompts}).
\item More details on quality control (\cref{sec:quality}).
\item Detailed evaluation setups (\cref{sec:setups}).
\item More experimental results (\cref{sec:exp_res}).
\item More cases and case studies (\cref{sec:cases}).
\item Comparisons with related works (\cref{sec:work}).
\item Broader impacts and limitations (\cref{sec:discussion}).
\end{itemize}

\section{Prompts}
\label{sec:prompts}
In this section, we provide our prompts used for annotation (\cref{sec:prompts4anno}), inference (\cref{sec:prompts4infer}), and evaluation (\cref{sec:prompts4eval}), as listed below:

\subsection{Prompts for Annotation}
\label{sec:prompts4anno}
The following are the prompts we use in our scene annotation pipeline.

\begin{lstlisting}[style=prompt, caption={The prompt for tag filtering}, label={lst:prompt4tag_filter}]
You are given a Python list of English words or short phrases. Your task is to clean and filter this list and then output a new Python list.

Follow these rules:

1. **Remove verbs**  
   - Delete all items that are verbs (actions), e.g., `"stir"`, `"attach"`.

2. **Keep only object-like nouns**  
   - Keep words/phrases that refer to concrete, physical objects or places.  
   - Remove items that are not objects, such as abstract nouns (e.g., `"job"`) or actions/events.  
   - Remove items that refer to human body parts (e.g., `"hand"`, `"arm"`, `"leg"`).

3. **Handle multiple words referring to people**  
   - If there are several words that all mean a generic human (e.g., `"cook"`, `"man"`, `"person"`), keep only `"person"` and remove the others.  
      - If `"person"` is not present, keep just one most generic word that refers to a human and remove the rest.

4. **Handle generic vs. specific objects**  
   - When there is a generic term and one or more more-specific terms of the same type, **keep only the most specific term**.  
     - Example: `"home appliance"` and `"oven"` -> keep `"oven"`, remove `"home appliance"`.  
     - Example: `"food"` and `"omelet"` -> keep `"omelet"`, remove `"food"`. Only keep `"food"` if there is no more specific food item in the list.

5. **Handle near-synonyms or overlapping objects**  
   - If two or more nouns/phrases refer to almost the same physical object, keep only one of them.  
   - If the difference is just an modifier (e.g., fuel type), keep the **longer, modified noun**.
     - Example: `"frying pan"` and `"pan"` -> keep `"frying pan"`.  Only keep `"pan"` if there is no word like `"frying pan"` in the list.
     - Example: `"gas stove"` and `"stove"` -> keep `"gas stove"`. Only keep `"stove"` if there is no word like `"gas stove"` in the list.

6. **Handle multiple places/locations**  
   - If the list contains several place words (e.g., `"basement"`, `"garage"`, `"park"`, `"room"`), choose **one** place that best matches the context implied by the other words in the list (for example, `"garage"` if there are tools and a bicycle), and remove the others.

7. **Near-synonyms in general**  
   - For any pair or group of near-synonyms (very similar meaning), keep only one term and delete the others.
     - Example: `"omelet"` and `"pancake"` -> keep `"omelet"`. 
     - Example: `"oven"` and `"microwave"` -> keep `"oven"`. 

8. **Output format**  
   - Return only the final, filtered list as a valid Python list of strings, in the same language and casing as the input items.  
   - Do not include explanations or any extra text-only the resulting list.

**Input list:**
{tags}
\end{lstlisting}

\begin{lstlisting}[style=prompt, caption={The prompt for attribute generation}, label={lst:prompt4attribute_gen}]
You are given an image and a set of detected objects in that image.  
Each object has:
- an object id: category name with its instance index (e.g., "table.0", "cup.1"), and
- a bounding box in the format [x_min, y_min, x_max, y_max].

The original width and height of the image are {img_wh}.  
The detected objects are: {obj_list}.

Your task:  
Based on the visual content inside each provided bounding box, describe the visual attributes of each object (e.g., shape, size, color, material/texture, pattern, state/condition). Produce phrases or 1 sentence per object.

Requirements:
- Output ONLY a valid JSON array (no markdown, no extra text). Use double quotes for all JSON strings. Do not use single quotes. Do not include trailing commas.
- The output must be a list where each element is a JSON object with exactly ONE key-value pair:
  {{ "object_id": "visual attribute description" }}
- object_id must come from the given objects and must use the instance ids exactly (e.g., "table.0").
- Descriptions should be based strictly on what is visible in the image region within the bbox; do not guess invisible attributes.
- Include attributes when visible, such as:
  - color(s) (e.g., "red", "black and white", "transparent")
  - shape/form (e.g., "round", "rectangular", "cylindrical", "flat")
  - relative size in the image (e.g., "small", "medium", "large"; or "thin", "wide", "tall")
  - material/texture (e.g., "metal", "wooden", "plastic", "glass", "leather", "furry", "smooth", "rough")
  - pattern/markings (e.g., "striped", "polka-dotted", "logo", "text")
  - condition/state (e.g., "open", "closed", "full", "empty", "broken", "dirty", "flat") when clearly visible
- Do **not** include any spatial or interactive relations between objects (e.g., "attached to tire", "held by person").

Return JSON in this exact example shape:
[
  {{"table.0": "large rectangular wooden table with a flat surface, light brown finish"}},
  {{"cup.1": "small white ceramic cup with a handle, glossy surface"}}
]
\end{lstlisting}

\begin{lstlisting}[style=prompt, caption={The prompt for relation generation}, label={lst:prompt4relation_gen}]
You are given an image and a set of detected objects in that image.
Each object has:
- a category name with its instance index (e.g., "person.0", "tire.1", "cup.2"), and
- a bounding box in the format [x_min, y_min, x_max, y_max].

The original width and height of the image are {img_wh}.
The detected objects are: {obj_list}.

Your task:
Based on the visual content of the image and the provided objects, produce the scene graph relationships.

Requirements:
- Output ONLY a valid JSON array (no markdown, no extra text).
- Each relationship must be a JSON array of 3 strings:
  ["subject_id", "relation", "object_id"]
- subject_id/object_id must come from the given objects and must use the instance ids (e.g., "person.0").
- relation should be a meaningful predicate (spatial/interaction/semantic), e.g.:
  "on", "under", "in front of", "behind", "next to", "near", "inside", "overlapping",
  "holding", "riding", "walking on", "sitting on", "standing on", "looking at", "wearing"
- Include as many correct and relevant relationships as you can see from the image.

Return JSON in this exact example shape:
[
  ["person.0", "holding", "tire.1"],
  ["cup.2", "on", "table.3"]
]
\end{lstlisting}

\begin{lstlisting}[style=prompt, caption={The prompt for annotation correction.}, label={lst:prompt4anno_correct}]
You are an expert visual scene analyzer and verifier.

You are given:
1) an input image  
2) an initial JSON list `info_list` describing objects in the image (it may contain mistakes and may be incomplete).

### Your task
Using the image as the ground truth, **correct and enhance** `info_list` and output a **final JSON list**.

### What you must do
1. **Verify and correct each listed object**
   - For every dict in `info_list`, check the object dict against the image.
   - You may correct and improve the object's `"description"`, and `"relation"` fields to best match the image.
   - If the object is not in the image at all, delete the object dict after **double checking**.
   - If the object's `"category"` and `"description"` have nothing to do with each other, remove the object dict from the provided list.
   - Remove the object dict whose `"category"` is human body parts (e.g., `"hand"`, `"arm"`, `"leg"`, `"waist"`).

2. **Add missing objects**
   - Detect **all reasonably identifiable objects** that appear in the image but are missing from `info_list`.
   - Add each missing object as a new dict entry in the same format.
   - If there are multiple instances of the same category (e.g., two chairs), create **separate entries**.

3. **Improve completeness of relations**
   - Ensure relations cover meaningful spatial/interaction links between objects (e.g., on, under, inside, next to, in front of, behind, holding, touching, wearing).
   - Add relations so that important object-to-object relationships are captured across the whole list.
   - Relations must be **short lowercase phrases** in the style:
     - `"cat lying on desk"`
     - `"keyboard on desk"`
     - `"tire next to car"`
   - Each relation string should use the current object as the **subject**.

### Overall requirements

1. **Exhaustive coverage of objects**  
   - Try to detect and list **as many distinct objects as possible** in the image.  
   - If multiple objects belong to the **same category but are different instances** (e.g., two cats, three chairs), you must create **separate dict entries** for each instance with distinctive `"description"`.

2. **Output format**

Return **only valid JSON**, in the following format (no extra explanation, no comments):

```json
[
  {
    "category": "tire",
    "description": "black rubber tire with deep treads, round and thick, appearing durable and designed for rugged terrain",
    "relation": [
      "tire on ground",
      "tire next to car"
    ]
  },
  {
    "category": "cat",
    "description": "orange and black cat with short fur, medium-sized body, triangular ears, and a long tail",
    "relation": [
      "cat lying on desk",
      "cat next to keyboard"
    ]
  },
  {
    "category": "keyboard",
    "description": "black rectangular keyboard with white printed keys, flat profile, and a wired connection",
    "relation": [
      "keyboard on desk"
    ]
  }
]
```

This example is **only a template**. For the actual image, adjust:
- the number of objects (list length),
- the categories,
- the descriptions,
- the relations.

### Field definitions

For **each object** in the image, create one dict with:

1. `category` (string)  
   - A **concise category name** of the object (e.g., `"person"`, `"cat"`, `"cup"`, `"car"`, `"tree"`, `"keyboard"`, `"lamp"`).  
   - Use lowercase English nouns, singular form.
   - Do not use broad category names like `"red object"`.

2. `description` (string)  
   - Write as a **detailed attribute phrase** like:  
     `"black rubber tire with deep treads, round and thick, appearing durable and designed for rugged terrain"`  
   - Descriptions should be based strictly on what is visible in the image; do not guess invisible attributes.
   - Include attributes when visible, such as:
     - color(s) (e.g., "red", "black and white", "transparent")
     - shape/form (e.g., "round", "rectangular", "cylindrical", "flat")
     - relative size in the image (e.g., "small", "medium", "large"; or "thin", "wide", "tall")
     - material/texture (e.g., "metal", "wooden", "plastic", "glass", "leather", "furry", "smooth", "rough")
     - pattern/markings (e.g., "striped", "polka-dotted", "logo", "text")
     - condition/state (e.g., "open", "closed", "full", "empty", "broken", "dirty", "flat") when clearly visible  
   - **Do not** include any relations (no "on/next to/in front of..."), actions, or interactions.

3. `relation` (list of strings)  
   - Write each relation as a **short, lowercased phrase without "The ... is ..."**, like:  
     - `"cat lying on desk"`  
     - `"keyboard on desk"`  
     - `"cup next to plate"`  
   - Format: **`subject + relation + object`** (or **`subject + relation + environment`**), keeping it brief and consistent.  
   - Include spatial and interaction relations when meaningful (on, under, inside, behind, next to, holding, touching, etc.).  
   - Ensure the **subject is this object** (the object of the current dict).

### Additional guidelines

- Try to **cover all visible objects** that are reasonably identifiable.  
- If multiple objects share the same category, distinguish them via descriptions and relations (do not use ids like cat1/cat2).  
- Do **not** include any commentary outside the JSON.  
- Do **not** include fields other than `category`, `description`, and `relation`.  
- Ensure the output is **valid JSON** (proper quotation, commas, brackets).

### Input
Initial `info_list` to correct and expand:
{info_list}

Now analyze the image, correct/improve the existing objects, delete invisible objects, add missing objects, and output the final corrected JSON list.
\end{lstlisting}

\subsection{Prompts for Inference}
\label{sec:prompts4infer}
Note that the prompt used in the first round of multi-turn inference is the same as the default prompt for one-go final-scene prediction.

\begin{lstlisting}[style=prompt, caption={The prompt for one-go final-scene prediction (default setting)}, label={lst:prompt4one_go}]
You are given an image showing a initial scene.
Then, a sequence of atomic actions occurs:
'{atomic_actions}'

Notes:
- 'C' refers to the camera wearer (the person who is wearing or holding the camera).
- 'X' refers to someone other than the camera wearer.
- Each atomic action description is separated by the `|` character.
- The atomic actions happen in the given order, from left to right.

Your task:
1. Mentally apply this sequence of atomic actions to the initial scene in the image.
2. Imagine what the final scene would look like after all actions have occurred.
3. Describe the final scene in rich, concrete visual detail, as if you are describing the final image.

Requirements:
- Focus **only** on the final scene after all actions, not the intermediate steps.
- Do **not** explain your reasoning or mention the actions explicitly.
- Do **not** mention that you are imagining or predicting; just describe the final scene directly.
- Provide rich, concrete visual detail: entities, appearances, spatial relationships, and interactions.
- Avoid any meta-commentary (no phrases like "the image would show", "I imagine that", etc.).

Now, describe the final scene in detail.
\end{lstlisting}

\begin{lstlisting}[style=prompt, caption={The prompt for stepwise single-turn inference}, label={lst:prompt4single_turn}]
You are given an image showing an initial scene.
Then a sequence of atomic actions occurs, provided in {segment_num} consecutive segments (Segment 1, Segment 2, ..., Segment {segment_num}). The full sequence is continuous in time, and the segments are in chronological order.

Atomic action segments:
{atomic_action_segments}

Notes:
- 'C' refers to the camera wearer (the person wearing/holding the camera).
- 'X' refers to someone other than the camera wearer.
- Within each segment, atomic actions are separated by the `|` character.
- Segments happen in order from Segment 1 to Segment {segment_num}. All actions within a segment happen before the next segment begins.

Your task:
- Starting from the initial scene in the image, mentally apply Segment 1's actions and describe the resulting scene (Scene 1).
- Then, using your previously described Scene 1 as the current state (grounded by the initial image), mentally apply Segment 2's actions and describe the resulting scene (Scene 2).
- Continue this process until Segment {segment_num}, producing Scene {segment_num}.
- Describe the scenes in rich, concrete visual detail, as if you are describing the images.

Description requirements for each scene:
- Describe **only** the scene after that segment's actions are complete (do not describe intermediate steps).
- Do **not** explain your reasoning or mention the actions explicitly.
- Do **not** mention that you are imagining or predicting; describe the scene directly.
- Provide rich, concrete visual detail: entities, appearances, spatial relationships, and interactions.
- Avoid meta-commentary (no phrases like "the image would show", "I imagine that", etc.).

Output format (strict):
- Output a valid JSON list with exactly {segment_num} elements (no nested lists, no extra keys, no extra text outside the list).
- The i-th element must be the scene description after completing Segment i.
- Each element must be a single string enclosed in double quotes.

Now output the list of {segment_num} scene descriptions.
\end{lstlisting}

\begin{lstlisting}[style=prompt, caption={The prompt for stepwise multi-turn inference}, label={lst:prompt4multi_turn}]
You are given an image showing the previous initial scene.

You are also given the current scene state description (after previous actions):
"{previous_scene}"

Then, the following atomic actions occur next:
'{atomic_actions}'

Notes:
- 'C' refers to the camera wearer (the person who is wearing or holding the camera).
- 'X' refers to someone other than the camera wearer.
- Each atomic action description is separated by the `|` character.
- The atomic actions happen in the given order, from left to right.

Your task:
1. Treat the provided current scene description as the starting state (grounded by the initial image), mentally apply the new actions.
2. Imagine what the resulting scene would look like after all actions have occurred.
3. Describe the resulting scene in rich, concrete visual detail, as if you are describing the final image.

Requirements:
- Describe **only** the resulting scene after all actions (no intermediate steps).
- Do **not** explain your reasoning or mention the actions explicitly.
- Do **not** mention that you are imagining or predicting; describe the scene directly.
- Provide rich, concrete visual detail: entities, appearances, spatial relationships, and interactions.
- Avoid meta-commentary (no phrases like "the image would show", "I imagine that", etc.).

Now write the resulting scene description as plain text in detail.
\end{lstlisting}

\subsection{Prompts for Evaluation}
\label{sec:prompts4eval}
We use the same system prompt but different user prompts during evaluation.

\begin{lstlisting}[style=prompt, caption={The system prompt for evaluation}, label={lst:prompt4evaluation}]
You are a Natural Language Processing (NLP) expert. A curious human will give you a sentence and a phrase. The human want you to help with analyzing whether the sentence includes similar concept with the given phrase and rate the similarity on a scale from 0 to 5, with 0 being 'completely lacks similar concepts' and 5 being 'extremely has similar concepts'. You need to give helpful and reliable answers to help the human.
\end{lstlisting}

\begin{lstlisting}[style=prompt, caption={The prompt for attribute scoring}, label={lst:prompt4s_att}]
Sentence: '{sentence}'. Phrase: '{phrase}'. Please check if the sentence contains parts that accurately describe the attributes referenced in the provided phrase and provide an integer score as a single number from 0 to 5 without explanation.
\end{lstlisting}

\begin{lstlisting}[style=prompt, caption={The prompt for relation scoring}, label={lst:prompt4s_rel}]
Sentence: '{sentence}'. Phrase: '{phrase}'. Please check if the sentence contains parts that accurately describe the relations referenced in the provided phrase and provide an integer score as a single number from 0 to 5 without explanation.
\end{lstlisting}

\begin{lstlisting}[style=prompt, caption={The prompt for key-state scoring}, label={lst:prompt4s_key}]
Sentence: '{sentence}'. Phrase: '{phrase}'. Please check if the sentence contains parts that accurately describe the states referenced in the provided phrase and provide an integer score as a single number from 0 to 5 without explanation.
\end{lstlisting}

\section{More Details on Quality Control}
\label{sec:quality}

\subsection{Data Filtration}
During data collection, we apply multiple filtering steps as described in the main text. Human reviewers first select videos that match our task setting based on their content (\eg, cooking). They then check the quality and consistency of the extracted initial- and final-scene images and ensure that the descriptions of atomic actions are sufficient for predicting the final scene (see the main text Sec. 3.2 for details). In the end, we obtain only 1,157 qualified instances from a total of 20,235 raw videos collected from open-source datasets and our own recordings. In the final dataset, 948 instances come from Ego4D and 189 from Ego-Exo4D, since Ego4D is larger in scale and offers more diverse scenarios. In addition, 20 self-recorded videos serve as the data source for the abnormal cases.

\subsection{Scene Annotation Correction}
For scene annotation correction, we first sample 5\% of the data and ask annotators from three data-labeling companies to conduct pilot annotations. We then select the company with the highest annotation quality and commission it to annotate the full dataset. Annotators are required to cross-check each other's work to ensure quality. During the annotation process, we hold multiple meetings with them to align on the annotation guidelines. After annotation is completed, our internal specialists review the quality and accept the deliverables only if they meet a 99\% accuracy requirement in spot-checking.
Below are the annotation guidelines provided to the human annotators from the data-labeling company. For brevity, we omit the examples.

\noindent{\bf Raw Data.}
Images and their paired JSON files containing scene annotations (files in each pair share the same filename). The format of each scene annotation is as follows:

\begin{lstlisting}[style=prompt, caption={Format of the scene annotation}, label={lst:anno_format}]
[
  {
    "category": "bowl",
    "description": "white ceramic bowl with a smooth surface",
    "relation": [
      "bowl on kitchen counter",
      "bowl next to sink"
    ]
  },
  {
    "category": "cabbage",
    "description": "shredded green cabbage, appearing fresh and crisp",
    "relation": [
      "cabbage in colander"
    ]
  },
  {
    "category": "colander",
    "description": "green plastic colander with a mesh texture",
    "relation": [
      "colander in sink"
    ]
  }
]
\end{lstlisting}

Each scene annotation contains a list, where each dictionary represents an object. In each dictionary:
\begin{itemize}
\item The value of the key \verb|"category"| specifies the object category (a noun or noun phrase).
\item The value of the key \verb|"description"| records the object's visual attributes (including but not limited to shape, size, color, material, texture, and state).
\item The value of the key \verb|"relation"| records relations between objects (subject + relation + object), where the subject is the \verb|"category"| value of the object represented by this dictionary, and the relations may include spatial relations or interaction relations.
\end{itemize}

\noindent{\bf Rules.} Correct each scene annotation by following the rules below.
\begin{itemize}
\item Correct and refine the values of the keys \verb|"description"| and \verb|"relation"| based on the image content.
\item Add dictionaries for objects missing from the annotations in the required format and fill in the necessary key-value pairs.
\item Delete dictionaries that represent objects not present in the image after repeated verification.
\item Delete any dictionary whose \verb|"category"| value is not a noun or noun phrase denoting an object category. Dictionaries whose \verb|"category"| values represent human body parts should also be deleted.
\item Delete adjectives that describe object attributes from the value of \verb|"category"|.
\item Delete phrases describing inter-object relations or object positions in the image from the value of \verb|"description"|.
\end{itemize}

\section{Detailed evaluation setups}
\label{sec:setups}
\noindent{\bf Human Evaluation Setup.}
Following VSI-Bench \cite{yang2025thinking}, during the evaluation of human performance on EXPLORE-Bench (tiny), human participants are allowed unlimited time to answer questions.

\noindent{\bf Model Configurations.}
For the open-source MLLMs we evaluate and the LLM used as the scorer, we set \verb|do_sample=False|. And we set \verb|max_new_tokens=8192| (if available) for the evaluated MLLMs. For the proprietary MLLMs we evaluate, we set \verb|temperature=0| and \verb|max_tokens=8192|.

\noindent{\bf Implementation Details.}
We use the official repository and checkpoint to evaluate each open-source MLLM. All evaluations are conducted using NVIDIA H200 GPUs.

\noindent{\bf LLM-based Scorers.}
To enable researchers to run evaluations on our benchmark using a single consumer-grade GPU, we use an open-source LLM with no more than 10B parameters as the scorer. In addition to Qwen3-8B \cite{yang2025qwen3}, we also try other LLMs as the scorer, including Mistralai-3-8B-Instruct \cite{mistral3} and Hunyuan-7B-Instruct \cite{hunyuan}. Their scores achieve Spearman correlations (\(\rho\)) of 0.874 and 0.815 with human evaluations, respectively. As mentioned in the main text, using Qwen3-8B yields \(\rho = 0.919\), so we currently adopt Qwen3-8B as the scorer.

\begin{table*}[h]
\caption{\textbf{Evaluation results of abnormal cases.} \colorbox{rank1green}{Dark green} and \colorbox{rank2green}{light green} indicate the best and the second best result among all models. For models with both thinking and non-thinking modes, we report results in the non-thinking mode by default.}
\label{tab:eval_abnormal_app}
\centering
\setlength{\tabcolsep}{3pt} 

\begin{tabular}{l|ccc >{\columncolor{lightgraybg}}c c >{\columncolor{lightgraybg}}c}
\toprule
\textbf{Methods} & $\bm{S_{obj}}$ & $\bm{S_{att}}$ & $\bm{S_{rel}}$  & $\bm{S_{uni}}$ & $\bm{S_{key}}$  & $\bm{S_{abn}}$ \\
\midrule
\rowcolor{lightcyanbg} Human & 95.19 & 3.64 & 3.88 & 80.32 & 4.65 & 91.64 \\
Embodied-Reasoner & 63.75 & 1.62 & 2.75 & 49.28 & 1.45 & 30.95 \\
Qwen2.5-VL-3B-Instruct & 74.32 & 2.30 & 3.07 & 59.24 & 1.84 & 39.07 \\
Qwen3-VL-2B-Instruct & 79.96 & 2.36 & 3.44 & 64.03 & 1.93 & 41.20 \\
Qwen2-VL-7B-Instruct & 68.84 & 2.13 & 2.98 & 55.98 & 2.20 & 45.20 \\
Ovis2.5-2B & 81.89 & 2.63 & 3.25 & 64.90 & 2.19 & 45.94 \\
MiMo-VL-7B-RL-2508 & 76.79 & 2.41 & 3.10 & 60.81 & 2.28 & 47.18 \\
LLaVA-OneVision-1.5-8B-Instruct & 74.25 & 2.37 & 3.00 & 59.18 & 2.31 & 47.47 \\
EgoThinker & 72.33 & 2.21 & 2.92 & 56.92 & 2.38 & 48.44 \\
Qwen2.5-VL-7B-Instruct & 78.22 & 2.46 & 3.22 & 62.50 & 2.36 & 48.77 \\
InternVL3.5-8B & 77.50 & 2.47 & 3.23 & 62.57 & 2.38 & 49.16 \\
Qwen3-VL-8B-Instruct & 83.83 & 2.82 & 3.51 & 68.75 & 2.54 & 52.63 \\
MiniCPM-V-4.5 & 81.86 & 2.61 & 3.35 & 65.58 & 2.69 & 55.01 \\
Keye-VL-1.5-8B & 74.75 & 2.32 & 3.04 & 59.26 & 2.75 & 55.43 \\
Gemini-3-Pro & 80.98 & 2.70 & 3.41 & 66.47 & 2.74 & 56.00 \\
Qwen3-VL-8B-Thinking & \cellcolor{rank2green}86.03 & \cellcolor{rank1green}2.89 & \cellcolor{rank1green}3.62 & \cellcolor{rank1green}70.71 & \cellcolor{rank2green}2.84 & \cellcolor{rank2green}58.22 \\
GPT-5.2-Chat & \cellcolor{rank1green}87.32 & \cellcolor{rank2green}2.82 & \cellcolor{rank2green}3.55 & \cellcolor{rank2green}69.94 & \cellcolor{rank1green}3.10 & \cellcolor{rank1green}62.79 \\
\bottomrule
\end{tabular}
\end{table*}


\begin{figure}[t]
  \centering
  \begin{subfigure}{0.49\linewidth}
    \centering
    \includegraphics[width=\linewidth]{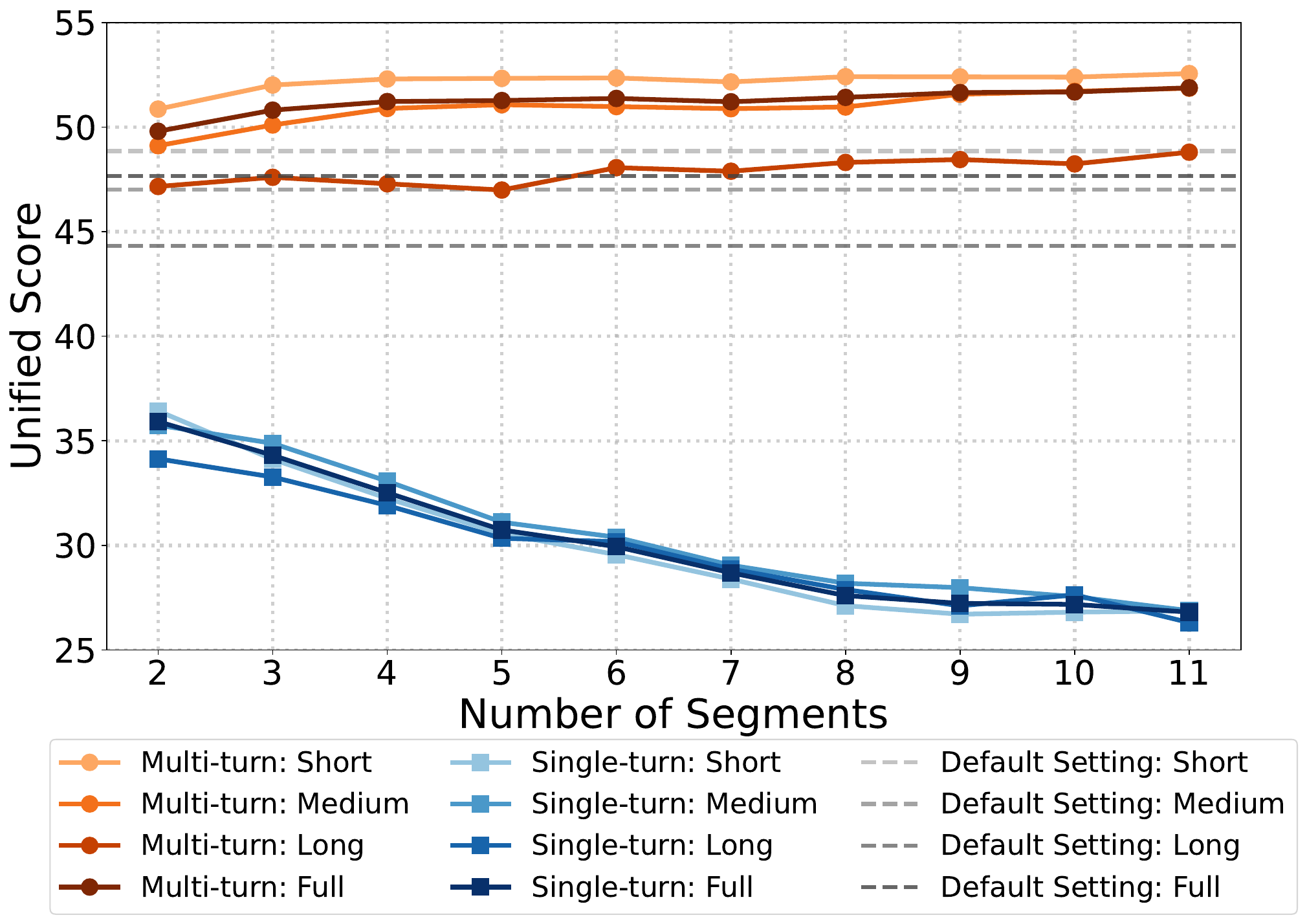} 
    \caption{Unified score \vs number of segments}
    \label{fig:score_segment_ovis}
  \end{subfigure}
  \hfill 
  \begin{subfigure}{0.49\linewidth}
    \centering
    \includegraphics[width=\linewidth]{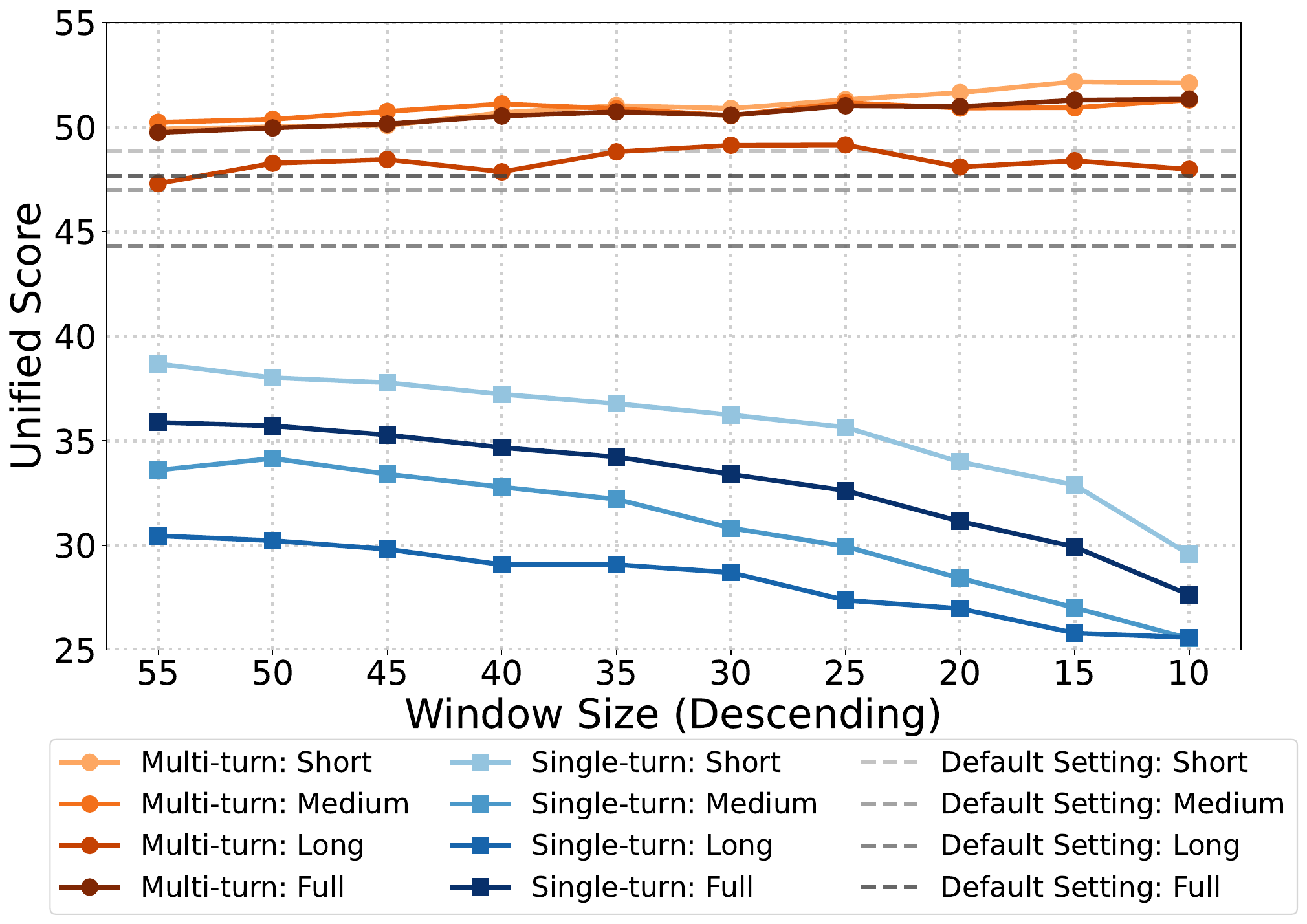}
    \caption{Unified score \vs window size}
    \label{fig:score_window_ovis}
  \end{subfigure}
  \caption{\textbf{Unified score $\bm{S_{uni}}$ of Ovis2.5-9B across subsets under different inference strategies.} \texttt{Short}, \texttt{Medium}, and \texttt{Long} denote the subsets with short, medium, and long atomic-action sequences. \texttt{Full} denotes the full dataset.}
  \label{fig:infer_score_ovis}
\end{figure}

\begin{figure}[t]
  \centering
  \begin{subfigure}{0.49\linewidth}
    \centering
    \includegraphics[width=\linewidth]{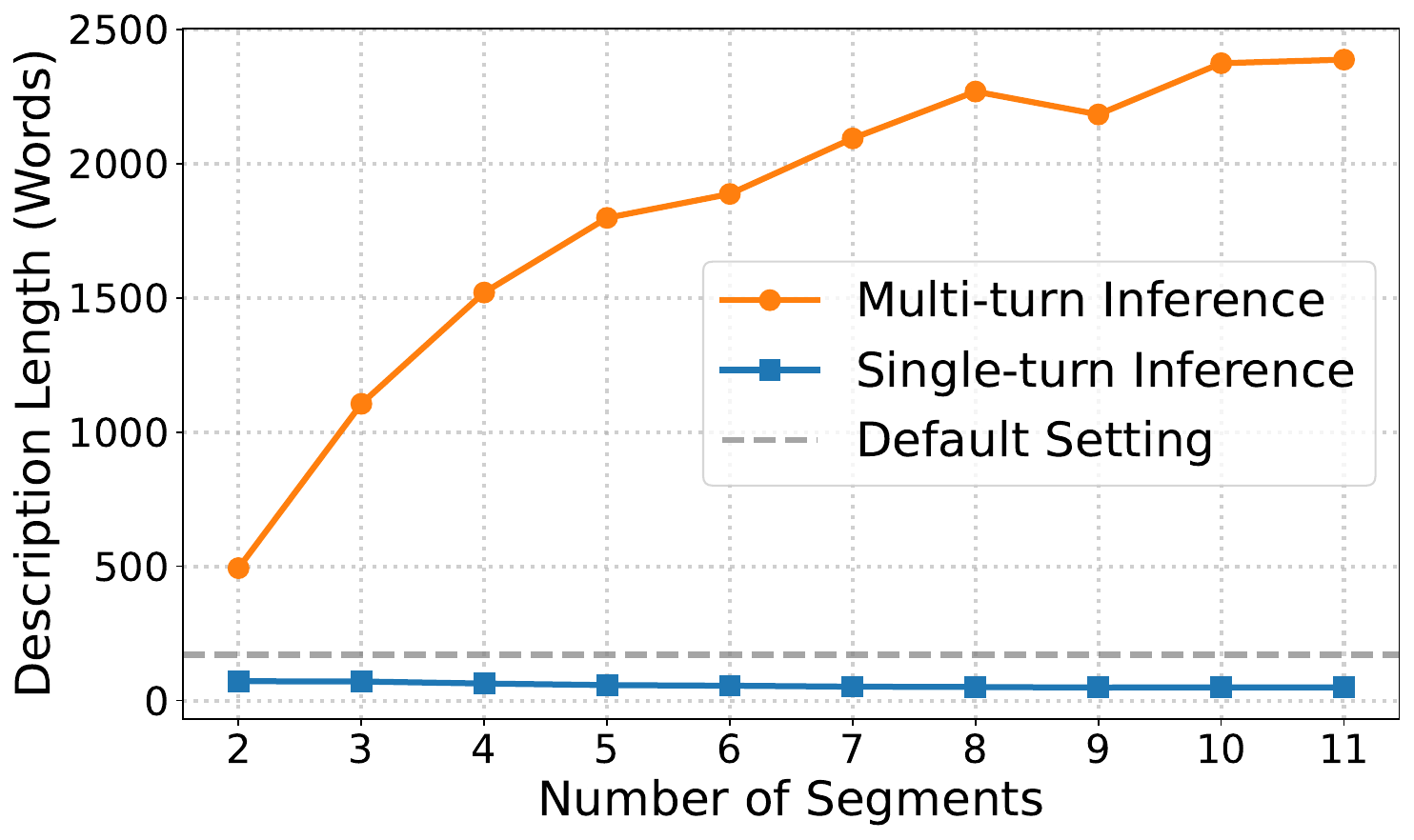} 
    \caption{Description length \vs number of segments}
    \label{fig:length_segment_ovis}
  \end{subfigure}
  \hfill 
  \begin{subfigure}{0.49\linewidth}
    \centering
    \includegraphics[width=\linewidth]{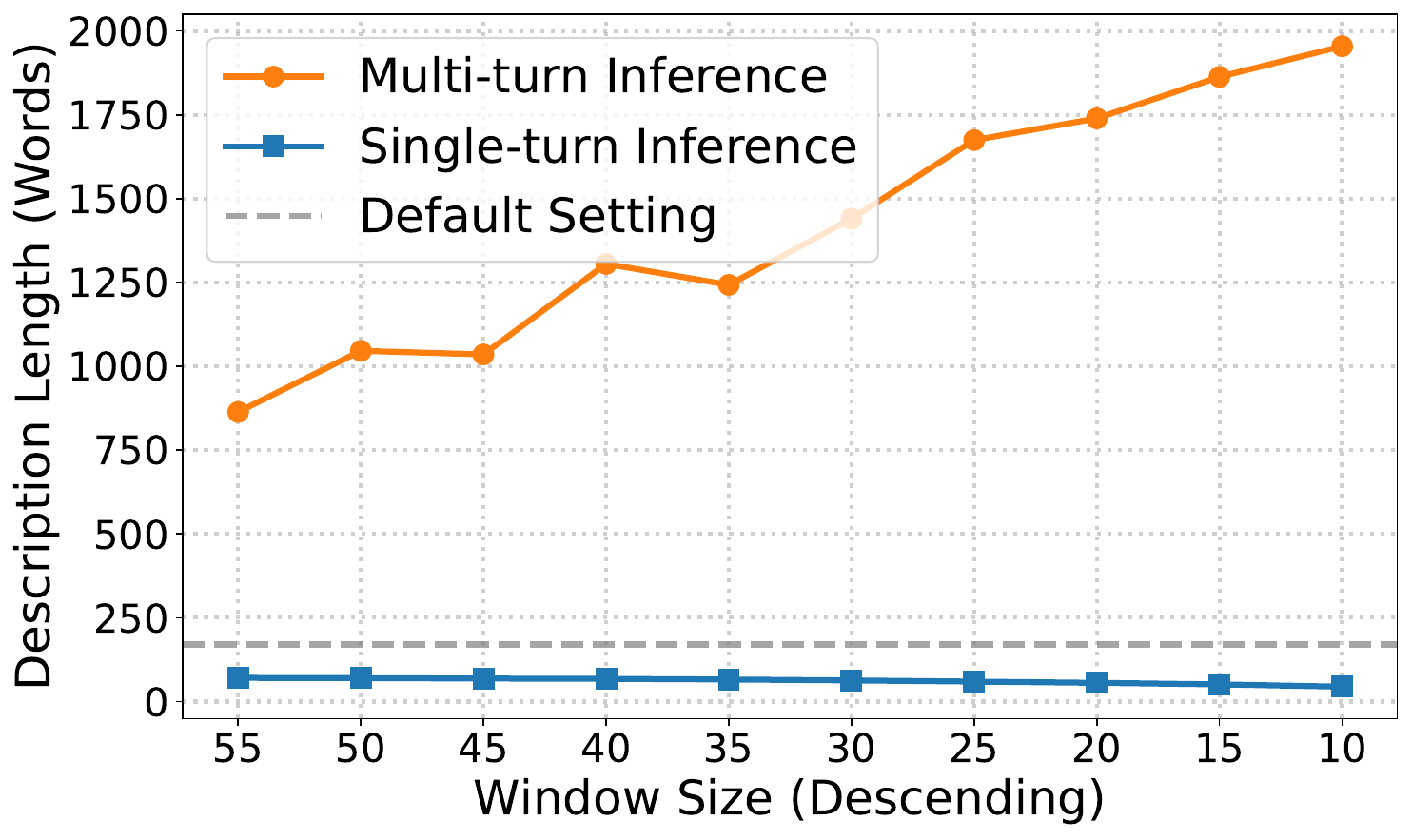}
    \caption{Description length \vs window size}
    \label{fig:length_window_ovis}
  \end{subfigure}
  \caption{\textbf{Average single-instance final-scene description length of Ovis2.5-9B under different inference strategies.}}
  \label{fig:infer_length_ovis}
\end{figure}

\begin{figure}[t]
  \centering
  \begin{subfigure}{0.49\linewidth}
    \centering
    \includegraphics[width=\linewidth]{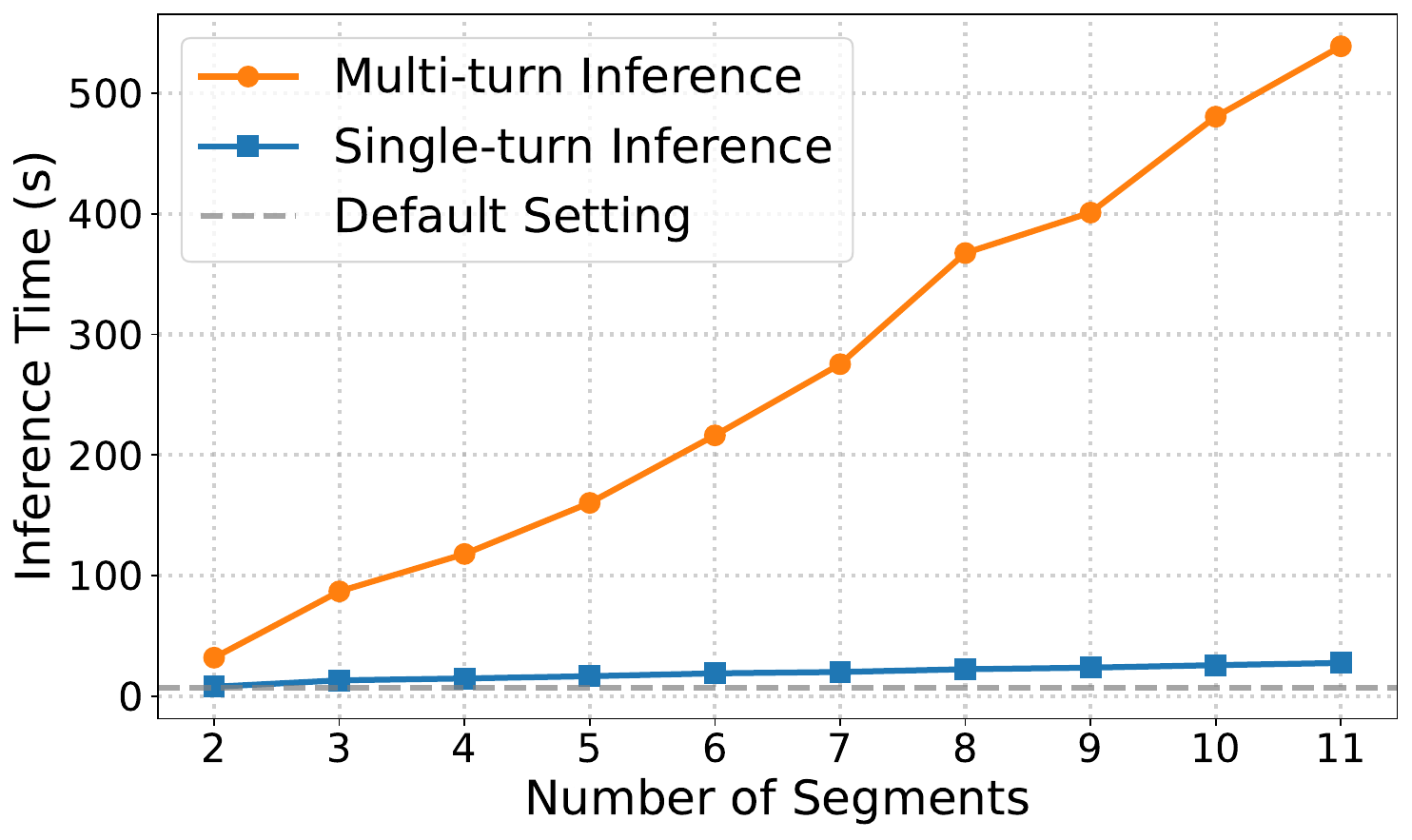} 
    \caption{Inference time \vs number of segments}
    \label{fig:time_segment_ovis}
  \end{subfigure}
  \hfill 
  \begin{subfigure}{0.49\linewidth}
    \centering
    \includegraphics[width=\linewidth]{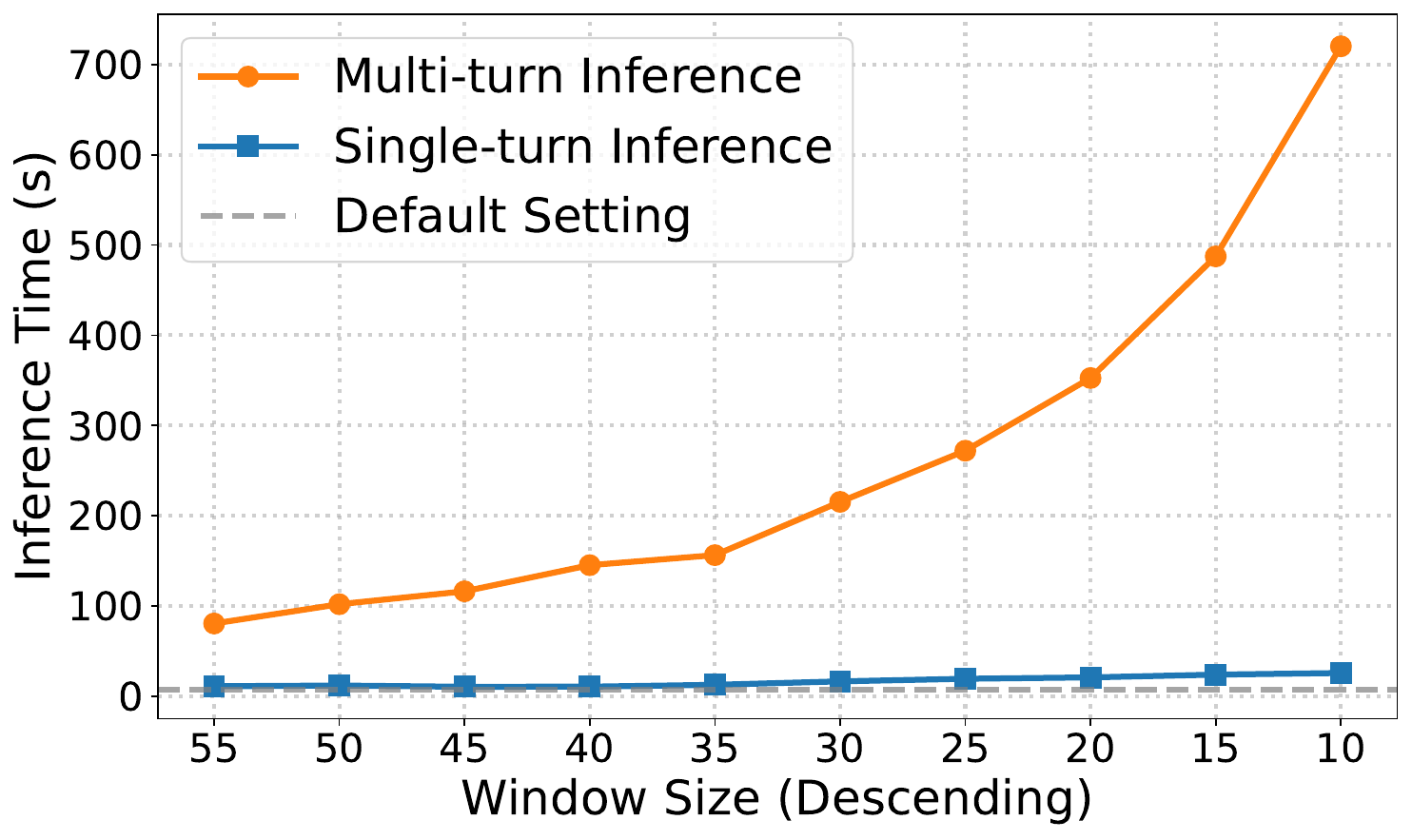}
    \caption{Inference time \vs window size}
    \label{fig:time_window_ovis}
  \end{subfigure}
  \caption{\textbf{Average single-instance inference time of Ovis2.5-9B on an NVIDIA H200 GPU under different inference strategies.}}
  \label{fig:infer_time_ovis}
\end{figure}

\clearpage

\begin{table*}[!tp]
\caption{\textbf{Evaluation results of Qwen3-VL-8B-Instruct under different inference strategies.} \texttt{Short}, \texttt{Medium} and \texttt{Long} mean the subsets with short, medium and long atomic-action sequences, respectively. \texttt{Full} denotes the full dataset. \colorbox{rank1green}{Dark green} and \colorbox{rank2green}{light green} indicate the best and the second best result among all strategies.}
\label{tab:eval_infer}
\centering
\renewcommand{\arraystretch}{1.1} 
\setlength{\tabcolsep}{2.5pt} 

\resizebox{\textwidth}{!}{
\begin{tabular}{l|cccc|cccc|cccc|cccc}
\toprule
\multirow{2}{*}{\textbf{Partitioning}} & \multicolumn{4}{c|}{\texttt{Short}} & \multicolumn{4}{c|}{\texttt{Medium}} & \multicolumn{4}{c|}{\texttt{Long}} & \multicolumn{4}{c}{\texttt{Full}} \\
 & {$\bm{S_{obj}}$} & {$\bm{S_{att}}$} & {$\bm{S_{rel}}$} & {$\bm{S_{uni}}$} & {$\bm{S_{obj}}$} & {$\bm{S_{att}}$} & {$\bm{S_{rel}}$} & {$\bm{S_{uni}}$} & {$\bm{S_{obj}}$} & {$\bm{S_{att}}$} & {$\bm{S_{rel}}$} & {$\bm{S_{uni}}$} & {$\bm{S_{obj}}$} & {$\bm{S_{att}}$} & {$\bm{S_{rel}}$} & {$\bm{S_{uni}}$} \\ \midrule \multicolumn{17}{c}{\textit{Default Setting}} \\ \midrule
\rowcolor{lightcyanbg} \texttt{segment\_num = 1} & 61.34 & 1.88 & 2.84 & 51.23 & 60.78 & 1.85 & 2.81 & 50.64 & 56.83 & 1.73 & 2.71 & 48.00 & 60.63 & 1.85 & 2.82 & 50.65 \\ \midrule
\multicolumn{17}{c}{\textit{Single-turn Inference with Various} \texttt{segment\_num}} \\ \midrule
\texttt{segment\_num = 11} & 37.20 & 0.89 & 1.77 & 29.69 & 36.42 & 0.86 & 1.75 & 29.10 & 33.31 & 0.79 & 1.72 & 27.59 & 36.48 & 0.87 & 1.75 & 29.24 \\
\texttt{segment\_num = 10} & 37.91 & 0.91 & 1.80 & 30.25 & 36.45 & 0.87 & 1.75 & 29.19 & 35.74 & 0.86 & 1.77 & 29.09 & 37.13 & 0.89 & 1.78 & 29.73 \\
\texttt{segment\_num = 9} & 37.88 & 0.89 & 1.80 & 30.11 & 37.82 & 0.90 & 1.81 & 30.24 & 34.59 & 0.83 & 1.80 & 28.86 & 37.49 & 0.89 & 1.80 & 30.02 \\
\texttt{segment\_num = 8} & 38.89 & 0.95 & 1.85 & 31.21 & 37.29 & 0.89 & 1.82 & 30.10 & 36.94 & 0.90 & 1.84 & 30.21 & 38.08 & 0.92 & 1.84 & 30.69 \\
\texttt{segment\_num = 7} & 40.06 & 0.98 & 1.89 & 32.00 & 38.98 & 0.94 & 1.90 & 31.52 & 37.52 & 0.93 & 1.95 & 31.50 & 39.38 & 0.96 & 1.90 & 31.77 \\
\texttt{segment\_num = 6} & 40.50 & 0.99 & 1.92 & 32.40 & 39.59 & 0.97 & 1.92 & 32.02 & 37.63 & 0.93 & 1.94 & 31.40 & 39.85 & 0.97 & 1.92 & 32.15 \\
\texttt{segment\_num = 5} & 39.57 & 0.97 & 1.91 & 32.00 & 40.29 & 0.99 & 1.93 & 32.44 & 37.78 & 0.97 & 1.96 & 31.95 & 39.64 & 0.98 & 1.93 & 32.16 \\
\texttt{segment\_num = 4} & 40.12 & 1.00 & 1.92 & 32.33 & 41.91 & 1.04 & 2.00 & 33.73 & 39.99 & 1.01 & 2.01 & 33.16 & 40.77 & 1.01 & 1.96 & 32.94 \\
\texttt{segment\_num = 3} & 42.23 & 1.07 & 2.02 & 34.19 & 42.97 & 1.09 & 2.06 & 34.89 & 41.26 & 1.07 & 2.08 & 34.47 & 42.40 & 1.08 & 2.04 & 34.48 \\
\texttt{segment\_num = 2} & 46.66 & 1.22 & 2.19 & 37.72 & 46.98 & 1.22 & 2.22 & 38.10 & 44.68 & 1.21 & 2.25 & 37.67 & 46.56 & 1.22 & 2.21 & 37.86 \\ \midrule
\multicolumn{17}{c}{\textit{Single-turn Inference with Various} \texttt{window\_size}} \\ \midrule
\texttt{window\_size = 10} & 38.08 & 0.96 & 1.89 & 31.35 & 33.37 & 0.80 & 1.68 & 27.41 & 31.25 & 0.76 & 1.64 & 26.20 & 35.57 & 0.88 & 1.78 & 29.32 \\
\texttt{window\_size = 15} & 41.25 & 1.06 & 2.01 & 33.86 & 36.09 & 0.87 & 1.78 & 29.32 & 31.84 & 0.78 & 1.63 & 26.50 & 38.29 & 0.96 & 1.88 & 31.36 \\
\texttt{window\_size = 20} & 42.93 & 1.13 & 2.10 & 35.44 & 37.67 & 0.91 & 1.86 & 30.68 & 31.64 & 0.75 & 1.64 & 26.26 & 39.73 & 1.01 & 1.96 & 32.66 \\
\texttt{window\_size = 25} & 43.45 & 1.14 & 2.11 & 35.74 & 38.23 & 0.93 & 1.89 & 31.21 & 32.67 & 0.79 & 1.71 & 27.37 & 40.31 & 1.02 & 1.99 & 33.13 \\
\texttt{window\_size = 30} & 44.71 & 1.21 & 2.19 & 37.16 & 40.08 & 1.00 & 1.96 & 32.70 & 33.62 & 0.80 & 1.69 & 27.54 & 41.77 & 1.09 & 2.05 & 34.44 \\
\texttt{window\_size = 35} & 46.06 & 1.24 & 2.23 & 38.01 & 39.58 & 0.98 & 1.95 & 32.35 & 35.50 & 0.88 & 1.85 & 29.82 & 42.48 & 1.10 & 2.08 & 35.00 \\
\texttt{window\_size = 40} & 46.48 & 1.25 & 2.24 & 38.34 & 40.72 & 1.02 & 2.00 & 33.30 & 35.68 & 0.86 & 1.83 & 29.61 & 43.15 & 1.12 & 2.11 & 35.50 \\
\texttt{window\_size = 45} & 47.40 & 1.29 & 2.26 & 39.01 & 41.42 & 1.03 & 2.02 & 33.73 & 36.84 & 0.91 & 1.91 & 30.83 & 44.01 & 1.15 & 2.13 & 36.14 \\
\texttt{window\_size = 50} & 47.82 & 1.31 & 2.29 & 39.44 & 41.37 & 1.05 & 2.05 & 34.11 & 36.70 & 0.91 & 1.87 & 30.50 & 44.19 & 1.17 & 2.16 & 36.47 \\
\texttt{window\_size = 55} & 48.11 & 1.32 & 2.32 & 39.83 & 42.90 & 1.09 & 2.09 & 35.06 & 37.12 & 0.95 & 1.94 & 31.42 & 44.96 & 1.20 & 2.19 & 37.13 \\ \midrule
\multicolumn{17}{c}{\textit{Mutil-turn Inference with Various} \texttt{segment\_num}} \\ \midrule
\texttt{segment\_num = 2} & 60.43 & 1.85 & 2.81 & 50.56 & 59.07 & 1.78 & 2.75 & 49.24 & 55.31 & 1.67 & 2.64 & 46.64 & 59.36 & 1.80 & 2.77 & 49.64 \\
\texttt{segment\_num = 3} & 60.94 & 1.87 & 2.83 & 50.94 & 58.79 & 1.76 & 2.74 & 48.90 & 55.90 & 1.67 & 2.67 & 46.99 & 59.59 & 1.81 & 2.77 & 49.74 \\
\texttt{segment\_num = 4} & 61.23 & 1.88 & 2.84 & 51.20 & 59.93 & 1.77 & 2.76 & 49.45 & 54.85 & 1.62 & 2.64 & 46.20 & 60.04 & 1.81 & 2.79 & 50.00 \\
\texttt{segment\_num = 5} & 61.70 & 1.88 & 2.85 & 51.39 & 60.75 & 1.81 & 2.81 & 50.35 & 54.85 & 1.60 & 2.62 & 45.90 & 60.59 & 1.82 & 2.81 & 50.39 \\
\texttt{segment\_num = 6} & 61.89 & 1.90 & 2.85 & 51.55 & 60.59 & 1.78 & 2.80 & 50.04 & 56.65 & 1.69 & 2.70 & 47.59 & 60.83 & 1.83 & 2.81 & 50.55 \\
\texttt{segment\_num = 7} & 62.87 & \cellcolor{rank2green}1.92 & 2.88 & 52.23 & 61.65 & 1.82 & 2.84 & 50.84 & 56.96 & 1.68 & 2.71 & 47.72 & 61.76 & 1.86 & 2.84 & 51.21 \\
\texttt{segment\_num = 8} & 62.77 & 1.92 & 2.90 & 52.35 & 61.88 & 1.85 & 2.86 & 51.28 & 56.87 & 1.68 & 2.73 & 47.87 & 61.78 & 1.87 & 2.87 & 51.46 \\
\texttt{segment\_num = 9} & \cellcolor{rank2green}63.08 & 1.92 & 2.90 & \cellcolor{rank2green}52.41 & 62.13 & 1.85 & 2.86 & 51.39 & 57.71 & 1.69 & 2.77 & 48.41 & 62.13 & 1.87 & 2.87 & 51.59 \\
\texttt{segment\_num = 10} & 62.83 & 1.92 & 2.89 & 52.29 & 62.84 & 1.86 & 2.86 & 51.57 & 59.01 & 1.75 & 2.74 & 48.90 & 62.41 & 1.88 & 2.86 & 51.65 \\
\texttt{segment\_num = 11} & \cellcolor{rank1green}64.07 & \cellcolor{rank1green}1.96 & \cellcolor{rank1green}2.94 & \cellcolor{rank1green}53.26 & \cellcolor{rank2green}63.36 & \cellcolor{rank2green}1.88 & \cellcolor{rank1green}2.91 & \cellcolor{rank1green}52.26 & 57.89 & 1.70 & 2.74 & 48.33 & \cellcolor{rank1green}63.12 & \cellcolor{rank1green}1.90 & \cellcolor{rank1green}2.91 & \cellcolor{rank1green}52.34 \\ \midrule
\multicolumn{17}{c}{\textit{Mutil-turn Inference with Various} \texttt{window\_size}} \\ \midrule
\texttt{window\_size = 55} & 61.36 & 1.88 & 2.85 & 51.29 & 59.59 & 1.78 & 2.76 & 49.41 & 55.96 & 1.67 & 2.70 & 47.32 & 60.10 & 1.82 & 2.80 & 50.15 \\
\texttt{window\_size = 50} & 60.99 & 1.86 & 2.82 & 50.85 & 60.02 & 1.80 & 2.78 & 49.82 & 55.88 & 1.67 & 2.70 & 47.20 & 60.07 & 1.82 & 2.79 & 50.06 \\
\texttt{window\_size = 45} & 61.26 & 1.87 & 2.85 & 51.16 & 60.67 & 1.80 & 2.78 & 49.98 & 58.18 & 1.71 & 2.71 & 48.25 & 60.70 & 1.82 & 2.81 & 50.40 \\
\texttt{window\_size = 40} & 61.02 & 1.87 & 2.83 & 50.99 & 59.69 & 1.79 & 2.77 & 49.60 & 57.69 & 1.70 & 2.73 & 48.11 & 60.16 & 1.82 & 2.80 & 50.16 \\
\texttt{window\_size = 35} & 60.93 & 1.86 & 2.83 & 50.86 & 60.09 & 1.80 & 2.77 & 49.79 & 57.87 & 1.70 & 2.75 & 48.37 & 60.28 & 1.82 & 2.80 & 50.19 \\
\texttt{window\_size = 30} & 60.89 & 1.86 & 2.82 & 50.77 & 60.28 & 1.81 & 2.80 & 50.11 & 58.72 & 1.73 & 2.77 & 48.98 & 60.42 & 1.82 & 2.81 & 50.32 \\
\texttt{window\_size = 25} & 61.18 & 1.87 & 2.83 & 51.00 & 61.64 & 1.83 & 2.85 & 50.96 & 57.74 & 1.68 & 2.74 & 48.10 & 60.97 & 1.83 & 2.83 & 50.67 \\
\texttt{window\_size = 20} & 61.59 & 1.88 & 2.85 & 51.36 & 61.65 & 1.83 & 2.84 & 50.94 & \cellcolor{rank2green}60.15 & \cellcolor{rank2green}1.76 & 2.80 & 49.77 & 61.46 & 1.85 & 2.84 & 51.03 \\
\texttt{window\_size = 15} & 61.24 & 1.87 & 2.85 & 51.14 & 62.85 & 1.86 & 2.88 & 51.78 & 59.89 & 1.75 & \cellcolor{rank2green}2.83 & \cellcolor{rank2green}49.83 & 61.69 & 1.85 & 2.86 & 51.23 \\
\texttt{window\_size = 10} & 62.68 & 1.91 & \cellcolor{rank2green}2.91 & 52.31 & \cellcolor{rank1green}63.39 & \cellcolor{rank1green}1.88 & \cellcolor{rank2green}2.89 & \cellcolor{rank2green}52.17 & \cellcolor{rank1green}61.59 & \cellcolor{rank1green}1.83 & \cellcolor{rank1green}2.90 & \cellcolor{rank1green}51.41 & \cellcolor{rank2green}62.83 & \cellcolor{rank2green}1.89 & \cellcolor{rank2green}2.90 & \cellcolor{rank2green}52.16 \\
\bottomrule
\end{tabular}
}
\end{table*}

\begin{table*}[!tp]
\caption{\textbf{Evaluation results of Ovis2.5-9B under different inference strategies.} \texttt{Short}, \texttt{Medium} and \texttt{Long} mean the subsets with short, medium and long atomic-action sequences, respectively. \texttt{Full} denotes the full dataset. \colorbox{rank1green}{Dark green} and \colorbox{rank2green}{light green} indicate the best and the second best result among all strategies.}
\label{tab:eval_infer_ovis}
\centering
\renewcommand{\arraystretch}{1.1} 
\setlength{\tabcolsep}{2.5pt} 

\resizebox{\textwidth}{!}{
\begin{tabular}{l|cccc|cccc|cccc|cccc}
\toprule
\multirow{2}{*}{\textbf{Partitioning}} & \multicolumn{4}{c|}{\texttt{Short}} & \multicolumn{4}{c|}{\texttt{Medium}} & \multicolumn{4}{c|}{\texttt{Long}} & \multicolumn{4}{c}{\texttt{Full}} \\
 & {$\bm{S_{obj}}$} & {$\bm{S_{att}}$} & {$\bm{S_{rel}}$} & {$\bm{S_{uni}}$} & {$\bm{S_{obj}}$} & {$\bm{S_{att}}$} & {$\bm{S_{rel}}$} & {$\bm{S_{uni}}$} & {$\bm{S_{obj}}$} & {$\bm{S_{att}}$} & {$\bm{S_{rel}}$} & {$\bm{S_{uni}}$} & {$\bm{S_{obj}}$} & {$\bm{S_{att}}$} & {$\bm{S_{rel}}$} & {$\bm{S_{uni}}$} \\ \midrule \multicolumn{17}{c}{\textit{Default Setting}} \\ \midrule
\rowcolor{lightcyanbg} \texttt{segment\_num = 1} & 59.51 & 1.74 & 2.72 & 48.85 & 57.84 & 1.65 & 2.63 & 47.01 & 53.83 & 1.55 & 2.50 & 44.32 & 58.26 & 1.68 & 2.66 & 47.66 \\ \midrule
\multicolumn{17}{c}{\textit{Single-turn Inference with Various} \texttt{segment\_num}} \\ \midrule
\texttt{segment\_num = 11} & 34.30 & 0.75 & 1.63 & 26.87 & 32.43 & 0.77 & 1.67 & 26.88 & 30.10 & 0.75 & 1.69 & 26.31 & 33.14 & 0.76 & 1.66 & 26.81 \\
\texttt{segment\_num = 10} & 34.16 & 0.74 & 1.63 & 26.81 & 33.33 & 0.80 & 1.70 & 27.55 & 31.76 & 0.80 & 1.76 & 27.64 & 33.59 & 0.77 & 1.67 & 27.18 \\
\texttt{segment\_num = 9} & 33.89 & 0.74 & 1.63 & 26.71 & 33.75 & 0.81 & 1.74 & 27.98 & 30.69 & 0.78 & 1.75 & 27.11 & 33.48 & 0.77 & 1.68 & 27.23 \\
\texttt{segment\_num = 8} & 33.91 & 0.77 & 1.66 & 27.12 & 34.22 & 0.82 & 1.74 & 28.19 & 31.84 & 0.81 & 1.78 & 27.89 & 33.80 & 0.79 & 1.70 & 27.60 \\
\texttt{segment\_num = 7} & 35.28 & 0.82 & 1.73 & 28.39 & 35.09 & 0.85 & 1.79 & 29.05 & 33.25 & 0.84 & 1.84 & 28.87 & 34.98 & 0.83 & 1.76 & 28.69 \\
\texttt{segment\_num = 6} & 36.24 & 0.87 & 1.80 & 29.56 & 36.54 & 0.91 & 1.86 & 30.39 & 34.55 & 0.90 & 1.91 & 30.18 & 36.16 & 0.89 & 1.84 & 29.94 \\
\texttt{segment\_num = 5} & 37.83 & 0.90 & 1.85 & 30.56 & 37.47 & 0.92 & 1.91 & 31.11 & 34.86 & 0.90 & 1.92 & 30.34 & 37.37 & 0.91 & 1.88 & 30.74 \\
\texttt{segment\_num = 4} & 39.51 & 0.96 & 1.95 & 32.25 & 39.79 & 1.02 & 2.00 & 33.07 & 36.85 & 0.96 & 1.99 & 31.91 & 39.32 & 0.98 & 1.98 & 32.52 \\
\texttt{segment\_num = 3} & 41.59 & 1.05 & 2.05 & 34.12 & 42.05 & 1.07 & 2.11 & 34.88 & 38.66 & 1.01 & 2.07 & 33.27 & 41.43 & 1.05 & 2.07 & 34.31 \\
\texttt{segment\_num = 2} & 44.37 & 1.15 & 2.16 & 36.43 & 43.02 & 1.13 & 2.14 & 35.75 & 39.06 & 1.07 & 2.11 & 34.14 & 43.28 & 1.13 & 2.15 & 35.92 \\ \midrule
\multicolumn{17}{c}{\textit{Single-turn Inference with Various} \texttt{window\_size}} \\ \midrule
\texttt{window\_size = 10} & 35.96 & 0.86 & 1.82 & 29.57 & 30.74 & 0.72 & 1.61 & 25.55 & 29.45 & 0.70 & 1.66 & 25.60 & 33.30 & 0.79 & 1.72 & 27.64 \\
\texttt{window\_size = 15} & 39.42 & 1.01 & 2.00 & 32.89 & 32.39 & 0.78 & 1.69 & 27.02 & 29.55 & 0.71 & 1.68 & 25.81 & 35.71 & 0.89 & 1.85 & 29.93 \\
\texttt{window\_size = 20} & 41.09 & 1.04 & 2.05 & 33.99 & 33.86 & 0.83 & 1.77 & 28.43 & 31.06 & 0.76 & 1.73 & 26.98 & 37.30 & 0.93 & 1.91 & 31.15 \\
\texttt{window\_size = 25} & 43.00 & 1.11 & 2.14 & 35.65 & 35.60 & 0.88 & 1.86 & 29.95 & 31.54 & 0.77 & 1.76 & 27.38 & 38.98 & 0.99 & 2.00 & 32.62 \\
\texttt{window\_size = 30} & 43.86 & 1.15 & 2.15 & 36.24 & 36.72 & 0.92 & 1.90 & 30.83 & 32.74 & 0.82 & 1.84 & 28.70 & 39.98 & 1.03 & 2.02 & 33.40 \\
\texttt{window\_size = 35} & 44.45 & 1.18 & 2.18 & 36.78 & 38.52 & 0.97 & 1.97 & 32.21 & 33.21 & 0.85 & 1.85 & 29.08 & 41.00 & 1.06 & 2.07 & 34.23 \\
\texttt{window\_size = 40} & 44.70 & 1.20 & 2.20 & 37.23 & 39.33 & 0.99 & 2.00 & 32.79 & 33.51 & 0.83 & 1.86 & 29.08 & 41.47 & 1.08 & 2.09 & 34.68 \\
\texttt{window\_size = 45} & 45.40 & 1.23 & 2.23 & 37.78 & 39.95 & 1.02 & 2.03 & 33.41 & 34.32 & 0.86 & 1.90 & 29.82 & 42.15 & 1.11 & 2.12 & 35.28 \\
\texttt{window\_size = 50} & 45.45 & 1.24 & 2.24 & 38.02 & 41.01 & 1.05 & 2.07 & 34.16 & 34.27 & 0.85 & 1.96 & 30.23 & 42.56 & 1.13 & 2.15 & 35.72 \\
\texttt{window\_size = 55} & 46.11 & 1.27 & 2.28 & 38.68 & 40.40 & 1.03 & 2.03 & 33.60 & 34.67 & 0.91 & 1.93 & 30.46 & 42.72 & 1.14 & 2.15 & 35.88 \\ \midrule
\multicolumn{17}{c}{\textit{Mutil-turn Inference with Various} \texttt{segment\_num}} \\ \midrule
\texttt{segment\_num = 2} & 62.92 & 1.80 & 2.82 & 50.86 & 60.90 & 1.70 & 2.75 & 49.11 & 58.15 & \cellcolor{rank2green}1.61 & 2.67 & 47.16 & 61.64 & 1.74 & 2.77 & 49.80 \\
\texttt{segment\_num = 3} & 65.05 & \cellcolor{rank2green}1.82 & 2.88 & 52.01 & 63.01 & 1.69 & 2.81 & 50.10 & 58.71 & \cellcolor{rank1green}1.62 & 2.70 & 47.60 & 63.59 & \cellcolor{rank2green}1.75 & 2.83 & 50.81 \\
\texttt{segment\_num = 4} & 65.58 & 1.81 & 2.90 & 52.30 & 64.82 & 1.68 & 2.86 & 50.89 & 59.08 & 1.56 & 2.70 & 47.29 & 64.58 & 1.74 & 2.86 & 51.22 \\
\texttt{segment\_num = 5} & 65.61 & 1.80 & 2.91 & 52.33 & 65.45 & 1.68 & 2.87 & 51.07 & 59.79 & 1.48 & 2.71 & 46.99 & 64.91 & 1.72 & 2.87 & 51.27 \\
\texttt{segment\_num = 6} & \cellcolor{rank2green}65.68 & 1.81 & 2.91 & 52.35 & 65.47 & 1.64 & 2.89 & 50.98 & 60.56 & 1.54 & 2.77 & 48.06 & 65.04 & 1.72 & 2.89 & 51.37 \\
\texttt{segment\_num = 7} & 65.31 & 1.81 & 2.90 & 52.16 & 64.95 & 1.66 & 2.88 & 50.88 & 60.87 & 1.51 & 2.77 & 47.89 & 64.69 & 1.72 & 2.88 & 51.21 \\
\texttt{segment\_num = 8} & 65.45 & 1.81 & \cellcolor{rank2green}2.93 & \cellcolor{rank2green}52.41 & 64.70 & 1.66 & 2.89 & 50.96 & 61.95 & 1.56 & 2.74 & 48.31 & 64.79 & 1.73 & 2.89 & 51.42 \\
\texttt{segment\_num = 9} & 65.52 & 1.81 & 2.92 & 52.40 & \cellcolor{rank1green}65.73 & 1.71 & 2.90 & 51.57 & 60.93 & 1.55 & 2.79 & 48.45 & \cellcolor{rank2green}65.09 & 1.74 & 2.90 & 51.65 \\
\texttt{segment\_num = 10} & 65.36 & 1.80 & \cellcolor{rank1green}2.93 & 52.39 & 65.51 & 1.69 & \cellcolor{rank1green}2.94 & \cellcolor{rank2green}51.71 & 60.45 & 1.53 & 2.80 & 48.24 & 64.87 & 1.73 & \cellcolor{rank1green}2.92 & \cellcolor{rank2green}51.68 \\
\texttt{segment\_num = 11} & \cellcolor{rank1green}65.79 & \cellcolor{rank1green}1.82 & 2.92 & \cellcolor{rank1green}52.56 & \cellcolor{rank2green}65.61 & \cellcolor{rank1green}1.73 & \cellcolor{rank2green}2.92 & \cellcolor{rank1green}51.86 & 62.24 & 1.55 & 2.80 & 48.80 & \cellcolor{rank1green}65.33 & \cellcolor{rank1green}1.75 & \cellcolor{rank2green}2.91 & \cellcolor{rank1green}51.88 \\ \midrule
\multicolumn{17}{c}{\textit{Mutil-turn Inference with Various} \texttt{window\_size}} \\ \midrule
\texttt{window\_size = 55} & 61.31 & 1.77 & 2.78 & 49.91 & 63.21 & 1.71 & 2.81 & 50.23 & 60.73 & 1.51 & 2.70 & 47.30 & 61.95 & 1.72 & 2.78 & 49.74 \\
\texttt{window\_size = 50} & 61.52 & 1.77 & 2.78 & 50.02 & 63.21 & 1.71 & 2.83 & 50.37 & 60.90 & 1.54 & 2.78 & 48.27 & 62.08 & 1.72 & 2.80 & 49.96 \\
\texttt{window\_size = 45} & 61.67 & 1.77 & 2.79 & 50.08 & 63.96 & \cellcolor{rank2green}1.72 & 2.84 & 50.75 & 61.18 & 1.53 & \cellcolor{rank2green}2.81 & 48.45 & 62.47 & 1.72 & 2.81 & 50.15 \\
\texttt{window\_size = 40} & 62.48 & 1.79 & 2.82 & 50.69 & 65.02 & 1.71 & 2.86 & 51.11 & 60.65 & 1.51 & 2.77 & 47.86 & 63.22 & 1.73 & 2.83 & 50.53 \\
\texttt{window\_size = 35} & 62.98 & 1.79 & 2.84 & 51.03 & 64.90 & 1.68 & 2.86 & 50.88 & \cellcolor{rank2green}62.64 & 1.56 & 2.78 & 48.82 & 63.66 & 1.73 & 2.84 & 50.73 \\
\texttt{window\_size = 30} & 63.03 & 1.78 & 2.83 & 50.89 & 63.97 & 1.67 & 2.86 & 50.55 & 61.73 & 1.58 & \cellcolor{rank1green}2.83 & \cellcolor{rank2green}49.13 & 63.24 & 1.72 & 2.84 & 50.57 \\
\texttt{window\_size = 25} & 63.81 & 1.79 & 2.85 & 51.31 & 65.11 & 1.69 & 2.89 & 51.17 & \cellcolor{rank1green}63.10 & 1.56 & 2.80 & \cellcolor{rank1green}49.15 & 64.22 & 1.73 & 2.86 & 51.02 \\
\texttt{window\_size = 20} & 64.58 & 1.79 & 2.87 & 51.65 & 64.96 & 1.67 & 2.87 & 50.91 & 60.85 & 1.53 & 2.77 & 48.09 & 64.31 & 1.72 & 2.86 & 50.98 \\
\texttt{window\_size = 15} & 65.09 & 1.81 & 2.91 & 52.17 & 64.90 & 1.67 & 2.88 & 50.93 & 61.34 & 1.56 & 2.76 & 48.39 & 64.61 & 1.73 & 2.88 & 51.29 \\
\texttt{window\_size = 10} & 65.02 & 1.80 & 2.91 & 52.10 & 64.71 & 1.71 & 2.89 & 51.30 & 60.53 & 1.54 & 2.76 & 47.98 & 64.41 & 1.74 & 2.88 & 51.34 \\
\bottomrule
\end{tabular}
}
\end{table*}

\section{More Experimental Results}
\label{sec:exp_res}
\noindent{\bf Quantitative Results of Abnormal Cases.}
More evaluation results of abnormal cases are provided in \cref{tab:eval_abnormal_app}.

\noindent{\bf More Results of Inference Strategies.}
To strengthen the generality of our conclusions, we also evaluate Ovis2.5-9B \cite{ovis2_5} (non-thinking mode) under two inference strategies. The trends shown in \cref{fig:infer_score_ovis}, \cref{fig:infer_length_ovis}, and \cref{fig:infer_time_ovis} are broadly consistent with those observed for Qwen3-VL-8B-Instruct \cite{qwen3vl} (in main text \cref{sec:non_thinking_model}). Notably, under multi-turn inference, Ovis2.5-9B achieves a larger gain in $S_{uni}$ than Qwen3-VL-8B-Instruct: when \texttt{seg\_num=11}, their $S_{uni}$ scores on the \texttt{Full} set increase by 4.22 and 1.69, respectively, relative to the default setting. However, the additional computational overhead remains non-negligible.

\noindent{\bf Quantitative Results of Inference Strategies.}
The quantitative results of Qwen3-VL-8B-Instruct and Ovis2.5-9B under different inference strategies are reported in \cref{tab:eval_infer} and \cref{tab:eval_infer_ovis}, respectively.

\begin{figure}[t]
\vspace{-2mm}
	\centering
	\small
		\begin{overpic}[width=0.90\linewidth]{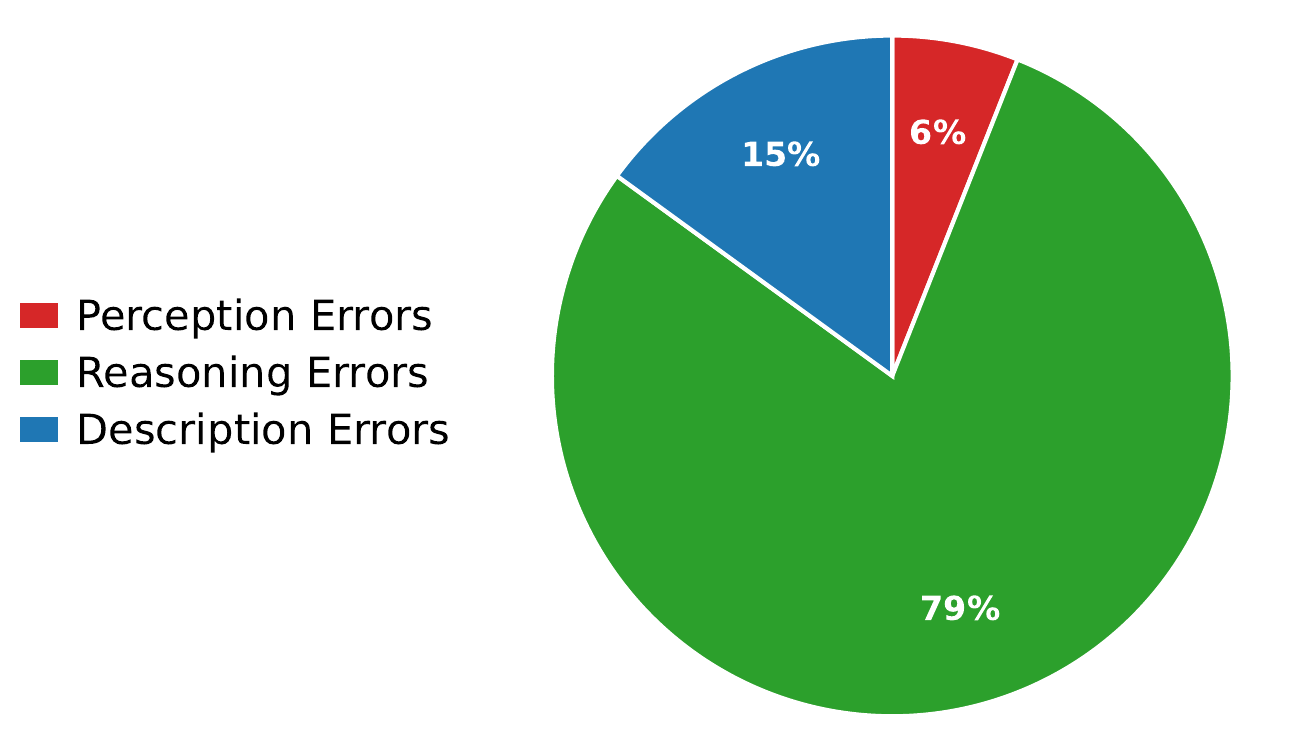}
	\end{overpic}
	\caption{Human-conducted analysis of errors by type.}
\label{fig:error_breakdown}
\vspace{-2mm}
\end{figure}

\noindent{\bf Error Analysis.} We analyze the errors of Qwen3-VL-8B-Instruct on a 100-instance subset, categorizing them into three distinct types: (1) perception errors, \eg, incorrect recognition of object category, color, or other visible attributes; (2) reasoning errors, \eg, incorrect inference about how actions change the scene or violations of world modeling; (3) description errors, \eg, missing relevant details or producing incomplete scene descriptions despite otherwise correct understanding. \cref{fig:error_breakdown} shows that 79\% of errors stem from faulty long-horizon reasoning.

\begin{figure}[t]
\vspace{-2mm}
  \centering
  \begin{subfigure}{0.49\linewidth}
    \centering
    \includegraphics[width=\linewidth]{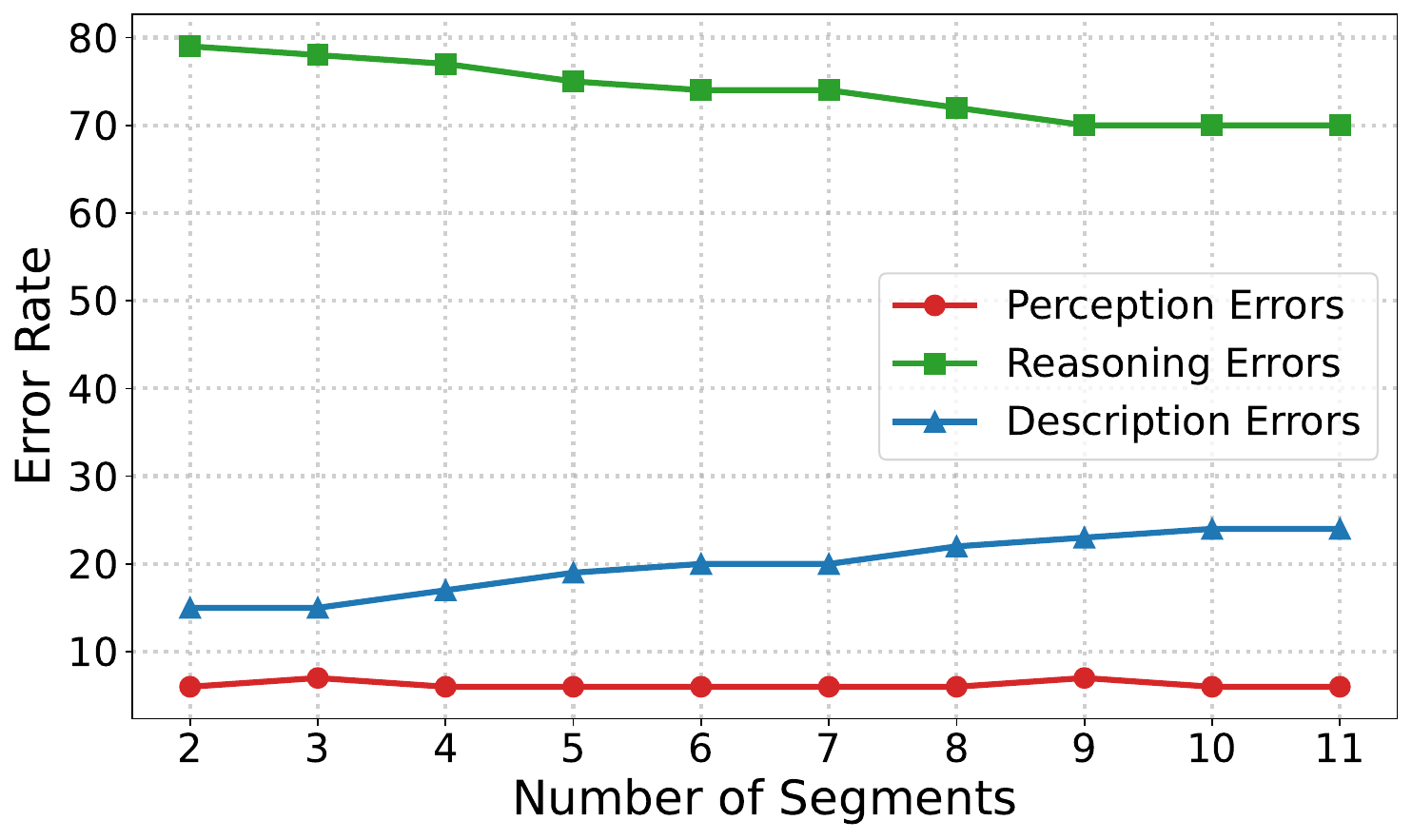} 
    \caption{Error rate \vs num. of segments}
    \label{fig:error_segment}
  \end{subfigure}
  \hfill
  \begin{subfigure}{0.49\linewidth}
    \centering
    \includegraphics[width=\linewidth]{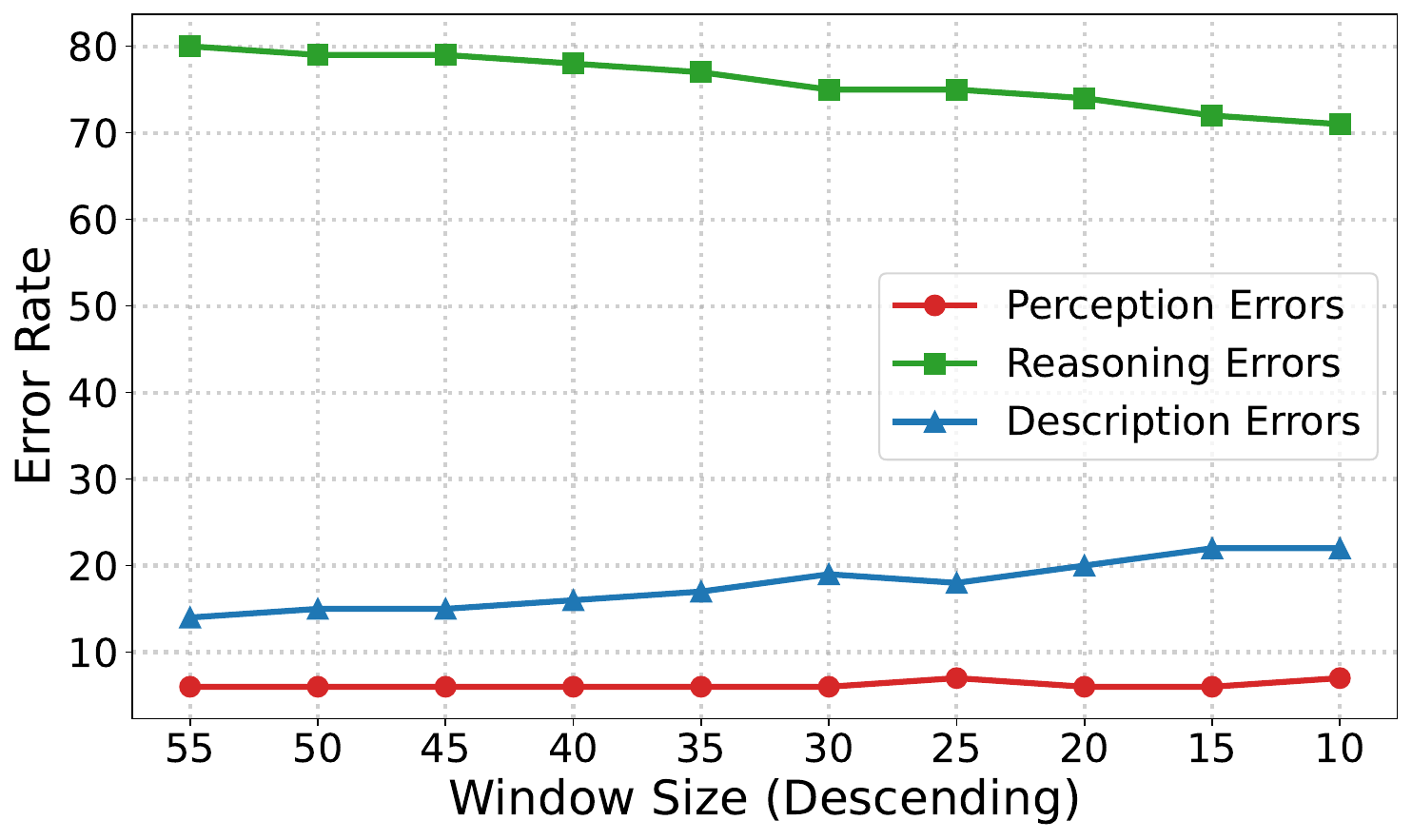}
    \caption{Error rate \vs window size}
    \label{fig:error_window}
  \end{subfigure}
  \caption{How different inference settings affect error types.}
  \label{fig:infer_error}
\vspace{-2mm}
\end{figure}

As shown in \cref{fig:infer_error}, for multi-turn inference, decomposing action sequences into finer-grained segments can reduce reasoning errors while increasing descriptive omissions, but has little effect on perception ability. This trend remains consistent in both single-turn and multi-turn inference.

\begin{figure}[t]
  \centering
  \includegraphics[width=\linewidth]{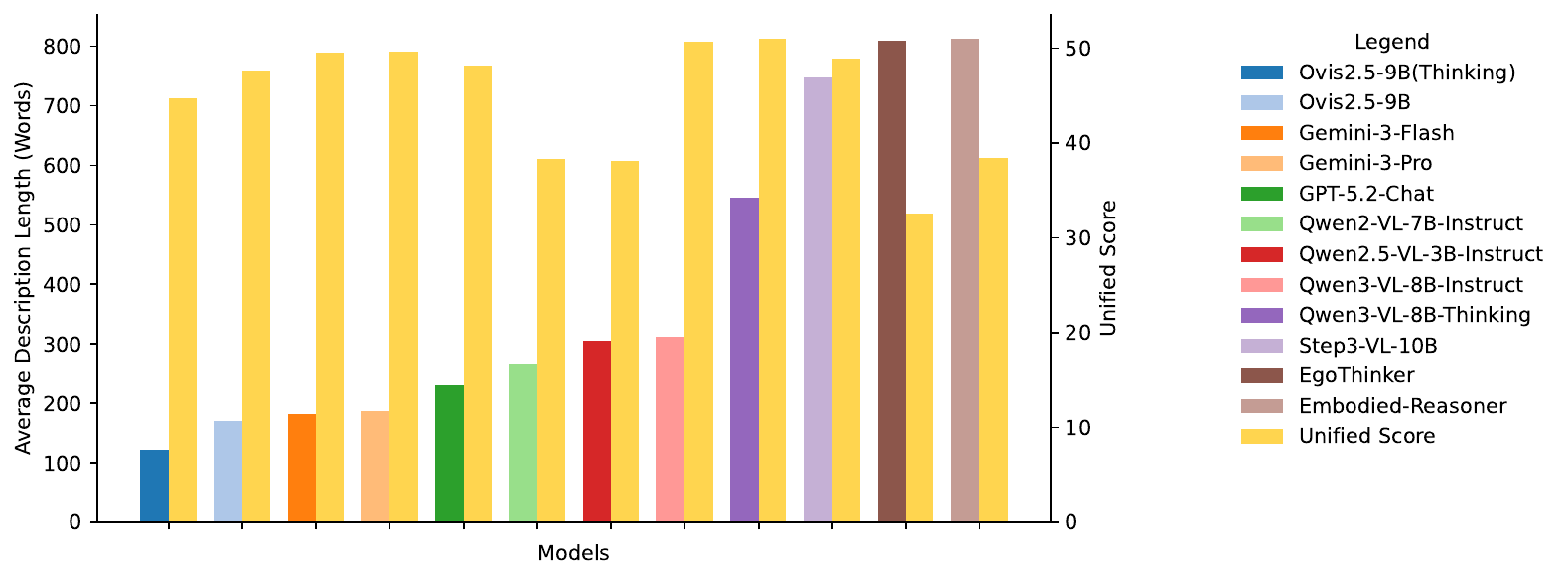}
  \caption{\textbf{Visual comparisons of the average description length and unified score $\bm{S_{uni}}$ on the full EXPLORE-Bench across different models.}}
  \label{fig:main_avg_len_score_bar}
\end{figure}

\begin{figure}[t]
  \centering
  \begin{subfigure}{0.235\linewidth}
    \centering
    \includegraphics[width=\linewidth]{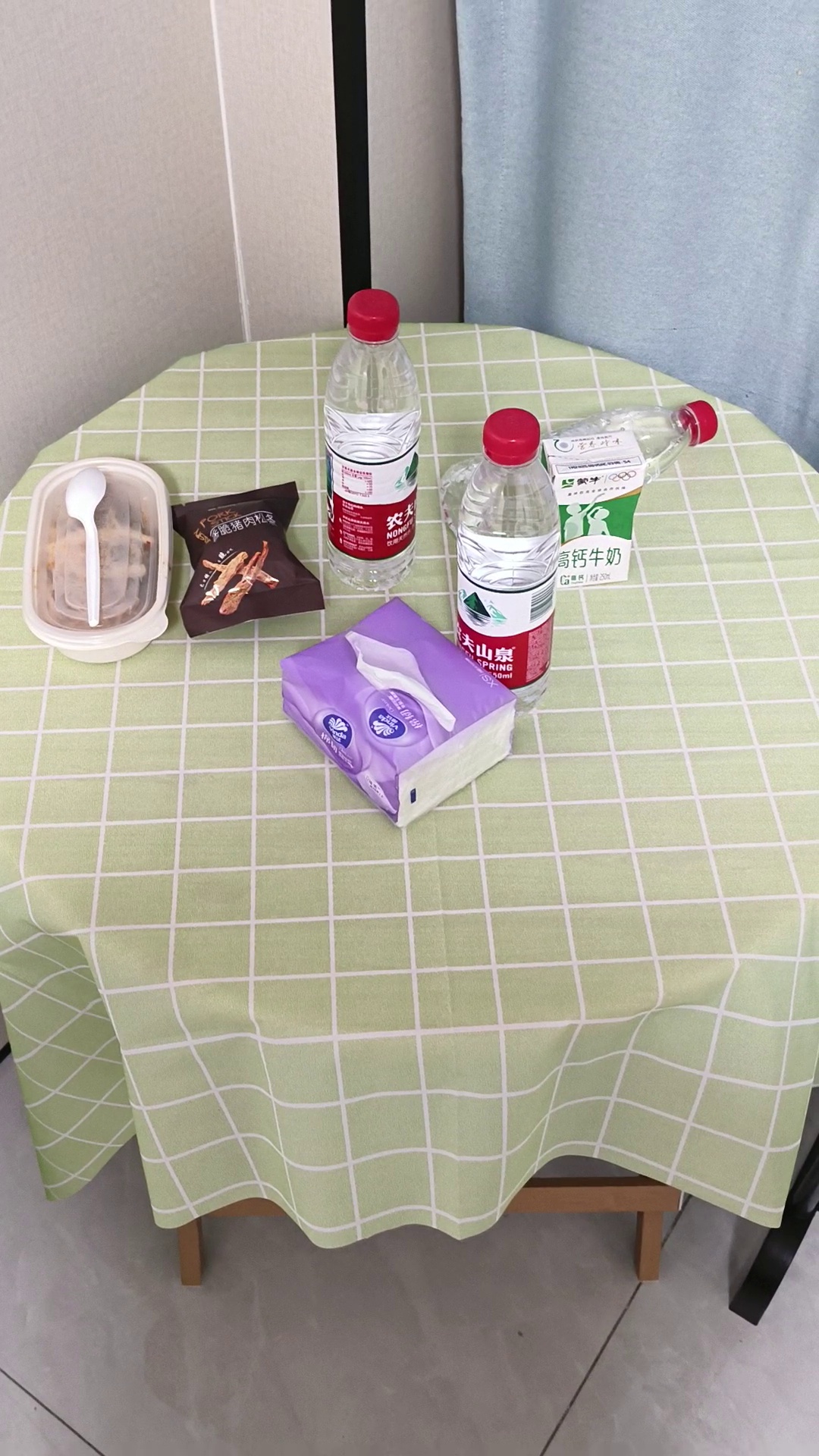} 
    \caption{Final-scene case 1}
    \label{fig:scene_1}
  \end{subfigure}
  \hfill 
  \begin{subfigure}{0.742\linewidth}
    \centering
    \includegraphics[width=\linewidth]{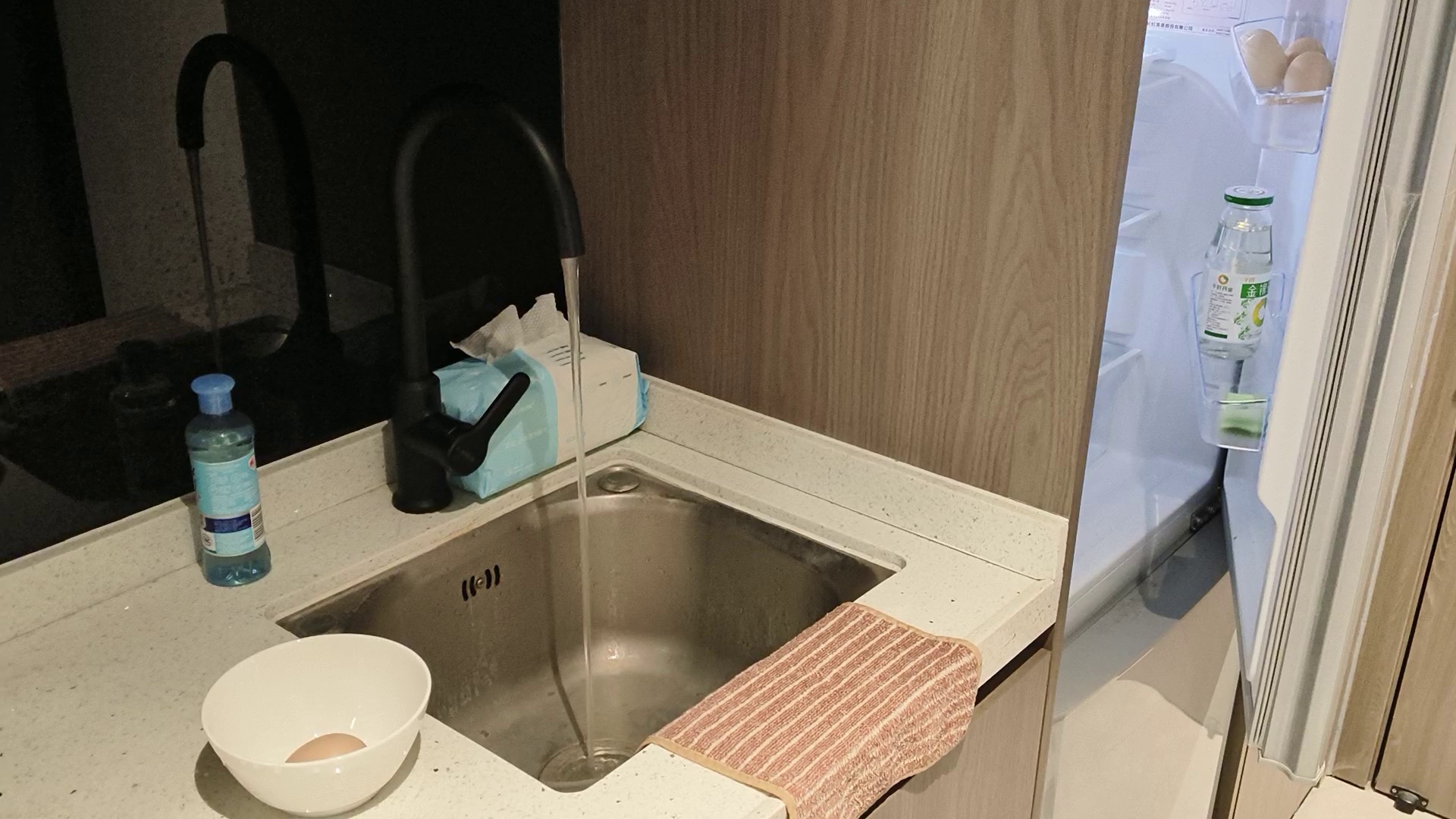}
    \caption{Final-scene case 2}
    \label{fig:scene_2}
  \end{subfigure}
  \caption{\textbf{Two cases of the final-scene images.}}
  \label{fig:scene_cases}
\end{figure}

\clearpage

\noindent{\bf Description Length.}
The visual comparisons of the average description length and unified score $S_{uni}$ on the full EXPLORE-Bench across different models are shown in \cref{fig:main_avg_len_score_bar}, which indicates that there is no clear correlation between description length and performance.

\section{More Cases and Case Studies}
\label{sec:cases}
In this section, we provide cases of final-scene annotations (\cref{sec:anno_case}) and model-predicted descriptions for the corresponding final scenes (\cref{sec:description_cases}). 

\subsection{Cases of Final-scene Annotations}
\label{sec:anno_case}
Here we present the annotations for the two final-scene images in \cref{fig:scene_cases}. \cref{lst:anno_case1} contains the annotation of the final scene shown in \cref{fig:abnormal_case_1v2} of the main text, so that readers can fully understand this case.

\begin{lstlisting}[style=prompt, caption={The scene annotation of case 1 (\cref{fig:scene_1})}, label={lst:anno_case1}]
[
  {
    "category": "table",
    "description": "round wooden table with light brown legs",
    "relation": [
      "table on floor",
      "table covered by tablecloth"
    ]
  },
  {
    "category": "tablecloth",
    "description": "light green fabric tablecloth with white grid pattern, slightly wrinkled and hanging down the sides",
    "relation": []
  },
  {
    "category": "bottle",
    "description": "transparent plastic water bottle with red cap and white and red label featuring green logo and white text, standing upright, filled with clear liquid",
    "relation": [
      "bottle on tablecloth",
      "bottle next to snack bag",
      "bottle next to bottle",
      "bottle next to tissue package",
      "bottle next to milk carton"
    ]
  },
  {
    "category": "bottle",
    "description": "transparent plastic water bottle with red cap and white and red label featuring green mountain logo and white text, standing upright, filled with clear liquid",
    "relation": [
      "bottle on tablecloth",
      "bottle next to bottle",
      "bottle next to tissue package",
      "bottle next to milk carton"
    ]
  },
  {
    "category": "bottle",
    "description": "transparent plastic water bottle with a red cap, lying horizontally, partially filled",
    "relation": [
      "bottle on tablecloth",
      "bottle next to bottle",
      "bottle next to milk carton"
    ],
    "key_state": "The bottle without label lies horizontally on the tablecloth."
  },
  {
    "category": "milk carton",
    "description": "white and green rectangular milk carton, standing upright",
    "relation": [
      "milk carton on tablecloth",
      "milk carton next to bottle"
    ]
  },
  {
    "category": "tissue package",
    "description": "purple rectangular tissue package with white text and logo, partially open with white tissue paper pulled out",
    "relation": [
      "tissue package on tablecloth",
      "tissue package next to bottle",
      "tissue package next to snack bag"
    ]
  },
  {
    "category": "lunch box",
    "description": "white elliptic plastic container with clear lid, containing food inside",
    "relation": [
      "lunch box on tablecloth",
      "lunch box next to snack bag"
    ]
  },
  {
    "category": "snack bag",
    "description": "dark brown plastic snack bag with printed text and images of meat sticks, sealed and lying flat",
    "relation": [
      "snack bag on tablecloth"
    ]
  },
  {
    "category": "spoon",
    "description": "white plastic spoon with a curved bowl and straight handle",
    "relation": [
      "spoon on lunch box"
    ]
  },
  {
    "category": "curtain",
    "description": "light blue fabric curtain hanging vertically",
    "relation": [
      "curtain behind table"
    ]
  },
  {
    "category": "wall",
    "description": "light beige wall with vertical seam and black vertical strip",
    "relation": [
      "wall behind table"
    ]
  },
  {
    "category": "floor",
    "description": "light gray tiled floor with visible grout lines, smooth surface",
    "relation": []
  }
]
\end{lstlisting}

\begin{lstlisting}[style=prompt, caption={The scene annotation of case 2 (\cref{fig:scene_2})}, label={lst:anno_case2}]
[
  {
    "category": "sink",
    "description": "stainless steel sink with rectangular basin",
    "relation": [
      "sink embodied in countertop"
    ]
  },
  {
    "category": "faucet",
    "description": "black matte-finish gooseneck faucet with lever handle, water flowing from spout",
    "relation": [
      "faucet over sink",
      "faucet on countertop",
      "faucet next to tissue box",
      "faucet near bottle"
    ],
    "key_state": "The water is flowing from the spout of the faucet."
  },
  {
    "category": "bottle",
    "description": "transparent plastic bottle with blue cap and label, partially filled with liquid",
    "relation": [
      "bottle on countertop"
    ]
  },
  {
    "category": "bowl",
    "description": "white ceramic bowl with smooth surface",
    "relation": [
      "bowl on countertop",
      "bowl next to sink"
    ]
  },
  {
    "category": "egg",
    "description": "brown oval-shaped egg with smooth shell",
    "relation": [
      "egg inside bowl"
    ]
  },
  {
    "category": "egg",
    "description": "brown oval-shaped egg with smooth shell",
    "relation": [
      "egg inside refrigerator"
    ]
  },
  {
    "category": "cloth",
    "description": "triped dishrag with alternating beige and brown horizontal stripes, textured terry cloth, spread out and laid flat",
    "relation": [
      "cloth on countertop",
      "cloth next to sink"
    ]
  },
  {
    "category": "countertop",
    "description": "light beige speckled countertop with smooth surface",
    "relation": [
      "countertop next to refrigerator"
    ]
  },
  {
    "category": "cabinet",
    "description": "wooden cabinet with light brown grain texture",
    "relation": [
      "cabinet under countertop"
    ]
  },
  {
    "category": "refrigerator",
    "description": "white refrigerator with open door, revealing interior shelves and door compartments",
    "relation": [
      "refrigerator containing eggs"
    ],
    "key_state": "The door of the refrigerator is open."
  },
  {
    "category": "tissue box",
    "description": "white and light blue tissue box with partially pulled-out tissue",
    "relation": [
      "tissue box on countertop"
    ]
  }
]
\end{lstlisting}

\subsection{Case Studies}
\label{sec:description_cases}
\begin{figure}[h]
  \centering
  \includegraphics[width=\linewidth]{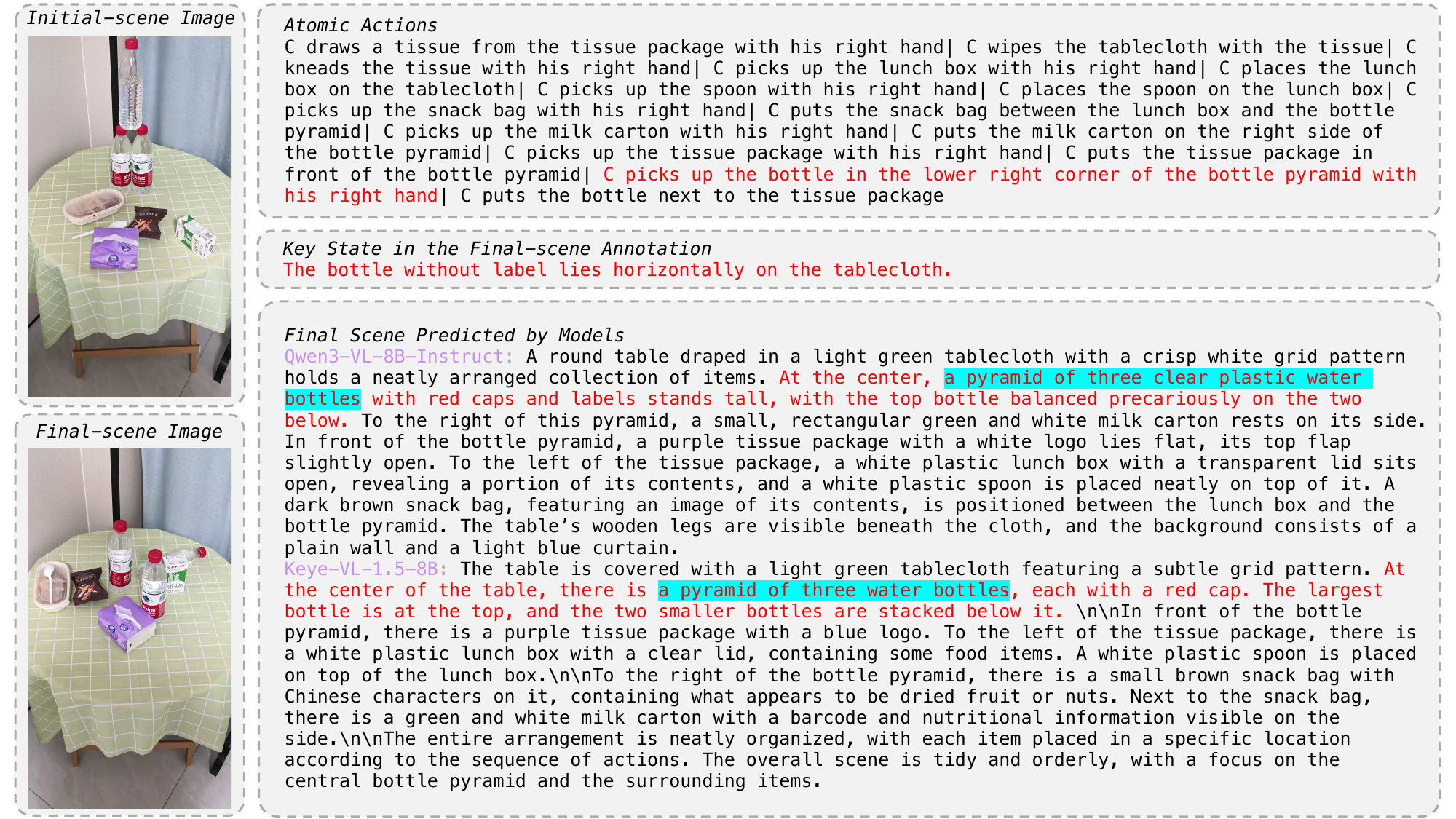}
  \caption{\textbf{Case study of case 1.} ``C'' refers to the camera wearer. Note that the scene in this case is the same as in \cref{fig:abnormal_case_1v2} of the main text, but we show predictions from different models, demonstrating their consistent failures in this case. Notably, Qwen3-VL-8B-Instruct and Keye-VL-1.5-8B cannot even realize that the bottle pyramid changes.}
  \label{fig:abnormal_case_1v3}
\end{figure}

\begin{figure}[h]
  \centering
  \includegraphics[width=\linewidth]{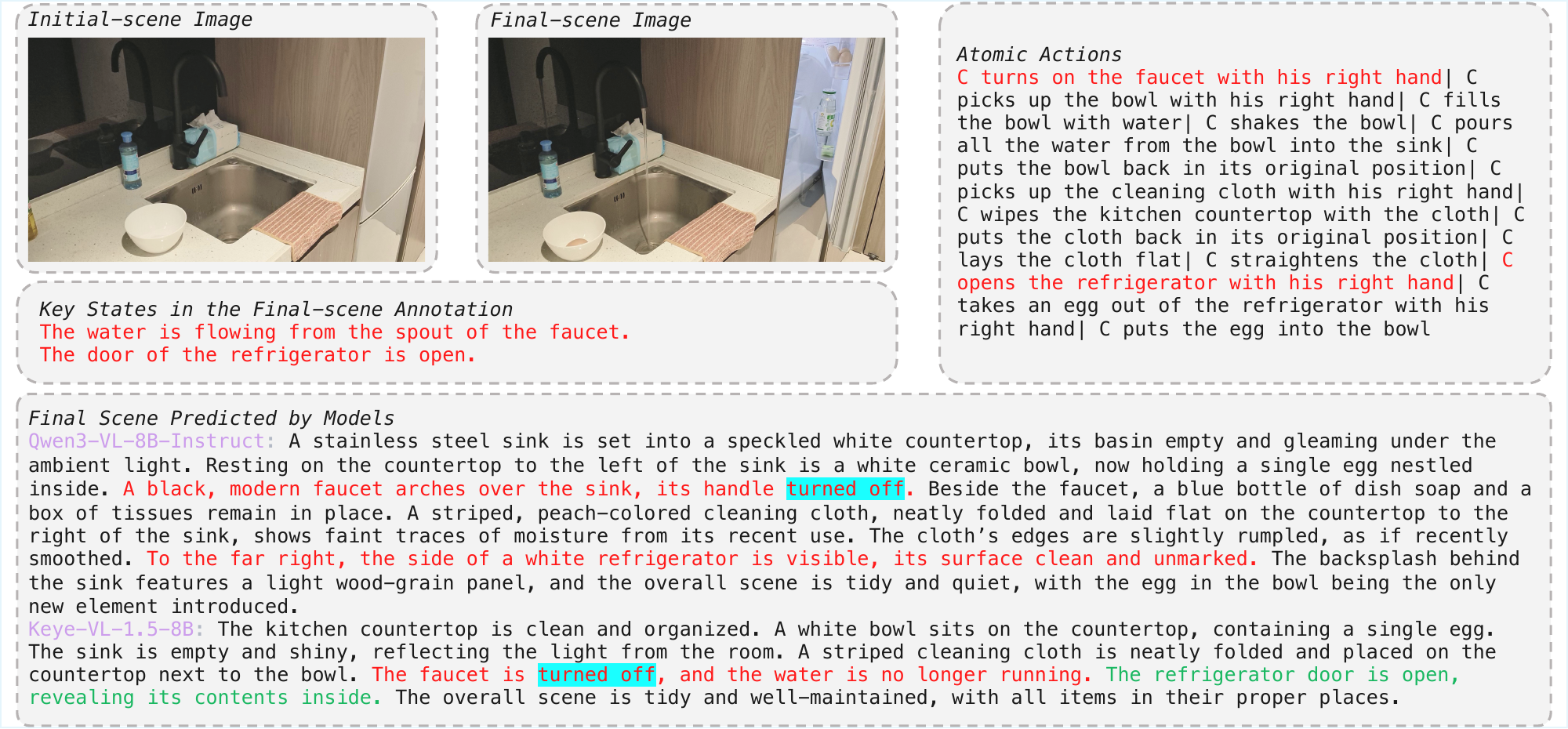}
  \caption{\textbf{Case study of case 2 (1).} ``C'' refers to the camera wearer. Qwen3-VL-8B-Instruct describes more objects than Keye-VL-1.5-8B, but it fails to mention that the refrigerator door is still open, whereas Keye-VL-1.5-8B notices this. In addition, both models claim that the faucet is turned off, which is inconsistent with the actual scene.}
  \label{fig:abnormal_case_2v1}
\end{figure}

\begin{figure}[h]
  \centering
  \includegraphics[width=\linewidth]{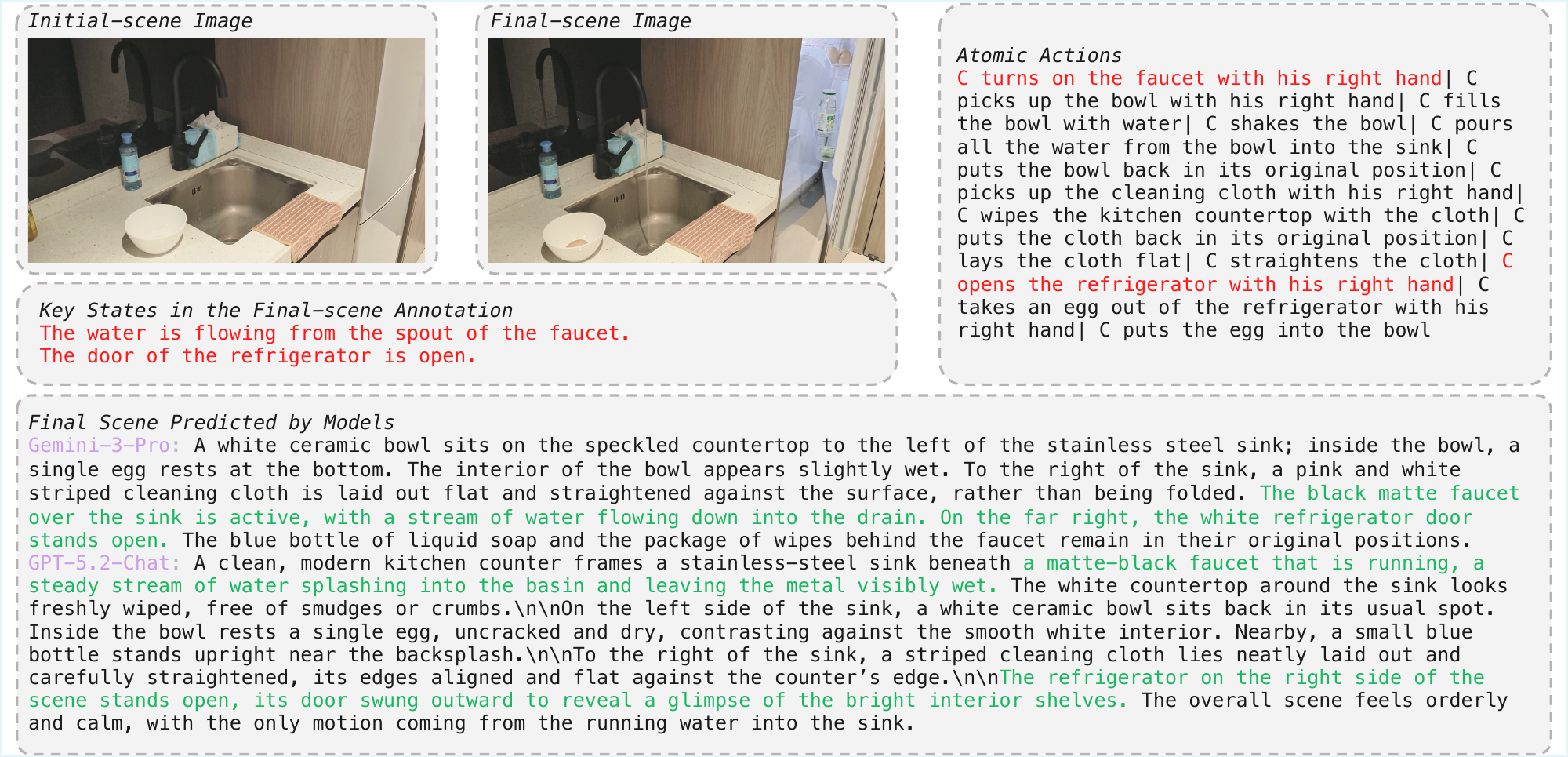}
  \caption{\textbf{Case study of case 2 (2).} ``C'' refers to the camera wearer. Both Gemini-3-Pro and GPT-5.2-Chat describe the key states in this scene, demonstrating their relative advantage in detecting safety hazards.}
  \label{fig:abnormal_case_2v2}
\end{figure}

\begin{figure}[h]
  \centering
  \includegraphics[width=\linewidth]{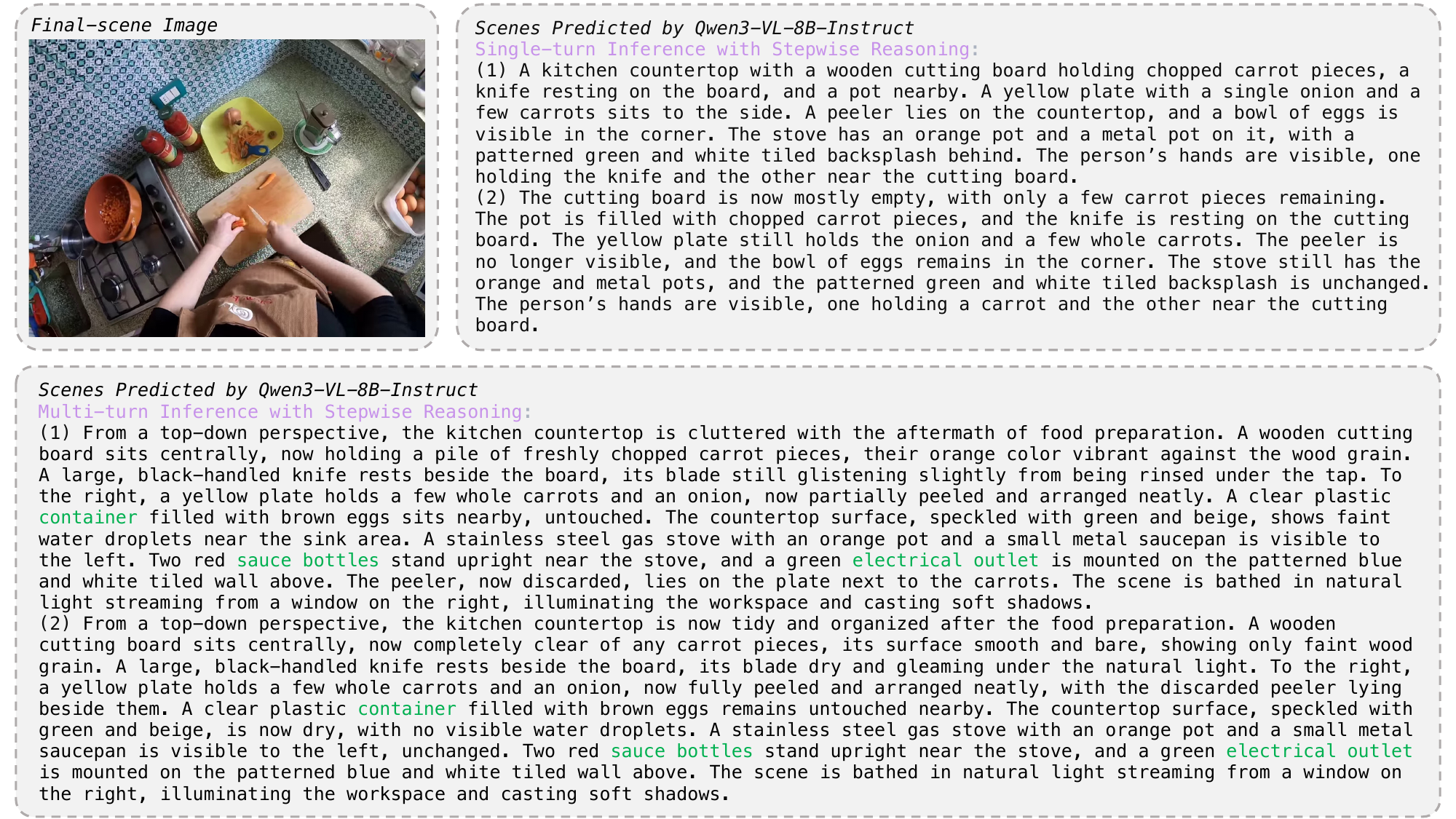}
  \caption{\textbf{Case study of stepwise reasoning implemented by single-turn inference \vs multi-turn inference.} In this case, \texttt{segment\_num=2}. (1) and (2) denote the model-predicted descriptions of the intermediate scene and the final scene, respectively. The descriptions generated by multi-turn inference are clearly longer than those generated by single-turn inference, covering more objects and providing more detailed descriptions of object attributes and relations.}
  \label{fig:single_vs_multi_turn}
\end{figure}

Detailed case studies of model-predicted final-scene descriptions are shown in \cref{fig:abnormal_case_1v3}, \cref{fig:abnormal_case_2v1}, and \cref{fig:abnormal_case_2v2}. A case study of stepwise reasoning implemented by single-turn inference \vs multi-turn inference is presented in \cref{fig:single_vs_multi_turn}.

\clearpage

\begin{figure}[h]
  \centering
  \includegraphics[width=\linewidth]{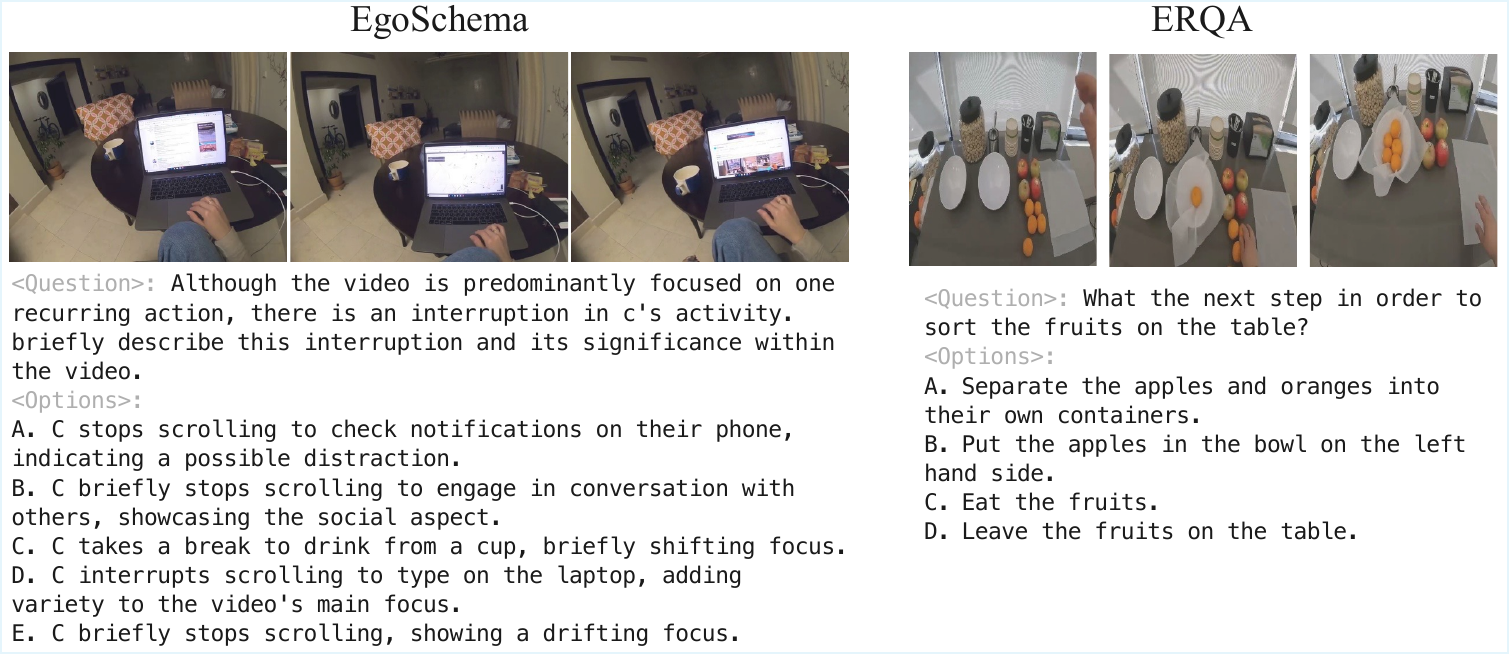}
  \caption{\textbf{Instances from other egocentric benchmarks.}}
  \label{fig:comparison}
\end{figure}


\section{Comparisons with related works}
\label{sec:work}
In works like EgoSchema \cite{mangalam2023egoschema}, ERQA \cite{team2025gemini}, and EgoPlan-Bench\cite{chen2026egoplan}, actions are mainly represented through event descriptions or next-step prediction. Their action sequences are short and are primarily evaluated in a multiple-choice format (as shown in \cref{fig:comparison}). Embodied Agent Interface \cite{li2024embodied} and EmbodiedBench \cite{yang2025embodiedbench} evaluates MLLMs for embodied planning by focusing on end-to-end execution success.
In contrast, our benchmark requires models to predict the causal effects of a sequence of atomic actions on the scene, with longer action sequences (as shown in \cref{fig:abnormal_case_1v3}) and open-ended questions. Moreover, we provide a fine-grained, systematic scene-level evaluation across objects, attributes, and relations.
Compared to ENACT's \cite{wang2025enact} symbolic forward/inverse world modeling (sequence reordering via scene-graph deltas in simulation), EXPLORE-Bench operates on real videos, conditions on long textual action sequences, and evaluates holistic visual end-state consistency, making it a complementary real-world testbed.

\section{Broader Impacts and Limitations}
\label{sec:discussion}
\noindent{\bf Broader Impacts.}
EXPLORE-Bench targets a capability central to embodied operation: anticipating the cumulative effects of actions, which is aligned with practical needs in household agents. The measured deficits of current models, especially on abnormal and safety-critical cases, highlight gaps that must be closed before real-world deployment. We hope EXPLORE-Bench can catalyze progress in consequence-aware reasoning and safer egocentric agents.

\noindent{\bf Limitations.} 
Although our work is meaningful, it still has limitations, including potential evaluation bias in LLM-as-a-judge \cite{gu2026survey} and the limited scale of abnormal cases. We hope this work encourages the community to address these limitations together with us.

\end{document}